\documentclass[fleqn,10pt]{wlscirep}

\usepackage{graphicx}

\usepackage{soul}

\usepackage[utf8]{inputenc} % allow utf-8 input
\usepackage[T1]{fontenc}    % use 8-bit T1 fonts
\usepackage[colorlinks=true]{hyperref}       % hyperlinks
\usepackage{url}            % simple URL typesetting
\usepackage{booktabs}       % professional-quality tables
\usepackage{amsfonts}       % blackboard math symbols
\usepackage{nicefrac}       % compact symbols for 1/2, etc.
\usepackage{microtype}      % microtypography
\usepackage{color}
\usepackage{subcaption}
\usepackage{lineno}
\usepackage{longtable}
\usepackage{enumitem}
\usepackage{amsmath}
\usepackage{amsthm}
\usepackage{float}
\usepackage{multirow}
\usepackage{adjustbox}
\usepackage{makecell}

\theoremstyle{definition}

\newtheorem{dataset}{Dataset}
\newtheorem{problem}{Problem}

\newcommand{\specialcell}[2][c]{%
  \begin{tabular}[#1]{@{}c@{}}#2\end{tabular}}

\title{A Comprehensive Benchmark for COVID-19 Predictive Modeling Using Electronic Health Records in Intensive Care}

\author[2,3,$\dag$]{Junyi Gao}
\author[1,$\dag$]{Yinghao Zhu}
\author[1,$\dag$]{Wenqing Wang}
\author[1]{Zixiang Wang}
\author[4]{Guiying Dong}
\author[5]{Wen Tang}
\author[6]{Hao Wang}
\author[1]{Yasha Wang}
\author[2]{Ewen M. Harrison}
\author[1,*]{Liantao Ma}
\affil[1]{Peking University, Beijing, CN}
\affil[2]{University of Edinburgh, Edinburgh, UK}
\affil[3]{Health Data Research UK}
\affil[4]{Peking University People's Hospital, CN}
\affil[5]{Peking University Third Hospital, CN}
\affil[6]{Kaihong Digital, CN}
\affil[*]{corresponding author: Liantao Ma (malt@pku.edu.cn)}

\affil[$\dag$]{these authors contributed equally to this work}
\begin{abstract}
The COVID-19 pandemic highlighted the need for predictive deep learning models in healthcare. However, practical prediction task design, fair comparison and model selection for clinical applications remain a challenge. To address this, we introduced and evaluated two new prediction tasks - Outcome-specific length-of-stay and Early mortality prediction for COVID-19 patients in intensive care - which better reflect clinical realities. We developed evaluation metrics, model adaptation designs, and open-source data preprocessing pipelines for these tasks, while also evaluating 18 predictive models, including clinical scoring methods, traditional machine learning, basic deep learning, and advanced deep learning models tailored for EHR data. Benchmarking results from two real-world COVID-19 EHR datasets are provided, and all results and trained models are released on an online platform for use by clinicians and researchers. Our efforts contribute to the advancement of deep learning and machine learning research in pandemic predictive modeling.

\end{abstract}

\begin{document}

\maketitle

\section{Introduction}
The COVID-19 pandemic has undoubtedly left an indelible impact on the global community. Although recent studies suggest that new COVID-19 variants are less lethal, their heightened transmissibility contributes to a continual surge in cases worldwide~\cite{callaway2021bad}. Given these circumstances, the critical need for early risk prediction and disease progression estimation, especially for COVID-19 patients in intensive care units (ICUs), is essential to optimally allocate medical resources and alleviate the strain on our healthcare system in the post-pandemic era.

%Electronic health record (EHR) data-based intelligent models have emerged as robust solutions to this dilemma. Over the past two years, numerous machine learning and deep learning models have been proposed for COVID-19 clinical predictive tasks, including diagnosis prediction~\cite{zoabi2021machine,feng2020early}, length-of-stay (LOS) prediction~\cite{ma2021distilling,dan2020machine}, and severity and mortality prediction~\cite{yan2020interpretable,wynants2020prediction,jamshidi2021using,martin2022characteristics,nachega2022assessment,dominguez2021bayesian,oliveira2022comparison,bennett2021clinical,elliott2021covid, gao2022medml,subudhi2021comparing, yan2021continuously}, among others.

Despite the commendable performance of existing COVID-19 predictive modeling works on specific datasets~\cite{yan2020interpretable,jamshidi2021using,martin2022characteristics,nachega2022assessment,dominguez2021bayesian,oliveira2022comparison,bennett2021clinical,elliott2021covid, gao2022medml}, researchers and clinicians often encounter difficulties when attempting to apply state-of-the-art prediction models to new data. Questions arise regarding whether to employ deep learning models or traditional machine learning models, and how different models compare in terms of prediction performances. Though there are a few descriptive review works~\cite{wynants2020prediction,shakeel2021covid}, there is still a lack of a fair quantitative comparative framework for existing models. Various models have been applied to different datasets, many of which are either not publicly available or entail strict access restrictions. Direct comparison of their results is therefore problematic, and model selection for new data is equally challenging as re-implementing all existing models on new data is resource-intensive, particularly for clinicians. This limitation curtails the practical usage of these models and hinders further research, underlining the necessity for a benchmark for COVID-19 predictive modeling that facilitates model comparison using identical data and evaluation strategies.

Existing electronic health record (EHR) prediction benchmarks~\cite{pirracchio2016mortality,purushotham2018benchmarking,harutyunyan2019multitask,yeche2021hirid} predominantly rely on publicly available ICU datasets such as MIMIC-III~\cite{johnson2016mimic} and MIMIC-II~\cite{lee2011open}. These benchmarks compare various machine learning and deep learning models across multiple standardized prediction tasks, including mortality prediction, patient phenotyping, length-of-stay (LOS) prediction, etc. However, these efforts primarily evaluate the performance of basic machine learning and deep learning models like recurrent neural networks (RNN) or long-short term memory networks (LSTM), despite the presence of numerous advanced deep learning models specifically designed for EHR data and clinical tasks. Notably, comprehensive benchmarking results for machine learning, deep learning and more advanced models on publicly available datasets remain absent for COVID-19 predictive tasks.

Another concern lies in the fact that current COVID-19 prediction works largely mirror the task settings of previous EHR data-mining endeavors. Risk or mortality prediction models, for instance, typically output the predicted risk score at the final timestep of available EHR sequences\cite{yan2020interpretable, gao2022medml}. This methodology can be problematic as it may be too late to initiate life-saving treatments for high-risk COVID-19 patients (especially ICU patients) by the time their status at the last timestep is identified as critical~\cite{rand2022early,reyna2019early}. Simultaneously, the length-of-stay (LOS) prediction task is commonly formulated as a regression task\cite{harutyunyan2019multitask}, with the model outputting the predicted remaining LOS at each timestep. This approach presents a notable issue where high-risk and low-risk patients may exhibit the same LOS value, as both death and ICU discharge are defined as the end of the stay in previous benchmarking works~\cite{harutyunyan2019multitask,ma2020covidcare,gao2020dr,yeche2021hirid,purushotham2018benchmarking}. Although the model can theoretically learn non-linearity and map these statuses to distinct locations in the latent embedding space, the practical application may prove challenging for clinicians. Adaptations are thus required for both tasks to cater to the unique intensive care setting associated with COVID-19. Consequently, it is imperative to compare existing deep learning and machine learning models using more clinically practical prediction tasks, identical publicly available data, and consistent evaluation settings.

In this study, we aim to bridge these gaps by proposing a standardized and comprehensive benchmark for COVID-19 prediction tasks in intensive care units. This benchmark enables comparison among machine learning, basic deep learning, and state-of-the-art EHR-specific models. Our contributions are three-fold:

\begin{itemize}
\item \textbf{Task, metric and model designing}: We introduce two tasks based on the clinical practice for COVID-19 patients in ICUs: \textit{Outcome-specific length-of-stay prediction} and \textit{Early mortality prediction}. Distinct from existing length-of-stay prediction frameworks, the outcome-specific length-of-stay prediction is formulated as a multi-target task. This task simultaneously outputs the patient outcome and corresponding length-of-stay, enabling clinicians to distinguish progression statuses between low-risk and high-risk patients. The early mortality prediction task is designed to alert clinicians to high-risk patients as early as possible, thereby preempting potential treatment delays. We design specific evaluation metrics for both tasks, namely the \textit{early mortality score (ES)} and the \textit{outcome-specific mean absolute error (OSMAE)}, to assess model performance. We have also designed the multi-task training architecture for the LOS prediction task and a time-aware loss term that can significantly improve deep learning models' early prediction performances.

\item \textbf{Data preparation and preprocessing pipelines}: We have established data preprocessing pipelines that include cleaning, filtering, missing value imputation, and cohort construction for two real-world COVID-19 EHR datasets. Both datasets consist of heterogeneous longitudinal EHR data from ICUs. Features include lab tests, vital signs, diagnoses, and static demographic information.

\item \textbf{Modeling and benchmarking}: We have implemented and evaluated 18 state-of-the-art predictive models across the two tasks, including 1 clinical scoring method, 5 machine learning models, 6 basic deep learning models, and 6 deep learning predictive models specifically designed for multivariate time-series EHR data. We conduct fair and reproducible comparisons and provide detailed benchmarking results to foster further research in this field.

\end{itemize}

To the best of our knowledge, this is the first benchmarking effort for patient-level COVID-19 prediction tasks in ICUs. We have made our code publicly available, enabling others to build complete benchmarks and reproduce all results. Our well-structured data preprocessing and modeling modules can also be easily applied to generate customized tasks and results. The benchmark code and documentations can be accessed at \url{https://github.com/yhzhu99/pyehr}. Moreover, we have released all the benchmark experiment results and trained models on an online platform, which includes model performances with all hyperparameter combinations for both tasks and makes the results easy to query and download. The platform can be accessed at \url{https://pyehr.netlify.app}.

\section{Dataset Description and Problem Formulation}

The datasets used in this study and the proposed prediction tasks are as follows:

\subsection{EHR datasets for COVID-19 patients in intensive care}

In this work, we utilize two COVID-19 EHR datasets to conduct benchmark experiments. After performing an exhaustive search for publicly available COVID-19 EHR datasets worldwide, we selected two datasets based on ease of access and absence of regional restrictions:

\begin{dataset}[\textbf{Tongji Hospital COVID-19 Dataset (TJH)}]~\cite{yan2020interpretable}
The \textcolor{blue}{TJH} dataset comprises anonymized EHR data from 485 COVID-19 patients admitted to Tongji Hospital, China, between January 10 and February 24, 2020. The dataset includes 74 lab tests and vital signs, all of which are numerical features, as well as 2 demographic features (age and gender). This dataset is publicly accessible via GitHub (\url{https://github.com/HAIRLAB/Pre_Surv_COVID_19}). The dataset download script is included in our benchmark code.
\end{dataset}

\begin{dataset}[\textbf{HM Hospitales Covid Data Save Lives (CDSL)}]~\cite{hmh}
The \textcolor{red}{CDSL} dataset originates from the HM Hospitales EHR system and contains anonymized records of 4,479 patients admitted with a diagnosis of COVID-19 or suspected COVID-19 infection. The dataset includes heterogeneous medical features such as detailed information on diagnoses, treatments, admissions, ICU admissions, diagnostic imaging tests, laboratory results, and patient discharge or death status. The dataset is open to global researchers and can be accessed upon request (\url{https://www.hmhospitales.com/prensa/notas-de-prensa/comunicado-covid-data-save-lives}). Prospective users must complete a request form.
\end{dataset}

These datasets have been utilized in several mortality prediction studies~\cite{yan2020interpretable, ma2021distilling}. However, we found no existing well-organized data preprocessing pipeline code or benchmarking results for them, which prompted us to provide a comprehensive benchmark analysis and data-processing suite for these publicly available datasets, thereby facilitating related research.

\subsection{Problem formulation and evaluation metrics}

In this study, we formulate two tasks—\textit{Outcome-specific length-of-stay prediction} and \textit{Early mortality prediction}—to evaluate and compare the performances of various machine learning and deep learning models. These tasks, which are adaptations of common length-of-stay and mortality prediction models, are tailored to suit the requirements of COVID-19 intensive care settings.

\begin{problem}[\textbf{Outcome-specific length-of-stay prediction}] Length-of-stay prediction is an important task in clinical practice, which better facilitates clinical resource management.
We define outcome-specific length-of-stay prediction as a multi-target prediction task, which encompasses a binary outcome classification task and a length-of-stay regression task. At each time step $t$, the model generates two outputs: a predicted outcome $\hat{y}_{m,t}\in \{0,1\}$ (i.e., 1 for mortality and 0 for survival) and a predicted length-of-stay (LOS) $\hat{y}_{l,t} \geq 0$ indicating the remaining days corresponding to the predicted outcome or the end of the ICU stay. Compared to traditional length-of-stay prediction models that output only $\hat{y}_{l,t}$, our model provides a more comprehensive evaluation of a patient's progression, allowing clinicians to differentiate length-of-stay according to different outcomes. The model learning process may also benefit from this setting as patients with varying health statuses are explicitly modeled in the latent space.

% \begin{figure}[htbp]
%     \centering
%     \includegraphics[width=0.8\linewidth]{figures/los.pdf}
%     \caption{Illustrations of the \textit{OSMAE} Metric.}
%     \label{fig:los}
% \end{figure}

% \begin{figure}[h!]
%     \centering
%     \includegraphics[width=0.3\linewidth]{figures/mortality.pdf}
%     \caption{Illustration of the early mortality prediction score.}
%     \label{fig:mortality}
% \end{figure}

\begin{figure}[htbp]
\centering
\begin{subfigure}{.7\textwidth}
  \centering
  \includegraphics[width=.9\linewidth]{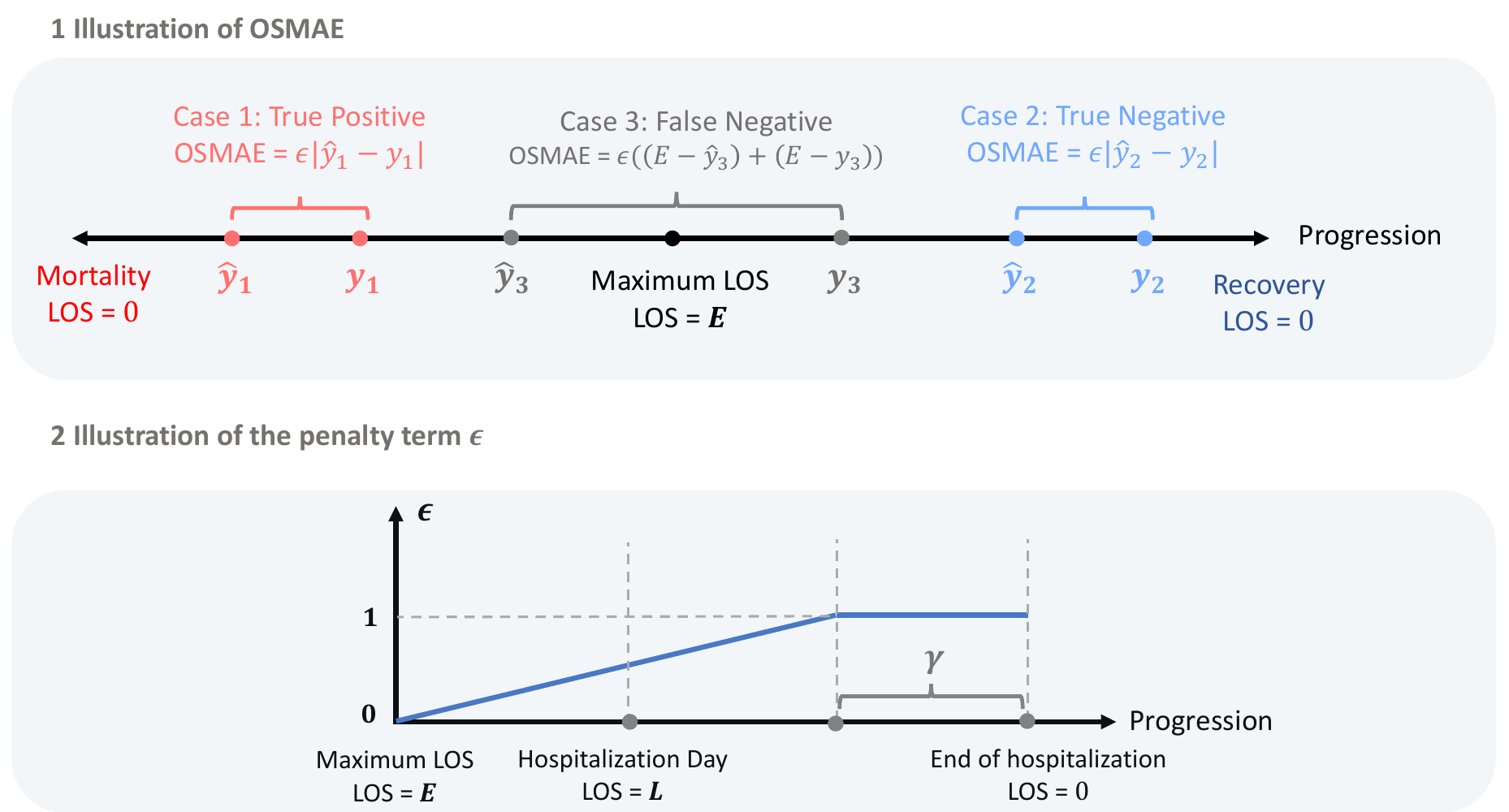}
  \caption{Illustrations of the \textit{OSMAE} metric.}
  \label{fig:los}
\end{subfigure}%
\begin{subfigure}{.3\textwidth}
  \centering
  \includegraphics[width=\linewidth]{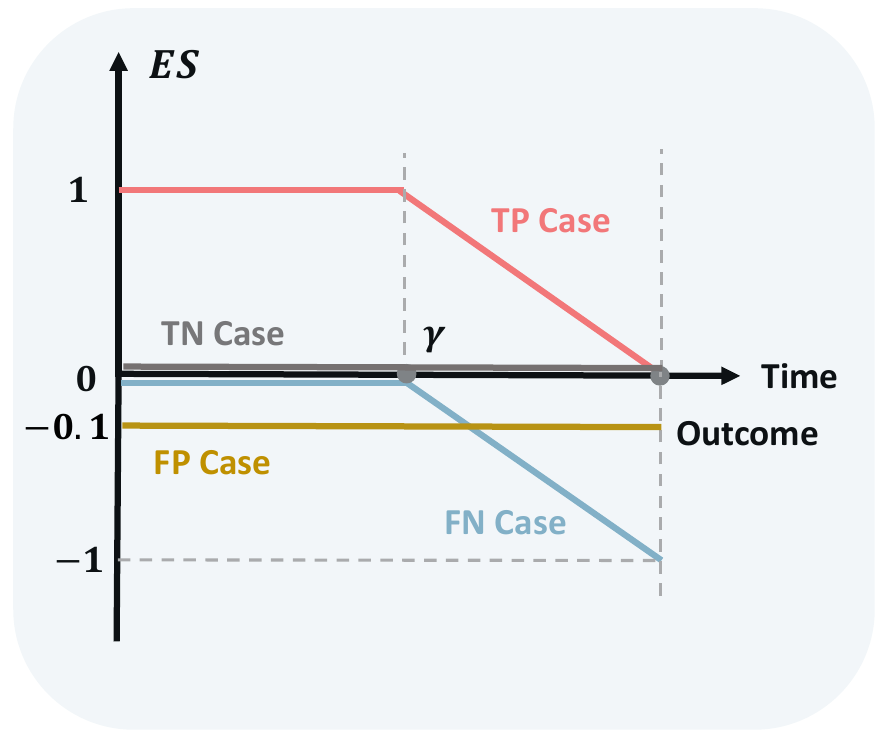}
  \caption{Illustration of the early mortality prediction score.}
  \label{fig:mortality}
\end{subfigure}
\caption{Illustrations of the proposed \textit{OSMAE} and \textit{ES} metrics.}
\label{fig:osmae_es_illustration}
\vskip -1em
\end{figure}

We use \textit{mean absolute error (MAE)} and \textit{mean squared error (MSE)} as evluation metrics. Furthermore, to evaluate this multi-target task comprehensively, we propose a new metric, \textit{Outcome-specific Mean Absolute Error (OSMAE)}, formulated as follows:

$$
    OSMAE = \left\{
    \begin{array}{ll}
        \epsilon(max(E-\hat{y}_l,0)+max(E-y_l, 0)) & , \hat{y}_m \neq y_m \\
        \epsilon|\hat{y}_l-y_l| & , \hat{y}_m = y_m
    \end{array} \right. \\
$$
$$
\epsilon = \left\{
    \begin{array}{ll}
        \frac{t}{E-L+\gamma} & ,0<t\leq L-\gamma \\
        1 & ,t>L-\gamma
    \end{array} \right. \\
$$

Here, $t$ represents the timestep, $L$ is the total length-of-stay, $E$ is the maximum value at the 95\% percentile of the length-of-stay of all patients. $\epsilon$ is a penalty term represented by a piecewise function. A lower \textit{OSMAE} indicates better model performance. Excluding the penalty term $\epsilon$, \textit{OSMAE} is equivalent to the conventional \textit{MAE} for true positive and true negative predictions, i.e., $\hat{y}_m = y_m$. However, for false positive and false negative patients, \textit{OSMAE} is significantly higher than the original \textit{MAE}, since the model's prediction for the patient's outcome deviates completely from the actual outcome. We introduce the penalty term to prevent the model from incurring a high \textit{OSMAE} during the early timesteps, as a patient's status may be uncertain during the initial ICU-stay phase. $\gamma$ serves as a penalty threshold, indicating that we expect the model to make accurate predictions within the final $\gamma$ days. The default value of $\gamma$ is $0.5*\bar{y}_{l,0}$, where $\bar{y}_{l,0}$ is the average value of all patients' total LOS. We provide a sensitivity analysis for $\gamma$ in the discussion section. Fig.~\ref{fig:los} illustrates the calculation of \textit{OSMAE} and the penalty term using three examples: true positive, false negative, and true negative cases.
\end{problem}

\begin{problem}[\textbf{Early mortality prediction}]
Early mortality prediction is defined as a binary sequential prediction task. For each timestep in the longitudinal EHR input sequence, the model will predict the mortality risk $\hat{y}_r\in \{0,1\}$ for this ICU stay, signifying whether the patient will succumb by the end of the ICU stay. An optimal early prediction model should raise an alert as soon as possible during the stay. Many existing mortality prediction models either feed the entire EHR sequence into the model and predict the mortality risk within a short future time window (e.g., 24h or 12h)~\cite{ma2020adacare,ma2020concare,yeche2021hirid,harutyunyan2019multitask}. However, this approach restricts the clinical applicability of these models. Particularly for COVID-19 patients in the ICU, their health statuses may have severely deteriorated by the final timestep, making the risks evident to clinicians. At this advanced stage, an alarm indicating high risk may be too late to initiate life-saving treatments or procedures. To address this limitation, we propose to assess early prediction performance using weighted metrics.

To evaluate models' early prediction performances, we not only utilize traditional measures such as \textit{Area Under the Receiver Operating Characteristic (AUROC)} and \textit{Area Under the Precision-Recall Curve (AUPRC)}, but we also introduce the \textit{Early Prediction Score (ES)} to specifically assess the early prediction performance of the models. Fig.\ref{fig:mortality} illustrates the \textit{ES}. Drawing from previous studies\cite{reyna2019early}, the idea is to assign a full score to early true positive predictions while penalizing late false negative predictions, i.e., the model fails to predict patients' outcomes accurately even in the final moments. We also give false alarms a small penalty (-0.1). Different penalty scores will not affect the relative comparison between different models. We provide details in the discussion section.  We normalize the total \textit{ES} for one patient as follows:
$$
    ES_{normalized} = \max{(ES_{total}/ES_{optimal},-1)}
$$
This normalization ensures that the ground truth (i.e., the highest possible score) receives a normalized score of 1 and the worst algorithm, which only outputs negative predictions, receives a normalized score of -1.
\end{problem}

%\yh{TOFIX: worst case may receives a normalized score of -1  (EStotal=-0.5, ESoptimal=0.1, then ES=-5, scaled to -1 (convert <-1 values to -1)) }

\section{Pipeline Design}
In this section, we introduce our benchmarking pipeline design, including data preprocessing, baseline selection, training and evaluation strategies.

\subsection{Data preprocessing}
Our benchmark data processing pipeline comprises four stages: data cleaning, merging, normalization, and imputation. Although both datasets are well-formatted data tables, they still contain artifacts that necessitate preprocessing. Previous machine learning and deep learning works on COVID-19 predictive modeling lack a uniform setting for these preprocessing details, leading to subtle performance differences. In this study, we propose a reproducible and reliable preprocessing pipeline to establish a fair comparison base for various models. 

\textbf{Data cleaning and merging: }Our first step in preprocessing the dataset into feature matrices is to structure the EHR data format for use in machine learning (ML) or deep learning (DL) models. Each patient's longitudinal EHR records are represented as a matrix, with each column representing a specific feature or label, and each row representing a record. We also extract demographic information for each patient as static data. During this step, we calculate ground truth labels, such as lengths of stay and mortality outcomes.

For error record cleaning, we first compute each feature's statistics, including the mean, standard deviation, minimum, maximum, median, and missing rate. For instance, in the \textcolor{blue}{TJH} dataset, we observed that all entries for the \textit{2019-nCoV nucleic acid detection} are identical, and some other feature values are negative. We remove these clear error records. For example, in the \textcolor{red}{CDSL} dataset, we remove obvious errors such as oxygen saturation values above 100 and maximum blood pressure exceeding 220. All filtered values are replaced with NaN (Not a Number). Detailed feature statistics can be found in Appendix I.

The raw datasets have record-level missing rates of 84.25\% for the \textcolor{blue}{TJH} dataset and 98.38\% for the \textcolor{red}{CDSL} dataset. High missing rates may impair models' prediction performance. Hence, following the previous benchmark preprocessing settings~\cite{harutyunyan2019multitask}, we merge data into day-level for the \textcolor{blue}{TJH} dataset and hour-level for the \textcolor{red}{CDSL} dataset, which reduces the missing rate of recorded physiological characteristics. It is worth noting that for the day/hour that does not have any record, we do not duplicate records from previous time steps. This data merging will not result in significant information loss since most records (over 95\%) do not record different values for the same feature on the same day or at the same hour. If there are multiple records for the same feature in the same time slot, we record their mean value. For the \textcolor{red}{CDSL} dataset, the feature dimensions are much larger and features are sparser than in the \textcolor{blue}{TJH} dataset: nearly 85\% of the features have a missing rate beyond 90\%, signifying that over 90\% of patients never have records for these features. Therefore, we remove features with a missing rate higher than 90\% among all patients.

\textbf{Data normalization and imputation:} We apply Z-score normalization to all demographic features, vital signs, lab test features, and the length-of-stays. Here, we calculate the mean and standard deviation based on the data in the 5\% to 95\% quantile range. Preventing future information leakage is crucial for maintaining the fairness of a benchmark. To address this, we implemented stringent strategies: (1) For missing values, we employed a forward-filling imputation method. Specifically, we replaced missing values with the most recent ones. If a patient lacked a prior record for a specific feature, we filled the missing data with the median value from all patients in the training set. (2) We utilized the means and standard deviations from the training set to normalize the entire dataset. This practice ensures no leakage regarding test set data distributions, a detail often overlooked in many previous benchmarking studies.

The data statistics of two processed datasets are shown in Table~\ref{tab:summary_statistics_tjh} and \ref{tab:summary_statistics_cdsl}.
\begin{table}[htbp]
    \footnotesize
    \centering
    \caption{\textit{Statistics of the \textcolor{blue}{TJH} dataset.} The reported statistics are of the form $Median [Q1, Q3]$.}
    \label{tab:summary_statistics_tjh}
\begin{tabular}{l|ccc}
\toprule
Mortality Outcome & Total & Alive & Dead\\
\midrule
\# Patients &  361  &  195 (54.02\%) & 166 (45.98\%)  \\
\# Records & 1704  & 1050 (61.62\%) & 654 (38.38\%) \\
\# Avg. records & 5.0 [3, 6]  & 5.0 [4, 7] & 3.0 [2, 5]  \\
\midrule
Age & 62.0 [46.0, 70.0] & 51.0 [37.0, 62.0] & 69.0 [62.25, 77.0]    \\
Age > Avg. (58) & 205 (56.79 \%) & 68 (34.87 \%) & 137 (82.53 \%) \\
Age $\leq$ Avg. (58) & 156 (43.21 \%) & 127 (65.13 \%) & 29 (17.47 \%) \\
\midrule
Gender & 58.7\% Male & 47.2\% Male & 72.3\% Male \\
Male & 212 (58.73 \%) & 92 (47.18 \%) & 120 (72.29 \%) \\
Female & 149 (41.27 \%) & 103 (52.82 \%) & 46 (27.71 \%) \\
\midrule
\# Features & \multicolumn{3}{c}{75} \\
Length of stay & 10.0 [4.0, 15.0] & 14.0 [9.0, 17.0] & 5.0 [3.0, 10.0]\\
\bottomrule
\end{tabular}
\end{table}

\begin{table}[htbp]
    \footnotesize
    \centering
    \caption{\textit{Statistics of the \textcolor{red}{CDSL} dataset.} The reported statistics are of the form $Median [Q1, Q3]$.}
    \label{tab:summary_statistics_cdsl}
\begin{tabular}{l|ccc}
\toprule
Mortality Outcome & Total & Alive & Dead \\
\midrule
\# Patients & 4255 & 3715 (87.31\%) & 540 (12.69\%) \\
\# Records  & 123044 & 108142 (87.89\%) & 14902 (12.11\%) \\
\# Avg. records & 24.0 [15, 39] & 25.0 [15, 39] & 22.5 [11, 37] \\
\midrule
Age & 67.2 {[}56.0, 80.0{]}      & 65.1 {[}54.0, 77.0{]}     & 81.6 {[}75.0, 89.0{]} \\
Age > Avg. (67) & 2228 (52.36 \%) & 1748 (47.05 \%) & 480 (88.89 \%) \\
Age $\leq$ Avg. (58) & 2027 (47.64 \%) & 1967 (52.95 \%) & 60 (11.11 \%) \\
\midrule
Gender & 59.1\% Male & 58.5\% Male & 63.3\% Male \\
Male & 2515 (59.11 \%) & 2173 (58.49 \%) & 342 (63.33 \%) \\
Female & 1740 (40.89 \%) & 1542 (41.51 \%) & 198 (36.67 \%) \\
\midrule
\# Features & \multicolumn{3}{c}{99} \\
Length of stay & 6.4 [4.0, 11.0] & 6.1 [4.0, 11.0] & 6.0 [3.0, 10.0]\\
\bottomrule
\end{tabular}
\end{table}

\subsection{Benchmarking experiment settings}
To provide a comprehensive benchmark comparison between existing models, we categorize the baseline models into four categories: clinical scoring models, machine learning models, basic deep learning models, and EHR-specific predictive models. Machine learning and basic deep learning models are popular for general classification and regression tasks. EHR-specific predictive models are designed explicitly for clinical predictive tasks with EHR data. We provide a list of baseline models below, and detailed descriptions of these models can be found in Appendix A:

\begin{enumerate}
    \item \textbf{Clinical scoring model}: 4C mortality score~\cite{knight2020risk}.
    \item \textbf{Machine learning models}: Decision tree (DT), Random forest (RF), Gradient Boosting Decision Tree (GBDT), XGBoost~\cite{chen2016xgboost}, CatBoost~\cite{dorogush2018catboost}.
    \item \textbf{Basic deep learning models}: Multi-layer perceptron (MLP), Recurrent neural network (RNN)~\cite{rumelhart1986learning}, Long-short term memory network (LSTM)~\cite{hochreiter1997long}, Gated recurrent units (GRU)~\cite{chung2014empirical}, Temporal convolutional networks (TCN)~\cite{bai2018empirical}, Transformer\cite{vaswani2017attention}.
    \item \textbf{EHR-specific predictive models}: RETAIN~\cite{choi2016retain}, StageNet~\cite{gao2020stagenet}, Dr.Agent~\cite{gao2020dr}, AdaCare~\cite{ma2020adacare}, ConCare~\cite{ma2020concare}, GRASP~\cite{zhang2021grasp}.
\end{enumerate}

\begin{figure}[htbp]
    \centering
    \includegraphics[width=\linewidth]{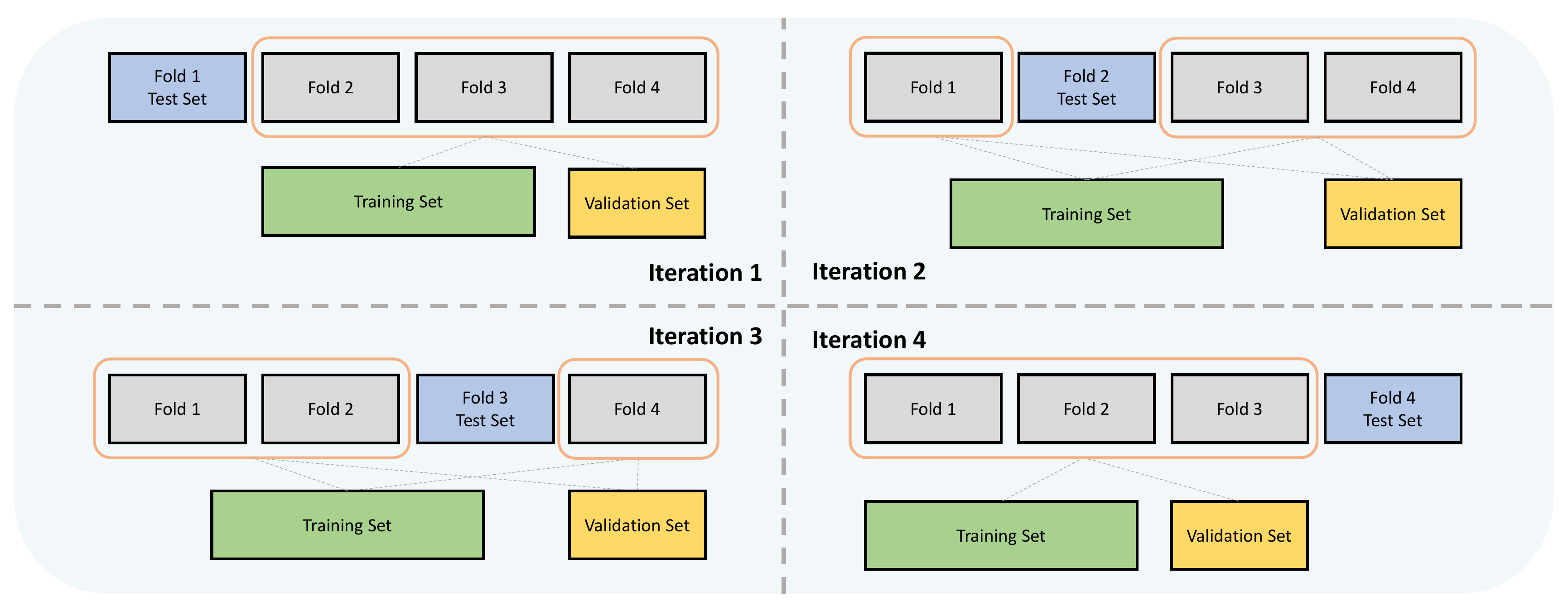}
    \caption{\textit{The K-fold cross-validation strategy}. We take 4-fold as an example in the figure. We use a stratified shuffle split to ensure the proportions of alive and dead patients on all folds are the same as the total cohort.}
    \label{fig:dataset_cross_validation}
\end{figure}

Machine learning models and some deep learning models (e.g., MLP) cannot handle sequential data as input. When training these models, we use the feature values at the current timestep as input and predict the target. The clinical scoring method is only applicable to the mortality prediction task. To minimize the randomness of the test sample selection process, we employ the stratified 10-fold cross-validation strategy to train and evaluate the models. We randomly divide patients into 10 groups or "folds". Then, we repeat the training and evaluation process 10 times. In the $i$-th iteration, we select the $i$-th fold as the test set and divide the remaining folds into training and validation sets at an 8:1 ratio. An illustration of this process is provided in Figure~\ref{fig:dataset_cross_validation}. Finally, we report the means and standard deviations of the model performances on all test sets from the 10 iterations. Moreover, to evaluate the model's generalizability over time more comprehensively, we have also conducted a standard holdout experiment. 
We divided the dataset into training, validation, and test sets in a 7:1:2 ratio, based on admission times (using the latest 20\% of patients as the holdout dataset). We retrained all models five times with different random seeds and report the standard deviations. The results for the hold-out performance are shown in Appendix D.

Experiments are conducted on a server equipped with dual Intel Xeon Silver 4210R CPUs, each with 10 cores supporting 20 threads, two NVIDIA RTX 3090 GPUs and 62GB RAM. Notably, all experiments are executed solely on one of the GPUs for consistent results. The model parameter size and average training time are shown in Figure~\ref{tab:computational_comparison}.

\begin{table}[htbp]
% \footnotesize
\centering
    \caption{\textit{Models' parameter size and training time (seconds).}}
    \label{tab:computational_comparison}
        \begin{tabular}{c|cc|cc}
        \toprule
        Dataset & \multicolumn{2}{c|}{TJH}                & \multicolumn{2}{c}{CDSL}   \\ \midrule
        Model   & \# Parameter & Runtime (1 epoch) & \# Parameter & Runtime (1 epoch) \\ \hline
        RF             & {/} & {1.71 s} & {/} & {26.69 s} \\
        DT             & {/} & {1.30 s} & {/} & {8.10 s} \\
        GBDT           & {/} & {4.04 s} & {/} & {128.55 s} \\
        CatBoost       & {/} & {1.54 s} & {/} & {7.40 s} \\
        XGBoost        & {/} & {1.50 s} & {/} & {9.40 s} \\ \midrule

        MLP            & {38.0 K} & {1.91 s} & {39.5 K} & {6.27 s} \\
        RNN            & {13.9 K} & {1.91 s} & {17.0 K} & {6.10 s} \\
        LSTM           & {41.0 K} & {1.89 s} & {48.6 K} & {6.30 s} \\
        GRU            & {31.9 K} & {1.85 s} & {38.1 K} & {6.84 s} \\
        TCN            & {55.8 K} & {3.31 s} & {60.4 K} & {31.60 s} \\
        Transformer    & {73.0 K} & {2.29 s} & {124 K} & {17.37 s} \\

        RETAIN         & {79.0 K} & {2.38 s} & {135 K} & {60.30 s} \\
        StageNet       & {603 K} & {2.31 s} & {622 K} & {17.81 s} \\
        Dr.Agent      & {41.9 K} & {2.89 s} & {48.1 K} & {15.81 s} \\
        AdaCare        & {114 K} & {3.05 s} & {130 K} & {21.38 s} \\
        GRASP          & {35.3 K} & {2.41 s} & {39.9 K} & {37.65 s} \\
        ConCare        & {66.4 K} & {2.59 s} & {89.1 K} & {11.98 s} \\

        \bottomrule
        \end{tabular}
\end{table}

Hyperparameters for the models are determined using a grid-search strategy on the validation set for each task. Hyperparameter settings can be found in Appendix B, and the grid search results are available on our online platform. All deep learning models are trained on the training set for 100 epochs, using an early stopping strategy. During each epoch, we assess the model's performance on the validation set of each fold. After training, we load the model parameters that yield the best performance on the validation set and then evaluate the model on the test set. For the outcome prediction task, we select models based on the \textit{AUPRC} score. For the length-of-stay prediction task, we use the \textit{MAE} score as the criteria. All metrics are computed on a per-record basis.

\subsection{Model adjustments for the proposed tasks}
This research additionally introduces model modifications tailored to the intended tasks in terms of both the loss function and the model architecture.

\begin{figure}[h!]
\centering
\includegraphics[width=.8\linewidth]{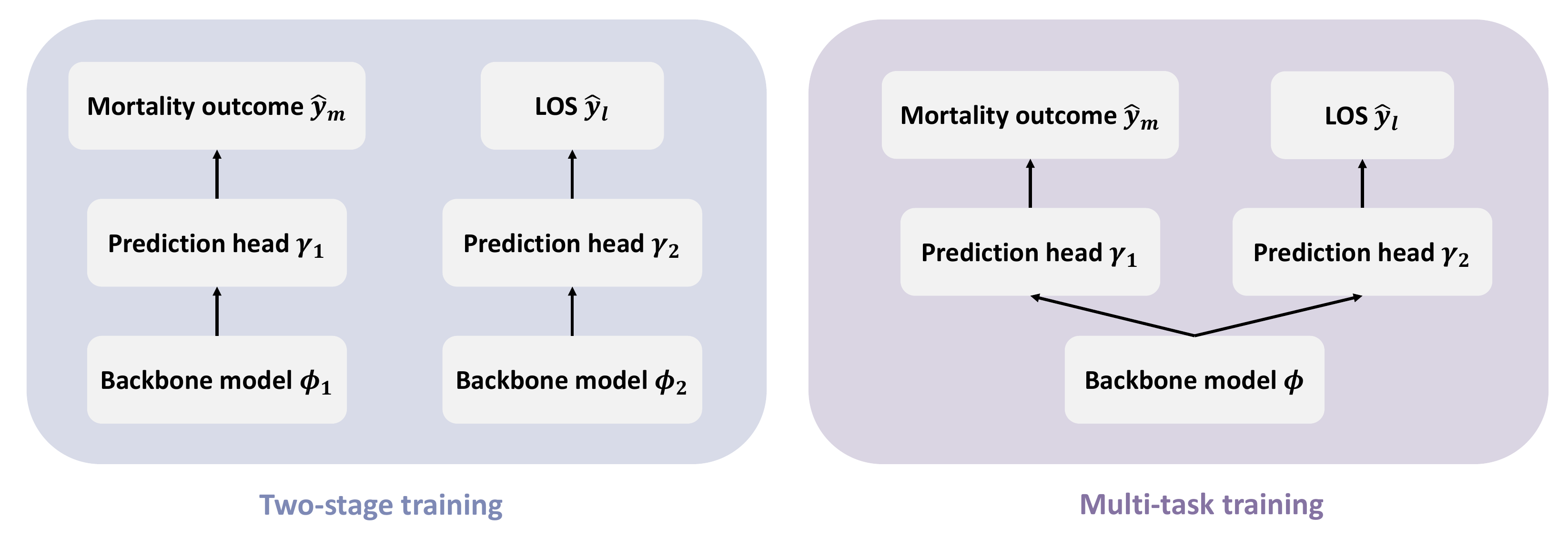}
\caption{Illustrations of the two-stage training and multi-task training settings}
\label{fig:settings}
\vskip -1em
\end{figure}

\textbf{Outcome-specific length-of-stay prediction:} The task of outcome-specific length-of-stay prediction is framed as a multi-target prediction task, for which we employ two distinct strategies. The first is \textit{end-to-end multi-task learning}, whereby a singular model backbone and two MLP prediction heads are trained - one for outcome prediction and the other for length-of-stay prediction. The backbone and prediction heads are co-trained in an end-to-end fashion. The second approach is the \textit{two-stage training}, in which two separate models, each with identical structures, are trained to predict the outcome and length-of-stay independently. Depictions of these two training scenarios can be found in Figure~\ref{fig:settings}. We include performance metrics for both approaches. It is important to note that traditional machine learning models only accommodate two-stage setting, hence we only present the performance metrics of the two-stage training approach for such models.

\textbf{Early mortality prediction:} Given the nature of the early mortality prediction task, we devised a straightforward yet potent loss function to guide deep learning models about the decision time threshold. The suggested \textit{time-aware loss}, is formulated as follows:

$$
    \mathcal{L}_{TA} = \frac{1}{T}\sum_t^T{(\mathcal{L}_{CE} * \exp(-\zeta l_t})) - \eta r
$$
$$
    r = \frac{1}{T}\sum_t^T{(l_t*|\hat{y}_t-y_t|)}
$$

 In this formula, $\mathcal{L}_{CE}$ is the cross-entropy loss, $l_t$ is the time to outcome at the $t$-th timestep, $\zeta$ denotes the decay rate, $\eta$ denotes the reward factor, and $r$ denotes the reward term. The second part confers rewards on earlier correct predictions and penalizes late incorrect predictions by either decreasing or increasing the loss values. The first part of the proposed loss function works to prevent early incorrect predictions from incurring excessive penalties through the application of an exponential decay function. This proposed time-aware loss function can be readily integrated into any deep learning model, thereby improving early mortality prediction performance. It should be noted that as the loss term is only employed during the training phase, the $l_t$ term is not required for the inference stage. Consequently, this ensures that there is no future information leakage at the inference stage.

\section{Results}
% [x] LOS distribution plot (wwq)
% [x] Feature distribution plot (wwq)
% [x] Calibration ECE table and curve (zyh)
% [x] Top 5 or 3 model ROC/PRC curve 4 figures (zyh AdaCare RETAIN Dr.Agent[TJH], ConCare RNN AdaCare [HM])
% [x] Top 5 EMP score bar plot 2 figures (wwq)
% [x] Top 5 OSMAE bar plot 2 figures (wwq)
% [x] Metric threshold (zyh \textcolor{red}{CDSL} dataset, OSMAE GRU, EMP GRU, use multitask pipeline)
In this section, we present comprehensive benchmarking results of all predictive models across all datasets and tasks. To assess the statistical significance of our model's performance, we performed t-tests on all results. The bootstrap T-test~\cite{dwivedi2017analysis} was employed to calculate p-values. We set the sample number for the bootstrap at 1000 and the number of sampling processes at 50. The comparisons were made between different variations of the model — for instance, comparing the model with and without Time-Aware (TA) loss in the early mortality prediction task, and between two-stage and multi-task settings in the LOS prediction task. The p-values are reported in Appendix G.

\subsection{Benchmarking performance of outcome-specific length-of-stay prediction}
The benchmarking performances for the task of outcome-specific length-of-stay prediction, under both multi-task and two-stage settings, are presented in Table~\ref{tab:outcome_specific_los_benchmark}. On the \textcolor{blue}{TJH} dataset, the GRU model with two-stage learning achieves the lowest \textit{MAE} and \textit{MSE}. Dr.Agent with multi-task learning demonstrates the lowest \textit{OSMAE}. On the \textcolor{red}{CDSL} dataset, StageNet with two-stage learning achieves the lowest \textit{MAE} and \textit{MSE}, while the TCN model with multi-task learning achieves the lowest \textit{OSMAE}.

\begin{table}[h!]
    \centering
    \caption{\textit{Benchmarking performance of outcome-specific length-of-stay prediction}}. The reported score is in the form of $mean \pm std$. Subscript $m$ signifies a multi-task learning strategy, while subscript $t$ indicates a two-stage learning strategy. \textbf{Bold} denotes the best performance. \underline{Underline} indicates that the multi-task setting outperforms the two-stage learning strategy. The asterisk * denotes that the performance improvement against the two-stage model is statistically significant (p-value < 0.05).
    \label{tab:outcome_specific_los_benchmark}
\begin{tabular}{l|ccc|ccc}
\toprule
Dataset & \multicolumn{3}{c}{TJH} & \multicolumn{3}{|c}{CDSL} \\
\midrule
Metric & MAE ($\downarrow$) & MSE ($\downarrow$) & OSMAE ($\downarrow$) & MAE ($\downarrow$) & MSE ($\downarrow$) & OSMAE ($\downarrow$) \\
\midrule
$\text{RF}_t$ & 4.83 $\pm$ 0.53 & 40.94 $\pm$ 8.77 & 6.14 $\pm$ 0.99 & 4.05 $\pm$ 0.13 & 31.30 $\pm$ 4.42 & 4.90 $\pm$ 0.16 \\ 
$\text{DT}_t$ & 5.07 $\pm$ 0.53 & 50.16 $\pm$ 12.06 & 7.02 $\pm$ 1.33 & 4.11 $\pm$ 0.15 & 32.23 $\pm$ 4.81 & 4.99 $\pm$ 0.14 \\ 
$\text{GBDT}_t$ & 4.79 $\pm$ 0.53 & 42.03 $\pm$ 9.58 & 5.92 $\pm$ 0.82 & 4.01 $\pm$ 0.14 & 30.74 $\pm$ 4.55 & 4.82 $\pm$ 0.19 \\ 
$\text{CatBoost}_t$ & 4.71 $\pm$ 0.53 & 39.72 $\pm$ 9.23 & 5.70 $\pm$ 0.85 & 4.01 $\pm$ 0.14 & 30.62 $\pm$ 4.61 & 4.79 $\pm$ 0.17 \\ 
$\text{XGBoost}_t$ & 4.78 $\pm$ 0.55 & 41.76 $\pm$ 9.47 & 5.77 $\pm$ 0.92 & 4.02 $\pm$ 0.13 & 31.10 $\pm$ 4.42 & 4.81 $\pm$ 0.15 \\ 
\midrule
\midrule
$\text{MLP}_t$ & 5.08 $\pm$ 0.48 & 47.21 $\pm$ 11.10 & 6.34 $\pm$ 0.97 & 4.05 $\pm$ 0.15 & 32.08 $\pm$ 4.96 & 4.80 $\pm$ 0.17 \\ 
$\text{MLP}_m$ & \underline{4.90 $\pm$ 0.52}* & \underline{44.35 $\pm$ 14.00}* & 7.26 $\pm$ 2.81 & 4.08 $\pm$ 0.16 & \underline{31.54 $\pm$ 4.53}* & 4.86 $\pm$ 0.21 \\ 
$\text{RNN}_t$ & 4.57 $\pm$ 0.48 & 39.82 $\pm$ 10.02 & 5.54 $\pm$ 1.30 & 3.90 $\pm$ 0.22 & 30.66 $\pm$ 5.58 & 4.37 $\pm$ 0.26 \\ 
$\text{RNN}_m$ & 4.68 $\pm$ 0.49 & 40.21 $\pm$ 9.29 & \underline{5.46 $\pm$ 1.43}* & \underline{3.87 $\pm$ 0.14}* & \underline{29.52 $\pm$ 4.98}* & 4.43 $\pm$ 0.18 \\ 
$\text{LSTM}_t$ & 4.49 $\pm$ 0.63 & 37.38 $\pm$ 12.64 & 4.88 $\pm$ 1.30 & 3.81 $\pm$ 0.16 & 29.29 $\pm$ 4.88 & 4.34 $\pm$ 0.23 \\ 
$\text{LSTM}_m$ & \underline{4.46 $\pm$ 0.60}* & 39.24 $\pm$ 10.30 & 5.53 $\pm$ 1.86 & 3.85 $\pm$ 0.13 & 30.13 $\pm$ 4.22 & 4.40 $\pm$ 0.19 \\ 
$\text{GRU}_t$ & \textbf{4.33 $\pm$ 0.49} & \textbf{35.10 $\pm$ 9.61} & 5.16 $\pm$ 1.31 & 3.75 $\pm$ 0.18 & 28.90 $\pm$ 4.68 & 4.34 $\pm$ 0.32 \\ 
$\text{GRU}_m$ & 4.80 $\pm$ 0.48 & 41.78 $\pm$ 9.49 & 5.38 $\pm$ 1.09 & 3.98 $\pm$ 0.18 & 32.14 $\pm$ 4.31 & 4.55 $\pm$ 0.22 \\ 
$\text{TCN}_t$ & 4.69 $\pm$ 0.60 & 43.29 $\pm$ 13.27 & 5.77 $\pm$ 0.92 & 3.75 $\pm$ 0.17 & 29.48 $\pm$ 4.78 & 4.31 $\pm$ 0.15 \\ 
$\text{TCN}_m$ & 4.79 $\pm$ 0.55 & \underline{41.56 $\pm$ 9.73}* & \underline{5.52 $\pm$ 0.85}* & 3.80 $\pm$ 0.16 & \underline{29.45 $\pm$ 4.47}* & \underline{\textbf{4.22 $\pm$ 0.18}}* \\ 
$\text{Transformer}_t$ & 5.04 $\pm$ 0.43 & 43.88 $\pm$ 7.22 & 5.96 $\pm$ 1.62 & 3.98 $\pm$ 0.20 & 32.35 $\pm$ 5.61 & 5.19 $\pm$ 0.19 \\ 
$\text{Transformer}_m$ & 5.06 $\pm$ 0.46 & 46.02 $\pm$ 11.63 & 6.61 $\pm$ 1.57 & 4.00 $\pm$ 0.20 & \underline{32.02 $\pm$ 4.68}* & \underline{5.13 $\pm$ 0.19}* \\ 
\midrule
\midrule
$\text{RETAIN}_t$ & 4.64 $\pm$ 0.61 & 41.22 $\pm$ 14.45 & 4.77 $\pm$ 1.18 & 3.83 $\pm$ 0.16 & 30.56 $\pm$ 5.99 & 4.41 $\pm$ 0.26 \\ 
$\text{RETAIN}_m$ & \underline{4.62 $\pm$ 0.55}* & \underline{40.06 $\pm$ 10.34}* & 5.03 $\pm$ 0.96 & 3.84 $\pm$ 0.20 & \underline{29.92 $\pm$ 4.48}* & 4.41 $\pm$ 0.30 \\ 
$\text{StageNet}_t$ & 4.60 $\pm$ 0.76 & 41.70 $\pm$ 14.21 & 5.59 $\pm$ 1.33 & \textbf{3.72} $\pm$ 0.15 & \textbf{28.81 $\pm$ 4.57} & 4.33 $\pm$ 0.22 \\ 
$\text{StageNet}_m$ & \underline{4.49 $\pm$ 0.42}* & \underline{39.55 $\pm$ 8.52}* & 7.10 $\pm$ 1.74 & 3.78 $\pm$ 0.18 & 29.93 $\pm$ 4.50 & \underline{4.28 $\pm$ 0.19}* \\ 
$\text{Dr.Agent}_t$ & 4.61 $\pm$ 0.58 & 40.85 $\pm$ 12.12 & 4.93 $\pm$ 1.14 & 3.80 $\pm$ 0.18 & 29.70 $\pm$ 4.96 & 4.46 $\pm$ 0.27 \\ 
$\text{Dr.Agent}_m$ & \underline{4.41 $\pm$ 0.58}* & \underline{37.55 $\pm$ 11.38}* & \underline{\textbf{4.75 $\pm$ 1.11}}* & 3.82 $\pm$ 0.19 & \underline{29.45 $\pm$ 4.70}* & 4.49 $\pm$ 0.28 \\ 
$\text{AdaCare}_t$ & 4.55 $\pm$ 0.61 & 39.67 $\pm$ 11.35 & 5.08 $\pm$ 0.62 & 3.80 $\pm$ 0.15 & 29.07 $\pm$ 4.57 & 4.42 $\pm$ 0.15 \\ 
$\text{AdaCare}_m$ & \underline{4.41 $\pm$ 0.63}* & \underline{38.86 $\pm$ 10.75}* & 5.24 $\pm$ 0.98 & 3.81 $\pm$ 0.16 & 32.05 $\pm$ 4.27 & 4.43 $\pm$ 0.18 \\ 
$\text{GRASP}_t$ & 4.57 $\pm$ 0.43 & 40.65 $\pm$ 11.46 & 5.51 $\pm$ 1.13 & 3.84 $\pm$ 0.16 & 29.95 $\pm$ 5.06 & 4.43 $\pm$ 0.25 \\ 
$\text{GRASP}_m$ & \underline{4.44 $\pm$ 0.50}* & \underline{37.79 $\pm$ 10.27}* & \underline{5.31 $\pm$ 1.31}* & 3.89 $\pm$ 0.12 & 30.01 $\pm$ 4.33 & 4.49 $\pm$ 0.28 \\ 
$\text{ConCare}_t$ & 4.69 $\pm$ 0.52 & 40.26 $\pm$ 11.13 & 5.29 $\pm$ 1.31 & 3.81 $\pm$ 0.17 & 30.16 $\pm$ 5.03 & 4.29 $\pm$ 0.23 \\ 
$\text{ConCare}_m$ & \underline{4.67 $\pm$ 0.56}* & 40.42 $\pm$ 10.77 & \underline{5.27 $\pm$ 1.35}* & 3.85 $\pm$ 0.11 & \underline{29.15 $\pm$ 3.90}* & 4.46 $\pm$ 0.24 \\ 
\bottomrule
\end{tabular}
\end{table}

We observe that the multi-task strategy generally outperforms the two-stage strategy on the \textcolor{blue}{TJH} dataset. On the TJH dataset, we find that EHR-specific models can better benefit from multi-task learning settings, suggesting that the multi-task setting may better facilitate these models' ability to map patients' diverse statuses in the latent space. We visualize the learned embeddings in the latent space using t-SNE in Appendix F. Additionally, it's important to note the discrepancy between \textit{OSMAE} and \textit{MAE}. Despite both metrics assessing absolute error, the \textit{OSMAE} values are notably larger than the \textit{MAE} values on both datasets. On average, the \textit{OSMAE} is 22\% higher than the \textit{MAE} on the \textcolor{blue}{TJH} dataset and 17\% higher on the \textcolor{red}{CDSL} dataset. This suggests that \textit{MAE} might potentially distort the evaluation of model performance, as it cannot measure the error for false negative and false positive predictions, despite its widespread use in previous LOS prediction studies. We also conduct the error analysis for the MAE in Appendix H. In our current setting, we utilize two identical model structures for two-stage training. However, future work could potentially involve the use of distinct models for the two tasks. Further research could also focus on designing a more sophisticated multi-task architecture or incorporating the principles of transfer learning.

\subsection{Benchmarking performance of early mortality prediction}

The comparative performances of the models in early mortality prediction are presented in Table~\ref{tab:early_motality_benchmark}. We also introduced the time-aware loss, labeling these adjusted models with a `-TA' suffix for all deep learning models. For the TJH dataset, the AdaCare-TA model delivered the highest \textit{AUPRC} and \textit{AUROC}, while Dr. Agent provided the highest \textit{ES}. In contrast, for the \textcolor{red}{CDSL} dataset, ConCare-TA achieved the highest \textit{AUPRC}, while StageNet-TA produced the highest \textit{AUROC} and \textit{ES}. Among traditional machine learning and basic deep learning models, CatBoost and GRU showed better performance on both datasets.

The results indicate that the proposed time-aware loss term can enhance the performance of most models, particularly in terms of the early prediction score. The average improvement of the \textit{ES} is 1.93\% for the \textcolor{blue}{TJH} dataset and 33.58\% for the \textcolor{red}{CDSL} dataset. This underscores that this simplistic loss term can effectively aid models in making correct early decisions. It's worth noting that our goal is to evaluate the direct effectiveness of the proposed time-aware loss term, so we did not specifically tune the hyper-parameters of all -TA models. So even under this adverse condition, most -TA models can still outperform the original version. Additionally, we observe that for the \textcolor{blue}{TJH} dataset, the performance of all models was higher, especially for the \textit{ES}, suggesting that the task is more straightforward on this dataset. Patients' initial status was also more severe in the \textcolor{blue}{TJH} dataset, resulting in a much higher \textit{ES} than in the \textcolor{red}{CDSL} dataset. This might be attributable to the fact that the \textcolor{blue}{TJH} data was collected during the initial months of the pandemic when the virus was significantly more lethal. The performance of EHR-specific models generally surpasses that of basic deep learning models, which in turn outperform traditional machine learning models. This is to be expected, as most EHR-specific models are better equipped to utilize characteristics present in EHR data, such as disease progression, while compared with deep learning models, traditional machine learning models are unable to leverage temporal relationships in sequential data.

% ~\wq{the order of the headers are different}

%%% outcome
% May 2023 newly update
    \begin{table}[h!]
        \centering
        \caption{\textit{Benchmarking performance on the task of early mortality prediction}. The reported score is of the form $mean \pm std$. `TA' denotes the model trained with the time-aware loss. \textbf{Bold} denotes the best performance. \underline{Underline} indicates that the model with time-aware loss outperforms the original model. The asterisk * denotes that the performance improvement against the model without TA version is statistically significant (p-value < 0.05). All three metrics are multiplied by 100 for readability purposes.}
        \label{tab:early_motality_benchmark}
    \begin{tabular}{l|ccc|ccc}
\toprule
Dataset & \multicolumn{3}{c}{TJH} & \multicolumn{3}{|c}{CDSL} \\
    \midrule
    Metric & AUPRC($\uparrow$) & AUROC($\uparrow$) & ES($\uparrow$) & AUPRC($\uparrow$) & AUROC($\uparrow$) & ES($\uparrow$) \\ 
\midrule
4C & 89.84 $\pm$ 4.46 & 94.16 $\pm$ 2.57 & - & 23.93 $\pm$ 2.99 & 76.15 $\pm$ 4.06 & - \\ 
RF & 95.56 $\pm$ 2.69 & 96.58 $\pm$ 2.20 & 72.63 $\pm$ 9.15 & 49.48 $\pm$ 3.79 & 84.35 $\pm$ 2.66 & -10.54 $\pm$ 3.57 \\ 
DT & 80.48 $\pm$ 7.26 & 87.41 $\pm$ 3.99 & 67.86 $\pm$ 11.24 & 38.27 $\pm$ 5.21 & 79.67 $\pm$ 4.61 & -8.23 $\pm$ 4.17 \\ 
GBDT & 95.13 $\pm$ 3.51 & 96.41 $\pm$ 2.35 & 76.45 $\pm$ 7.13 & 50.32 $\pm$ 5.15 & 85.15 $\pm$ 2.83 & 2.25 $\pm$ 4.86 \\ 
CatBoost & 95.99 $\pm$ 2.61 & 97.14 $\pm$ 1.81 & 74.91 $\pm$ 9.00 & 50.86 $\pm$ 4.34 & 85.09 $\pm$ 2.86 & -3.94 $\pm$ 3.89 \\ 
XGBoost & 95.70 $\pm$ 2.98 & 96.84 $\pm$ 2.09 & 76.41 $\pm$ 9.90 & 49.70 $\pm$ 5.06 & 84.59 $\pm$ 3.09 & -3.12 $\pm$ 4.44 \\ 
\midrule
\midrule
MLP & 93.78 $\pm$ 2.80 & 95.95 $\pm$ 1.62 & 72.87 $\pm$ 8.22 & 48.60 $\pm$ 3.57 & 84.15 $\pm$ 2.72 & -0.76 $\pm$ 3.95 \\ 
MLP-TA & 93.21 $\pm$ 2.78 & 95.65 $\pm$ 1.85 & \underline{73.94 $\pm$ 8.96} & \underline{48.67 $\pm$ 3.07} & \underline{84.24 $\pm$ 2.56} & \underline{0.55 $\pm$ 3.71} \\ 
RNN & 96.03 $\pm$ 3.42 & 97.41 $\pm$ 2.19 & 78.33 $\pm$ 11.35 & 57.57 $\pm$ 6.55 & 87.81 $\pm$ 2.55 & 19.66 $\pm$ 8.86 \\ 
RNN-TA & 95.97 $\pm$ 3.65 & 97.38 $\pm$ 2.29 & \underline{79.79 $\pm$ 11.06}* & \underline{58.21 $\pm$ 6.34}* & \underline{88.01 $\pm$ 2.36}* & \underline{20.15 $\pm$ 11.25}* \\ 
LSTM & 94.82 $\pm$ 7.65 & 97.08 $\pm$ 3.65 & 84.80 $\pm$ 10.68 & 55.53 $\pm$ 8.54 & 87.02 $\pm$ 3.01 & 19.63 $\pm$ 11.41 \\ 
LSTM-TA & \underline{95.40 $\pm$ 5.53}* & \underline{97.10 $\pm$ 3.05}* & 83.78 $\pm$ 8.30 & \underline{56.89 $\pm$ 7.34} & \underline{87.76 $\pm$ 2.50} & \underline{20.92 $\pm$ 13.10} \\ 
GRU & 96.33 $\pm$ 3.18 & 97.59 $\pm$ 2.13 & 78.33 $\pm$ 15.45 & 56.78 $\pm$ 8.02 & 87.93 $\pm$ 2.89 & 21.66 $\pm$ 8.86 \\ 
GRU-TA & \underline{96.50 $\pm$ 3.04} & \underline{97.70 $\pm$ 2.06} & \underline{80.93 $\pm$ 13.84}* & \underline{57.75 $\pm$ 5.36}* & \underline{87.98 $\pm$ 2.62}* & \underline{23.54 $\pm$ 8.10}* \\ 
TCN & 93.13 $\pm$ 5.10 & 96.39 $\pm$ 1.89 & 74.66 $\pm$ 13.03 & 57.09 $\pm$ 5.84 & 87.40 $\pm$ 2.67 & 13.48 $\pm$ 12.04 \\ 
TCN-TA & \underline{93.31 $\pm$ 4.97} & \underline{96.74 $\pm$ 1.87}* & \underline{76.47 $\pm$ 12.27} & \underline{57.71 $\pm$ 5.72} & \underline{87.59 $\pm$ 2.60} & \underline{21.59 $\pm$ 12.68}* \\ 
Transformer & 93.47 $\pm$ 6.73 & 96.86 $\pm$ 2.13 & 79.21 $\pm$ 13.05 & 38.34 $\pm$ 5.07 & 81.32 $\pm$ 3.46 & -3.96 $\pm$ 13.23 \\ 
Transformer-TA & \underline{93.86 $\pm$ 6.42} & \underline{97.01 $\pm$ 2.01} & \underline{80.73 $\pm$ 12.47}* & \underline{40.18 $\pm$ 5.00} & \underline{82.36 $\pm$ 3.61}* & \underline{-0.78 $\pm$ 11.42}* \\ 
\midrule
\midrule
RETAIN & 96.49 $\pm$ 2.05 & 97.84 $\pm$ 1.44 & 82.59 $\pm$ 12.29 & 54.54 $\pm$ 7.97 & 85.02 $\pm$ 3.93 & 7.32 $\pm$ 10.67 \\ 
RETAIN-TA & 95.99 $\pm$ 2.42 & 97.82 $\pm$ 1.51 & \underline{82.85 $\pm$ 11.36}* & 54.30 $\pm$ 6.50 & \underline{85.43 $\pm$ 2.81} & \underline{9.00 $\pm$ 7.66} \\ 
StageNet & 95.71 $\pm$ 3.77 & 97.27 $\pm$ 2.38 & 77.44 $\pm$ 12.52 & 56.57 $\pm$ 6.82 & 87.09 $\pm$ 3.54 & 23.93 $\pm$ 9.26 \\ 
StageNet-TA & \underline{95.79 $\pm$ 3.86}* & \underline{97.32 $\pm$ 2.41}* & \underline{78.88 $\pm$ 12.29}* & \underline{58.19 $\pm$ 6.48}* & \textbf{\underline{88.18 $\pm$ 2.82}}* & \textbf{\underline{24.55 $\pm$ 9.32}} \\ 
Dr.Agent & 97.22 $\pm$ 2.44 & 98.00 $\pm$ 1.75 & 85.80 $\pm$ 7.97 & 54.17 $\pm$ 8.92 & 86.76 $\pm$ 3.64 & 17.53 $\pm$ 11.25 \\ 
Dr.Agent-TA & 97.16 $\pm$ 2.56 & 98.00 $\pm$ 1.97 & \textbf{\underline{86.01 $\pm$ 7.66}} & 53.06 $\pm$ 7.34 & \underline{87.06 $\pm$ 3.30}* & \underline{19.35 $\pm$ 9.60}* \\ 
AdaCare & 97.86 $\pm$ 1.09 & 98.53 $\pm$ 0.77 & 78.39 $\pm$ 7.51 & 55.32 $\pm$ 4.45 & 86.59 $\pm$ 1.99 & 17.46 $\pm$ 8.66 \\ 
AdaCare-TA & \textbf{\underline{98.11 $\pm$ 1.13}} & \textbf{\underline{98.64 $\pm$ 0.89}} & \underline{84.51 $\pm$ 8.17}* & \underline{55.62 $\pm$ 5.44} & \underline{87.00 $\pm$ 2.18} & \underline{20.09 $\pm$ 9.86}* \\ 
GRASP & 96.04 $\pm$ 3.15 & 97.30 $\pm$ 2.30 & 81.63 $\pm$ 9.71 & 53.95 $\pm$ 8.37 & 84.44 $\pm$ 4.40 & 15.27 $\pm$ 15.17 \\ 
GRASP-TA & \underline{96.16 $\pm$ 3.17} & \underline{97.40 $\pm$ 2.21} & \underline{82.82 $\pm$ 9.08} & 53.50 $\pm$ 7.89 & \underline{84.91 $\pm$ 3.73}* & \underline{15.61 $\pm$ 13.76}* \\ 
ConCare & 97.01 $\pm$ 2.46 & 97.89 $\pm$ 1.62 & 80.97 $\pm$ 13.66 & 58.59 $\pm$ 6.49 & 87.76 $\pm$ 3.02 & 17.58 $\pm$ 12.34 \\ 
ConCare-TA & \underline{97.04 $\pm$ 2.51} & \underline{97.94 $\pm$ 1.62} & \underline{82.31 $\pm$ 13.08}* & \textbf{\underline{58.68 $\pm$ 7.04}} & \underline{87.96 $\pm$ 3.23} & \underline{20.90 $\pm$ 10.43} \\ 
\bottomrule
\end{tabular}
\end{table}

\section{Discussions}
In this section, we further explore models' performances in terms of the proposed metrics ES and OSMAE. These metrics show models' performances on the \textit{early} and \textit{outcome-specifc} perspectives of two tasks. We also discuss the limitations and future directions of this work.

\subsection{Analysis of early prediction performance}
\begin{figure}[htbp]
\centering
% \begin{subfigure}{0.45\textwidth}
%   \centering
%   \includegraphics[width=\textwidth]{figures/compare_ta_loss_auroc_score_simulated.pdf} % first figure itself
%   \caption{AUROC of early prediction}
%   \label{fig:compare_ta_loss_auroc_score_simulated}
% \end{subfigure}%
% \begin{subfigure}{0.45\textwidth}
  % \centering
  \includegraphics[width=0.8\textwidth]{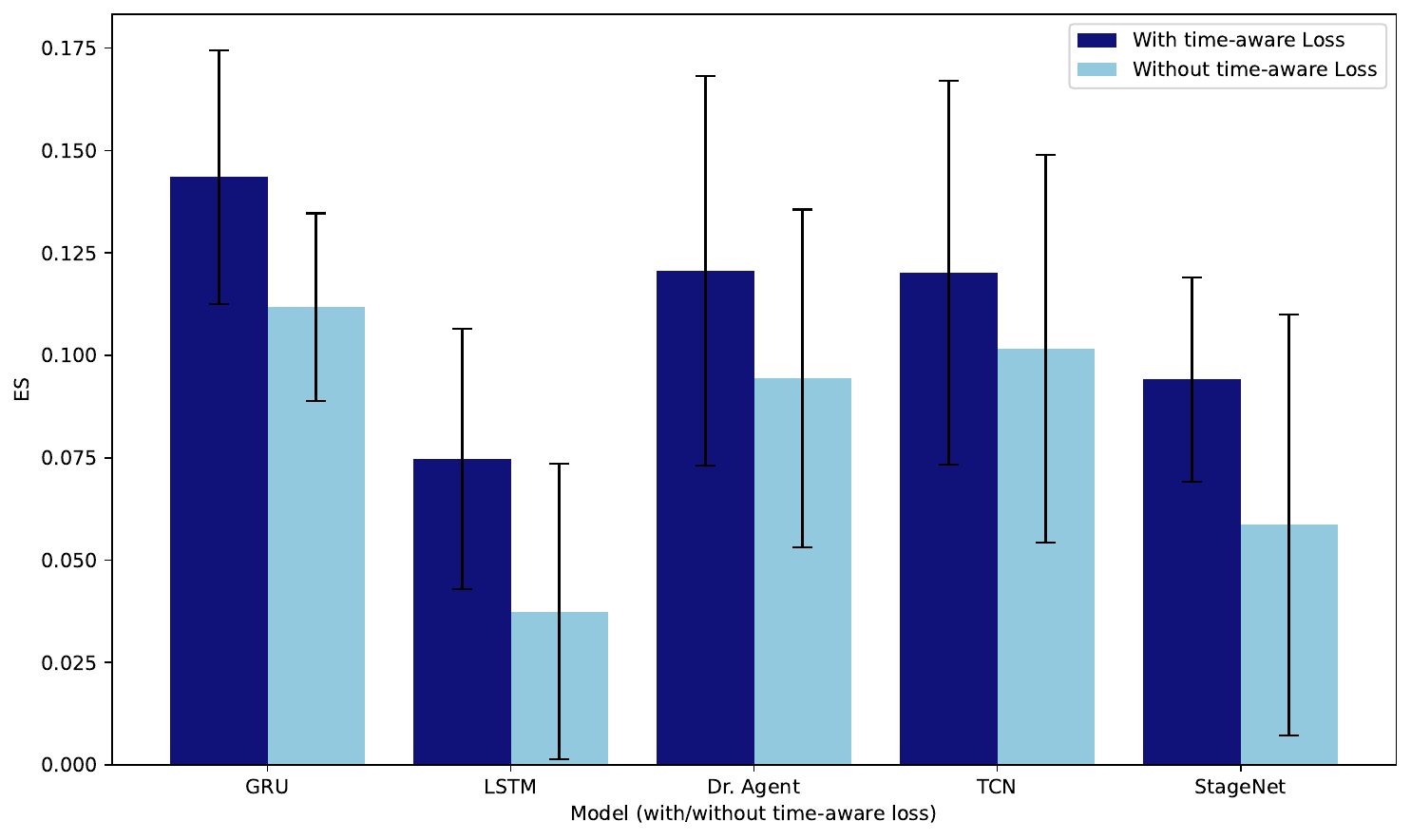} % second figure itself
  % \caption{ES of early prediction}
  \label{fig:compare_ta_loss_es_score_simulated}
% \end{subfigure}
\caption{\textit{Early prediction performance of 5 models with the highest ES on the CDSL dataset.} All models are trained using the first half of patient records. Error bars are standard deviations. All performance improvements are statistically significant (p-value < 0.05).}
\label{fig:compare_es_score_simulated_cdsl}
\end{figure}

By enhancing early prediction performance, models can discern patients' health risks in the early stages. This can address the `cold start' issue of predictive models, which is particularly beneficial during the early phases of a pandemic when predictions must rely on sparse data. To assess the efficacy of our proposed early prediction loss term in improving early-stage predictions, we simulated a scenario where the model generates forecasts using only the first half of a patient's visit records. The ES scores for the top five best-performing models (both with and without the time-aware loss) on the CDSL dataset are illustrated in Figure~\ref{fig:compare_es_score_simulated_cdsl}. The findings demonstrate that models incorporating the time-aware loss consistently achieve superior ES scores in the early stages, which directly assesses the true positive rate of predictions. This underscores the adaptability of our proposed time-aware loss in enhancing various models' early prediction capabilities. 

\begin{figure}[htb!]
    \centering
    \includegraphics[width=0.5\textwidth]{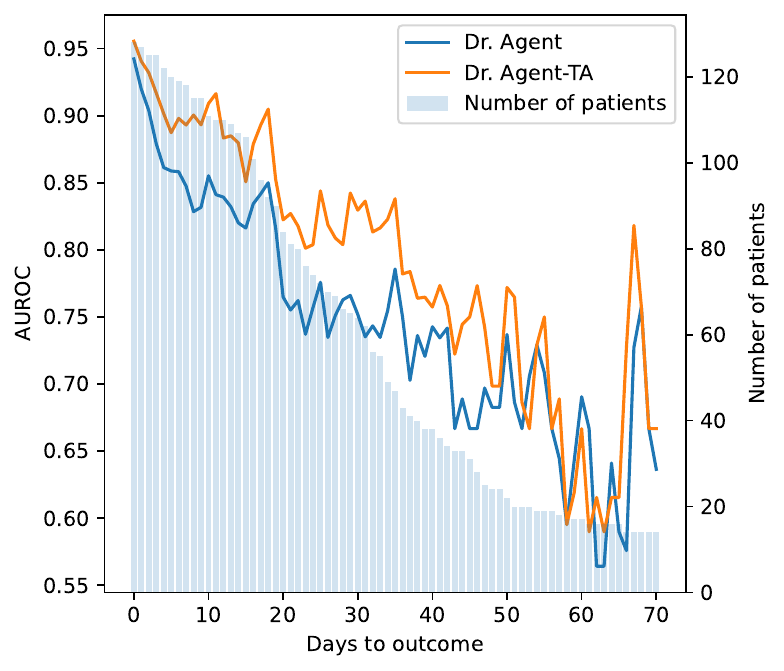}
    \caption{\textit{AUROC of Dr. Agent and Dr. Agent-TA at each visit}}
    \label{fig:agent_es}
\end{figure}

To offer a more granular view of the performance gains attributed to the time-aware loss, we've charted the AUROC at different visits (days to outcome) for the Dr. Agent model on the CDSL dataset as a case study in Figure~\ref{fig:agent_es}. The figure shows that most patients' LOS are less than 30 days, and the AUROC of the model with the time-aware loss is significantly higher. This suggests that, for most patients, the time-aware loss is more effective in identifying early health risks.

% \begin{figure}
%      \centering
%      \begin{subfigure}[h]{0.45\textwidth}
%          \centering
%          \includegraphics[width=\textwidth]{figures/auroc_model.pdf}
%          \caption{AUROC of early prediction}
%      \end{subfigure}
%      \begin{subfigure}[h]{0.45\textwidth}
%          \centering
%          \includegraphics[width=\textwidth]{figures/es_model.pdf}
%          \caption{ES of early prediction}
%      \end{subfigure}
% \caption{Early prediction performance of top 5 models on the CDSL dataset by using just the first half of patient records.}
% \label{fig:epf}
% \end{figure}

\subsection{Case Study of Prediction Metrics}

\begin{figure}[htbp]
\centering
\begin{subfigure}{0.45\linewidth}
  \centering
  \includegraphics[width=\textwidth]{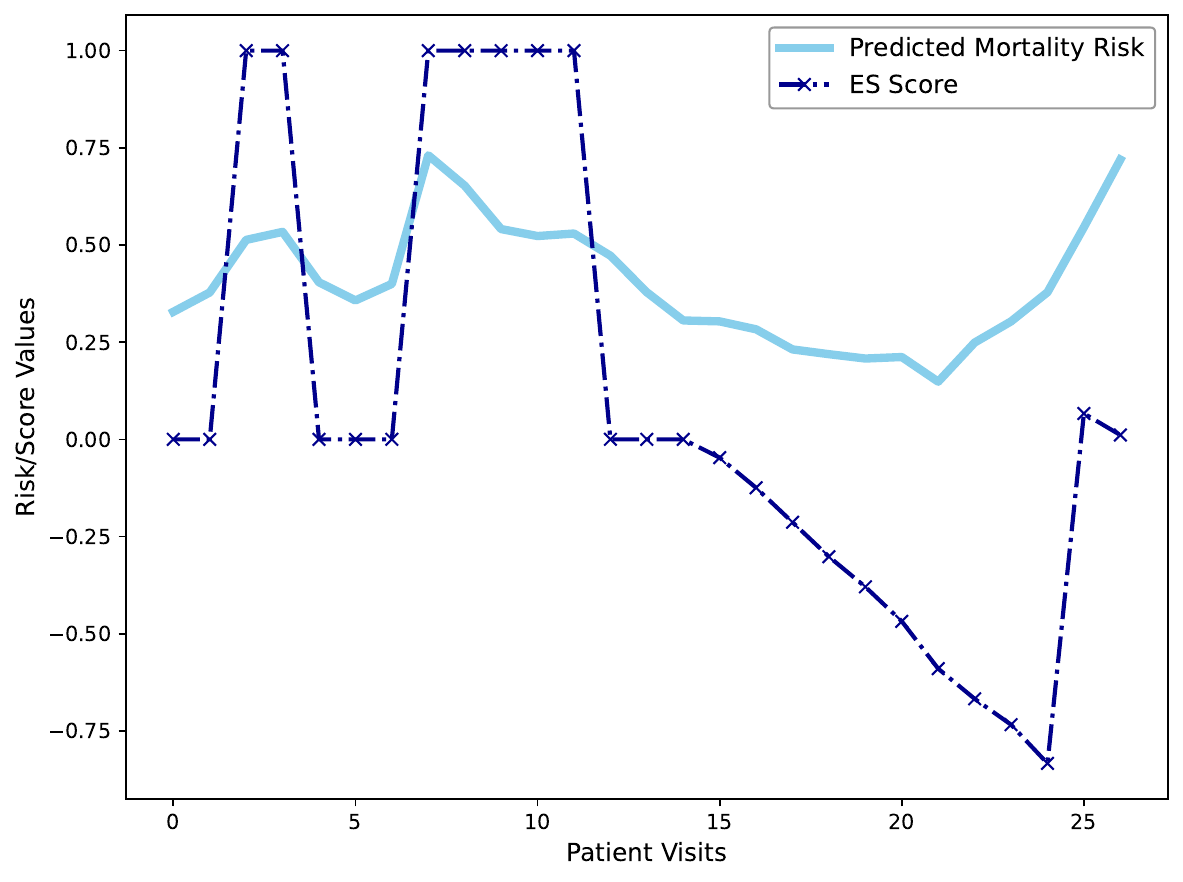} % first figure itself
  \caption{ES and risk curve}
  \label{fig:case_study_es_pid407}
\end{subfigure}%
\begin{subfigure}{0.45\linewidth}
  \centering
  \includegraphics[width=\textwidth]{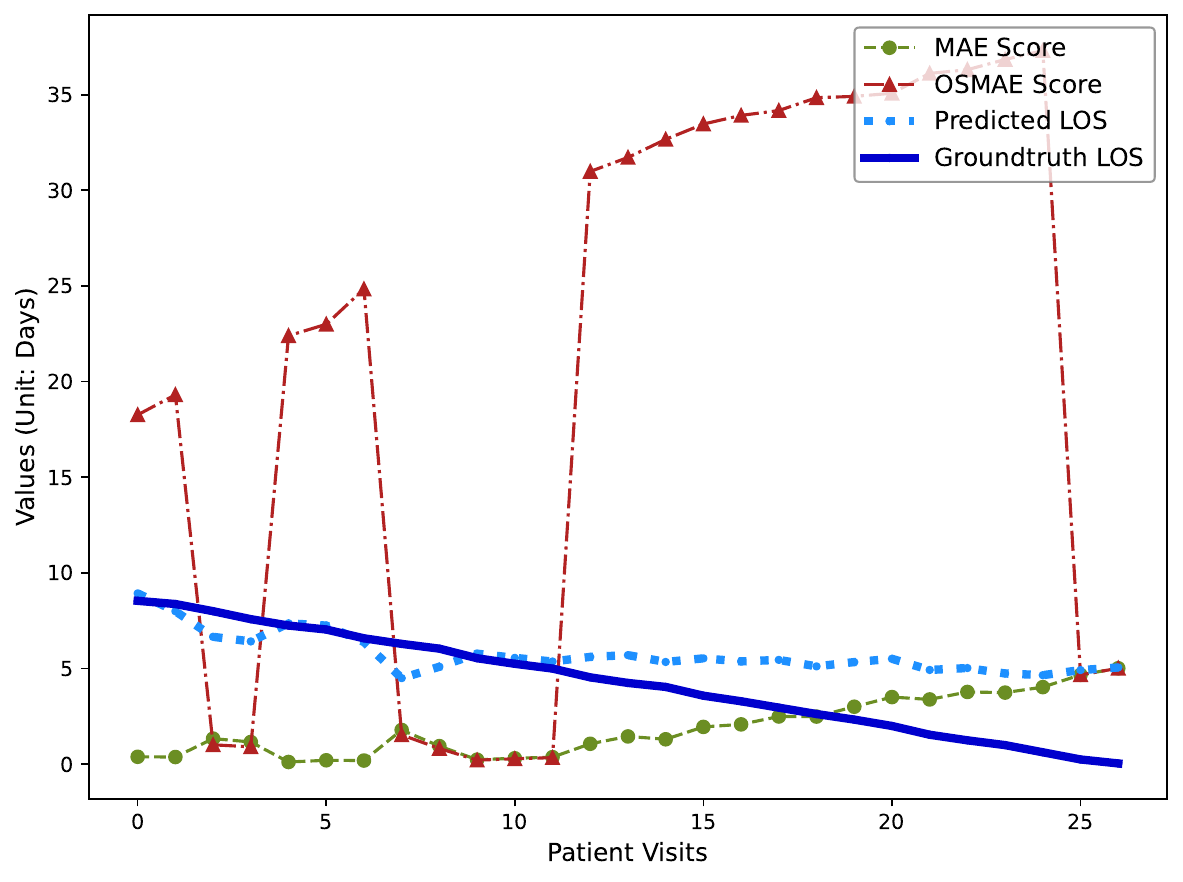} % second figure itself
  \caption{OSMAE, MAE, predicted LOS and ground truth LOS curve}
  \label{fig:case_study_los_pid407}
\end{subfigure}
\caption{\textit{Case prediction plot on the CDSL dataset.} We use the TCN model with the multi-task setting to generate the prediction results.}
\label{fig:case2_cdsl}
\end{figure}

The proposed ES and OSMAE metrics are inherently interpretable as they are based solely on straightforward score calculations, independent of the models. This makes the calculation process transparent and easy to understand. To further elucidate our metrics, we have included a case study here, illustrated in Figure~\ref{fig:case2_cdsl}.

In the left figure, we plot the predicted risk probability over time alongside the ES score at each timestep for a patient whose actual outcome was mortality. Initially, the model accurately identifies early risk by predicting a risk probability above 0.5, earning a full ES score (1.0) for these early correct predictions. However, over time, as the model incorrectly assesses the patient's condition as improving, it receives progressively lower ES scores. This decline reflects the model's erroneous predictions as the patient approaches the final outcome. Near the end, even though the model predicts high risks again, the rewards for these correct predictions are lower due to their proximity to the final outcome.

In the right figure, we show the predicted Length of Stay (LOS), the actual LOS, the MAE, and the OSMAE for the same patient. At first glance, the predicted LOS (dotted blue line) closely matches the actual LOS (solid blue line), resulting in a consistently low MAE (dotted green line). However, the higher OSMAE (red line) indicates that the model is making incorrect outcome predictions, suggesting a recovery when the patient's condition is actually deteriorating. When outcomes are predicted accurately, OSMAE aligns with MAE. Using traditional MAE as a sole metric could lead to biased decisions, as it may not reflect the patient's actual health trajectory.

This case study demonstrates that our proposed ES and OSMAE metrics can more accurately represent a model's ability to assess early patient risk and understand disease progression, aspects often overlooked in traditional classification and regression metrics.

\subsection{Outcome-specific prediction performance}
Another primary objective of our study is to determine how the proposed outcome-specific LOS prediction task can enhance clinical decision-making. In Table~\ref{tab:dis_cdsl}, we display the prediction discrepancies for the top 5 models with the highest $\Delta$MAE (OSMAE-MAE). To demonstrate how $\Delta$MAE can retain information about the correctness of outcome predictions while computing the MAE, we also include the number and percentage of patients with incorrect outcome predictions. 

\begin{table}[htb]
\centering
\caption{\textit{Prediction discrepancy of 5 models on the CDSL dataset.}}
\begin{tabular}{lllll}
\toprule
Model       & Type      & $\Delta$MAE & \specialcell{Outcome Wrong \#} & \specialcell{Outcome Wrong \%} \\
\midrule
Transformer & Two-stage  & 1.21      & 1469           & 11.89\%\\ %
Transformer & Multi-task & 1.13      & 1389           & 11.24\%\\ %  
MLP         & Multi-task & 0.78      & 1322           & 10.70\%\\ %  
MLP         & Two-stage  & 0.75      & 1295           & 10.48\%\\ %  
Dr.Agent     & Multi-task & 0.67      & 1108           & 8.97\%\\ %  
\bottomrule
\end{tabular}
\label{tab:dis_cdsl}

\end{table}

The results indicate that as $\Delta$MAE increases, the percentage of incorrect outcome predictions also rises. This suggests that an increase in outcome prediction error contributes to the discrepancy between our new OSMAE metric and the traditional MAE. This conclusion further implies that in traditional Length of Stay (LOS) prediction tasks, relying solely on MAE to assess prediction accuracy can lead to erroneous conclusions and potentially compromise clinical decision-making. Consequently, refining the OSMAE metric—rather than solely concentrating on the conventional MAE—is crucial for future predictive modeling endeavors, possibly through a more judicious multi-task learning framework or regularizer.
% \begin{table}[]
% \caption{Prediction discrepancy of 6 models on the TJH dataset.}
% \begin{tabular}{lllllll}
% \toprule
% Model       & Type      & $\Delta$MAE & \specialcell{Both \\Correct \#} & \specialcell{Both \\Correct \%} & \specialcell{Outcome \\Correct \#} & \specialcell{Outcome \\Correct \%} \\
% \midrule
% StageNet    & Multitask & 2.61 & 58 & 33.53\% & 87  & 50.29\% \\
% MLP         & Multitask & 2.36 & 30 & 17.34\% & 79  & 45.66\% \\
% Transformer & Multitask & 1.55 & 49 & 28.32\% & 101 & 58.38\% \\
% MLP         & TwoStage  & 1.26 & 26 & 15.03\% & 93  & 53.76\% \\
% LSTM        & Multitask & 1.07 & 70 & 40.46\% & 85  & 49.13\% \\
% StageNet    & TwoStage  & 0.99 & 31 & 17.92\% & 95  & 54.91\% \\
% \bottomrule
% \end{tabular}
% \end{table}

\subsection{Sensitivity analysis of OSMAE and ES}
\begin{figure}[h!]
\centering
\begin{subfigure}{.5\textwidth}
  \centering
  \includegraphics[width=.9\linewidth]{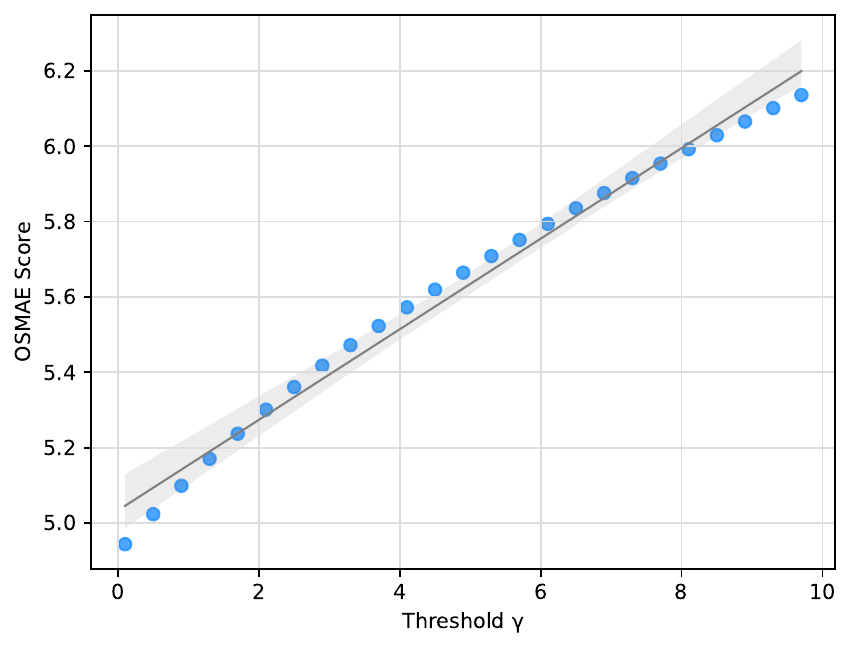}
  \caption{OSMAE with $\gamma$ from 0-10}
  \label{fig:osmae_trend}
\end{subfigure}%
\begin{subfigure}{.5\textwidth}
  \centering
  \includegraphics[width=.9\linewidth]{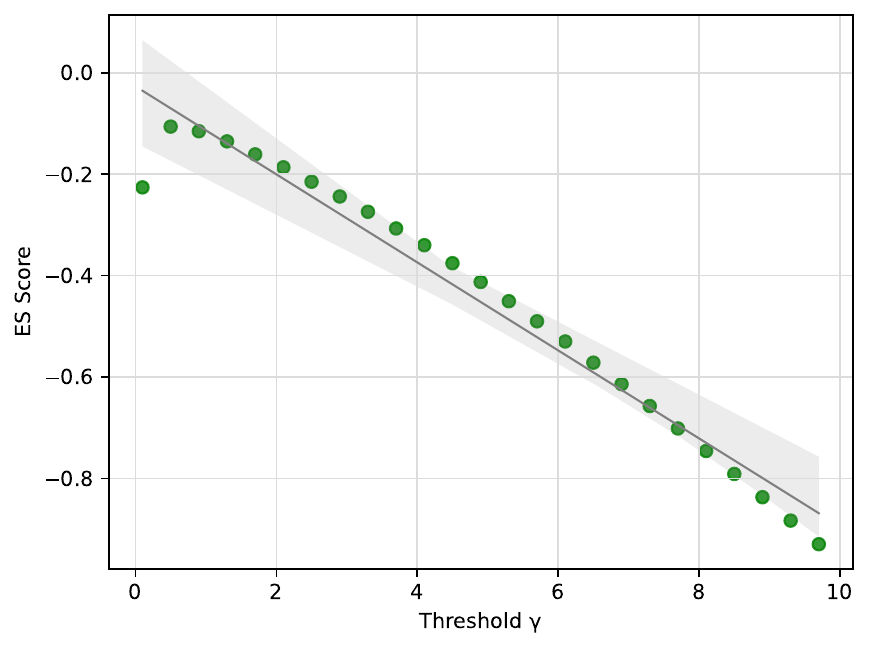}
  \caption{\textit{ES} with $\gamma$ from 0-10}
  \label{fig:emp_trend}
\end{subfigure}
\caption{\textit{OSMAE and ES values with different $\gamma$.} Scores are calculated using the same TCN multi-task model on \textcolor{red}{CDSL} dataset.}
\label{fig:metrics_trend}
\end{figure}

We investigate the influence of the threshold parameter, $\gamma$, on the \textit{OSMAE} and the \textit{ES}. The \textit{OSMAE} and \textit{ES} values for various $\gamma$ thresholds, using the same GRU multi-task model on the \textcolor{red}{CDSL} dataset, are plotted in Figure~\ref{fig:metrics_trend}. Evidently, as $\gamma$ increases, \textit{OSMAE} increases and \textit{ES} decreases. This occurs because a larger $\gamma$ necessitates the model to accurately predict the target earlier, thereby increasing the task's difficulty. In practical clinical settings, clinicians can choose an appropriate threshold based on their specific requirements. Another advantage of this $\gamma$ setting is making both metrics invariant to the record length, which means patients with too many visits bring less bias to the proposed metrics compared to traditional AUROC and AUPRC. This is because both metrics are normalized by the visit time, ensuring that the length of a patient's record does not disproportionately influence the metric values, regardless of the duration of the patient record. No matter how long the patient record is, the visits before the penalty time $\gamma$ are regularized to have less effect on the metric values.

\begin{figure}[htbp]
\centering
\begin{subfigure}{0.45\textwidth}
  \centering
  \includegraphics[width=\textwidth]{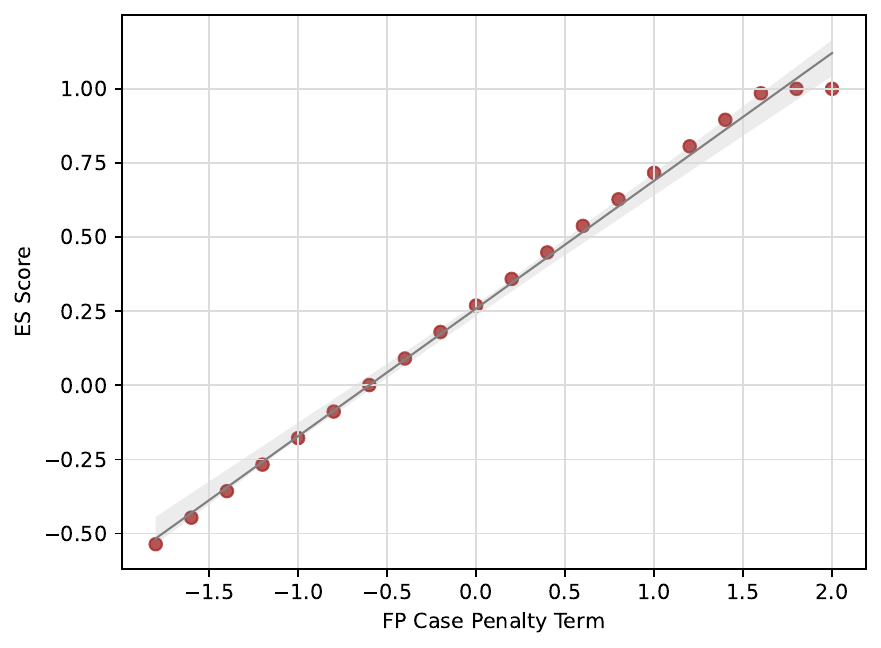} % first figure itself
  \caption{FP case penalty term}
  \label{fig:es_with_penalty_term_cdsl}
\end{subfigure}%
\begin{subfigure}{0.45\textwidth}
  \centering
  \includegraphics[width=\textwidth]{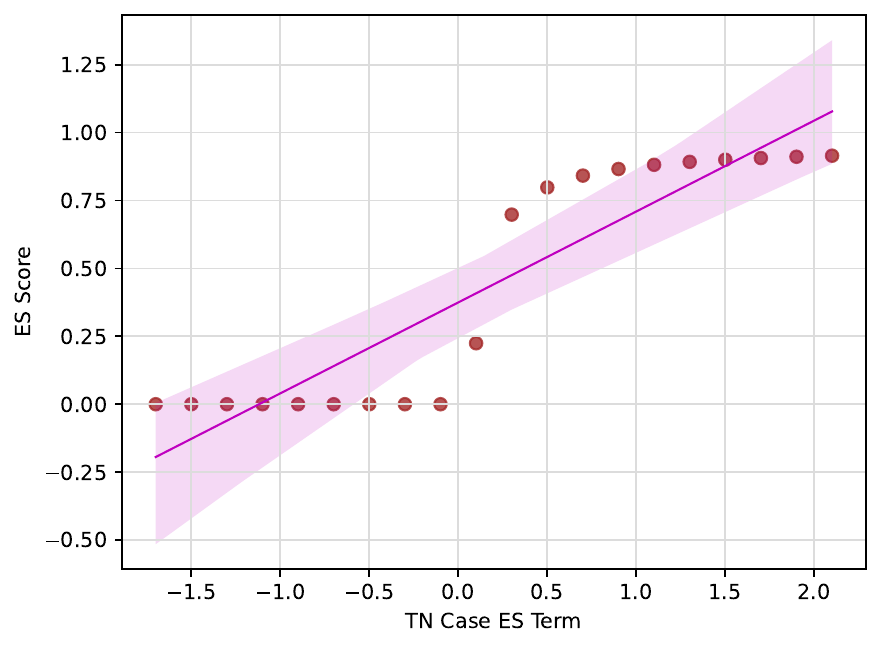} % second figure itself
  \caption{TN case penalty term}
  \label{fig:es_with_tn_term_cdsl}
\end{subfigure}
\caption{\textit{ES values with different penalty terms for TN case and FP case ranging from $-2$ to $2$.} Scores are calculated using the same TCN multi-task model on the CDSL dataset.}
\label{fig:es_score_observation_tn_fn}
\end{figure}

We also aim to explore the effect of penalty score value in ES for FN and FP cases. We selected -0.1 as the penalty score for false positive (FP) cases to mirror scenarios in real-world clinical settings, where the costs associated with false negative predictions are typically higher than those for false positives, particularly in intensive care environments~\cite{mallett2012interpreting, clifford2015reducing}. Thus, a lower penalty score is assigned to false positive predictions. However, clinicians may adjust this penalty score to suit their specific needs in various applications. For example, in a less critical clinical setting, clinicians may set higher penalty scores to select the model that produces fewer false alarms. Since the Effectiveness Score (ES) solely functions as a metric, it does not influence model performance. Figure \ref{fig:es_score_observation_tn_fn} illustrates how the penalty score impacts the ES value. As demonstrated in the figure, while the penalty score influences the absolute value of ES, it does not alter the underlying trend. True negative cases receive a score of zero, aligning with the ES metric's aim to evaluate models for early risk detection in patients. Since true negative predictions do not trigger alarms, the model is neither rewarded nor penalized for these outcomes. It is important to note that the number of true positive predictions is 649, compared to 10,381 true negative predictions. Awarding points for true negative predictions would introduce significant bias into the final score.

\subsection{Model interpretability, limitations and future works}
Interpretability is an important aspect of models, especially in clinical settings. With model interpretability, the clinicians can understand the model's prediction process and make reliable clinical decisions. Many models benchmarked in this work claim they have different levels of interpretability. We group them into four groups: 1. visit-level; 2. feature-level; 3. others. Visit-level interpretability indicates that the model can identify the most important visit from longitudinal EHR sequences and feature-level interpretability means that the model can identify the most important risk factors from a single visit data. Some models can provide other information to aid clinical decisions such as disease progression score. We summarize the characteristics of all interpretable models in Table~\ref{tab:interpret}. 

\begin{table}[htb]
\centering
\caption{Interpretability characteristics of all interpretable models}
\begin{tabular}{llll}
\toprule
Model       & Visit-level      & Feature-level & Others \\
\midrule
4C & - & \checkmark & - \\
RF & - & \checkmark & - \\
DT & - & \checkmark & - \\
GBDT & - & \checkmark & - \\
CatBoost & - & \checkmark & - \\
XGBoost & - & \checkmark & - \\
\midrule
RETAIN & \checkmark & \checkmark & - \\
StageNet & - & - & disease progression score \\
Dr. Agent & \checkmark & - & - \\
AdaCare & - & \checkmark & - \\
GRASP & - & - & patient clusters \\
ConCare & - & \checkmark & feature correlations \\
\bottomrule
\end{tabular}
\label{tab:interpret}

\end{table}

We find most machine learning models are interpretable as they are tree-based models. These models can only provide feature-level interpretability (i.e. feature importance score) because they cannot process temporal dependencies. All naive deep learning models are black-box models and they are not interpretable. However, all EHR-specific deep learning models studied in this work are designed to provide some interpretability. RETAIN can provide both visit-level and feature-level interpretabilities. StageNet, GRASP and ConCare can provide additional information including disease progression score, patient clustering information and feature correlations.

We aim to establish a comprehensive benchmark for COVID-19 predictive modeling. However, our study has certain limitations. Firstly, within our two-stage approach, the models predicting mortality and Length of Stay (LOS) might have distinct architectures. For comparison purposes, due to the inherent complexity, we only assessed scenarios where both models are identical. Additionally, the multi-task setting may benefit from more intricate multi-task learning structures. Exploring various model combinations and learning structures is a promising avenue for future work. Secondly, during data preprocessing, we imputed missing values using the most recent measurement. While this method has been adopted in numerous studies~\cite{harutyunyan2019multitask,ma2020adacare,ma2020concare}, it is essential to investigate how different imputation techniques influence the outcomes in a clinical context. Lastly, when employing the OSMAE metric, other consequential clinical outcomes, like re-admission, besides mortality and recovery, should be considered. Future studies should account for these competing factors by devising more inclusive tasks and metrics. Our proposed tasks align with clinical practice requirements. Beyond task and metric design, the introduced early prediction loss and multi-task learning framework are innovative approaches yielding encouraging results for the tasks presented. We anticipate further refinements in future research endeavors. 

We envision that our proposed benchmark analysis pipeline and coding framework can be extended to a wider range of clinical tasks and datasets. In this work, to bolster the comprehensiveness and validity of our evaluation, we incorporated an expansive selection of ICU datasets, namely MIMIC-III~\cite{johnson2016mimic} and MIMIC-IV~\cite{johnson2020mimic}. We undertook the in-hospital mortality prediction task using these two datasets, with results detailed in Appendix E. While our efforts remain exploratory in nature, we aspire for our proposed tasks to be integrated across diverse clinical datasets, thereby enhancing the quality of healthcare delivery. Another line for future research pertains to benchmarking the fairness of model predictions. We have undertaken preliminary experiments using the GRU model with four fairness metrics, with results detailed in Appendix C. Subsequent efforts might focus on a thorough analysis aimed at minimizing bias and enhancing the fairness across various predictive models.

\section{Conclusions}
This paper presents a fair, comprehensive, and open-source benchmark for predictive modeling of COVID-19, using Electronic Health Records (EHR) data from ICUs. We introduce two innovative clinical prediction tasks – early mortality prediction and outcome-specific length-of-stay prediction – based on the clinical practice of COVID-19 prediction scenarios on two real-world datasets. Our approach ensures fairness and reproducibility through the implementation of meticulous data processing and model training pipelines. We provide benchmarking performances for a total of 17 predictive models, including 5 traditional machine learning models, 6 basic deep learning models, and 6 state-of-the-art EHR-specific deep learning models. To improve accessibility, we have made our benchmarking results and trained models readily available online. This not only facilitates quick access to prediction results using our trained models, but also grants clinicians and researchers the ability to obtain swift prediction results using our trained models. Our endeavors aim to spur continuous development and advancement in the field of deep learning and machine learning, particularly in the context of pandemic predictive modeling.
\section{Data and Code Availability}

This research did not involve the collection of new patient EHR data. The \textcolor{blue}{TJH} EHR dataset~\cite{yan2020interpretable} utilized in this study is publicly available on GitHub (\url{https://github.com/HAIRLAB/Pre_Surv_COVID_19}). The \textcolor{red}{CDSL} dataset~\cite{hmh} is open to global researchers and can be accessed on request (\url{https://hmhospitales.com/prensa/notas-de-prensa/comunicado-covid-data-save-lives}). To gain access, applicants should fill out the request form. We use these datasets under their respective licenses.

\begin{figure}[h!]
        \centering
        \includegraphics[width=\textwidth]{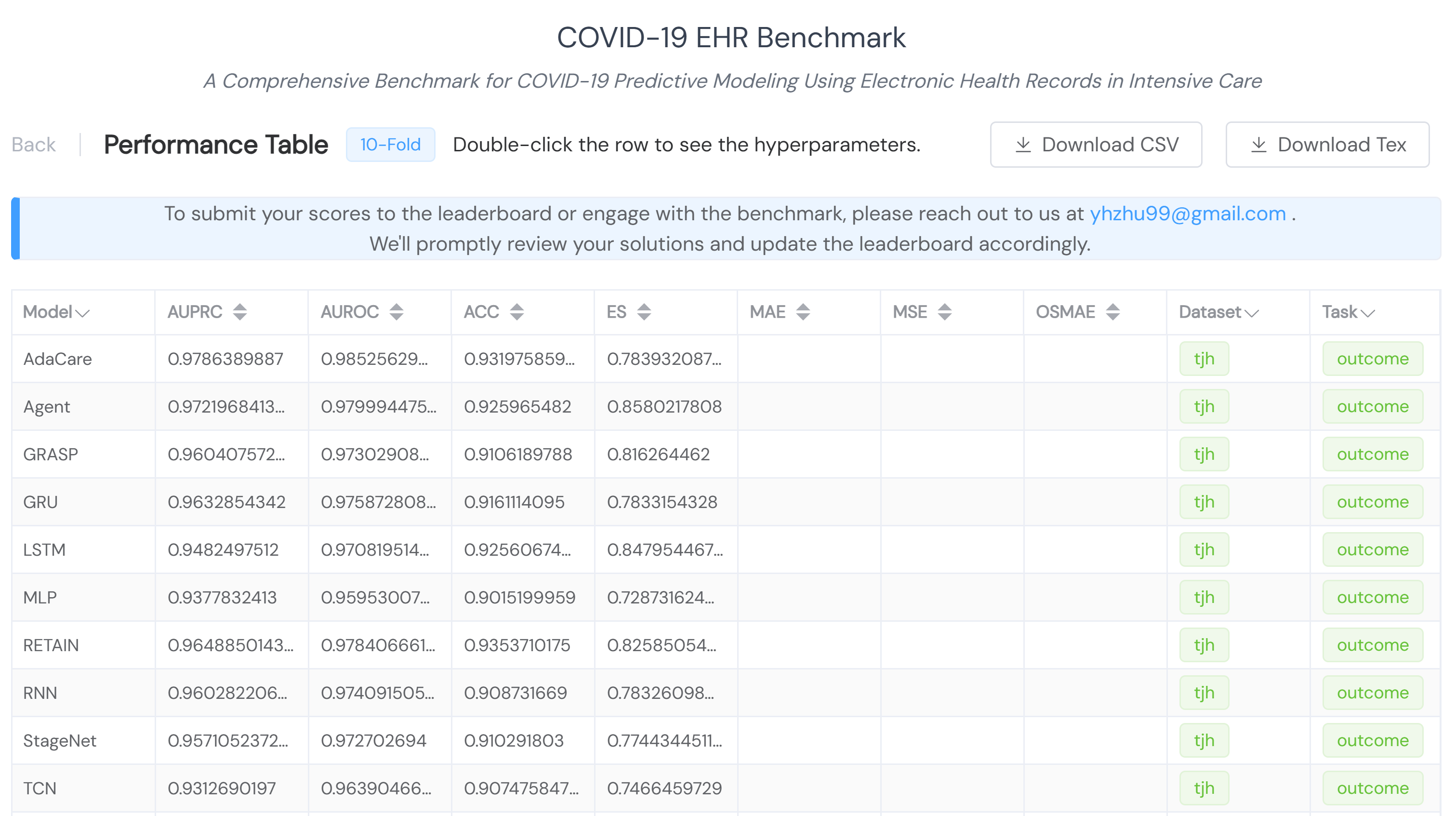}
        \caption{Screenshot of the online benchmark results visualization system.}
        \label{fig:screenshots}
\end{figure}

Furthermore, we have made our benchmarking system available online at \url{https://pyehr.netlify.app}, as illustrated in Figure~\ref{fig:screenshots}. All model performances with their associated hyperparameter combinations on both tasks are accessible in the system. These results were used to generate all the performance analysis figures presented in this work. The tables are designed for easy querying, comparison, and sorting. Users can select specific rows and download the CSV files or LaTeX codes. Additionally, all trained models are also released online, enabling clinicians and researchers to conveniently use these models to obtain prediction results with their own test samples. We have also made the source code of this online system publicly available at \url{https://github.com/yhzhu99/pyehr-playground} so that researchers can easily deploy an offline version.

\begin{figure}[h!]
    \centering
    \includegraphics[width=1.0\linewidth]{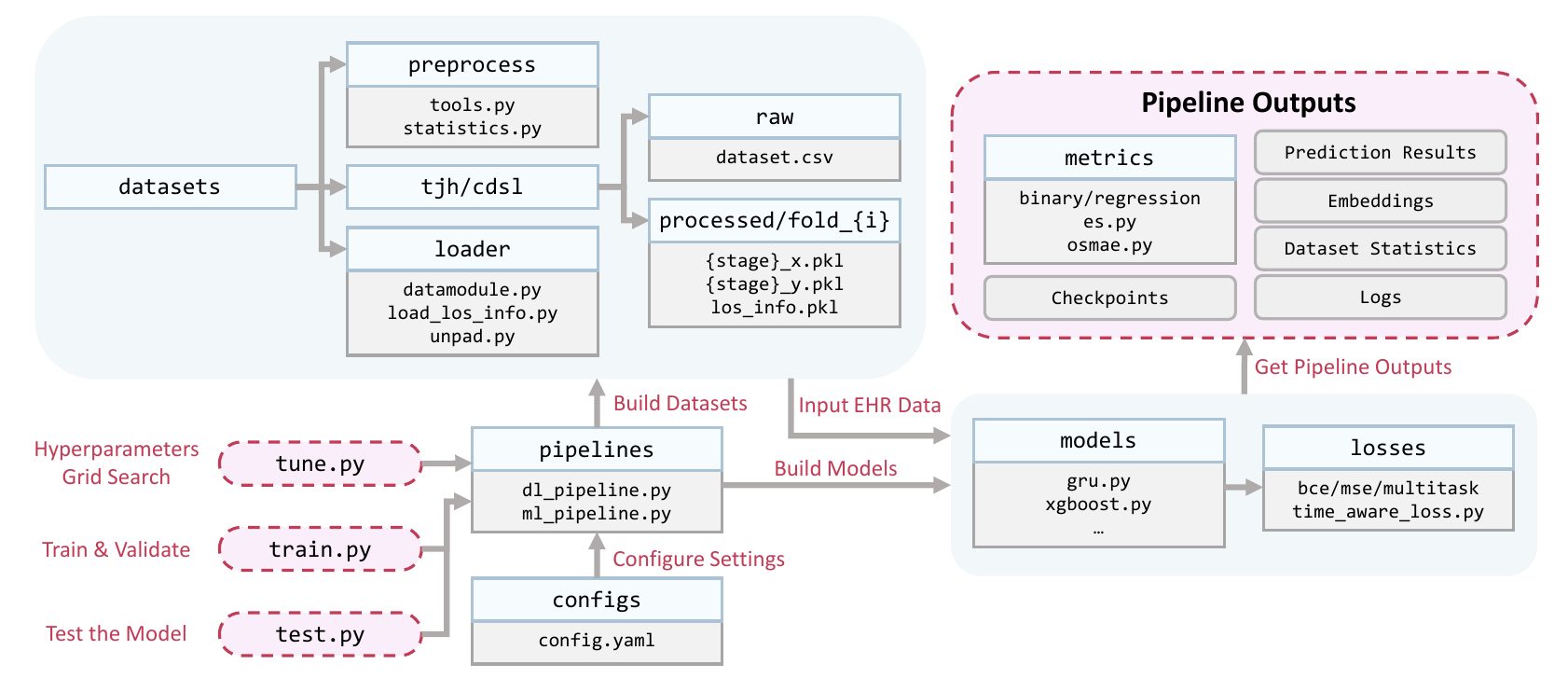}
    \caption{Benchmark code structure.}
    \label{fig:toolkit_tree}
\end{figure}

The code for this benchmarking work can be accessed on GitHub (\url{https://github.com/yhzhu99/pyehr}). The code structure of this benchmarking pipeline is depicted in Figure~\ref{fig:toolkit_tree}. The diagram is essentially self-explanatory. All runtime configurations for model parameters are stored in \verb|configs/| directory. All deep learning and machine learning models are implemented with various modules. This clearly structured and modular design allows users to effortlessly extend the existing models, design new models and tasks, or even introduce new datasets by adding corresponding components without impacting the downstream benchmark prediction and evaluation process. The analytical files generated in this work are also publicly available online. These data analysis files are housed in the \verb|datasets/| folder of the GitHub repository.

\section*{Acknowledgements}
This work was supported by the National Natural Science Foundation of China (No.82241052) , China Postdoctoral Science Foundation (2021TQ0011, 2022M720237).
Junyi Gao acknowledges the receipt of studentship awards from the Health Data Research UK-The Alan Turing Institute Wellcome PhD Programme in Health Data Science (Grant Ref: 218529/Z/19/Z).

\section*{Author Contributions Statement}

Junyi Gao and Liantao Ma designed this study, including tasks, pipelines, experiments and result analysis. Yinghao Zhu and Wenqing Wang processed the datasets and built the models. All authors wrote and reviewed the manuscript. 

\section*{Declaration of Interests}

All authors declare that they have no conflicts of interest.

\newpage
\bibliography{ref}

\appendix

% red line: dead
% green line: alive

% \section{T-SNE Visualizations of the Learned Embeddings}
% The hidden states learned by the Temporal Convolutional Network (TCN) model are visualized using t-SNE for the \textcolor{red}{CDSL} dataset, as depicted in Figure~\ref{fig:hidden_state_tsne}. Each green dot represents the embedding of a patient who survived, whereas a red cross signifies the embedding of a patient who died. We observe that the embeddings of patients with similar outcomes tend to cluster more closely in the TCN model trained with the multi-task strategy. Consequently, multi-task models achieve lower OSMAE scores.

% \begin{figure}[h!]
% \centering
% % \begin{subfigure}{.33\textwidth}
% %   \centering
% %   \includegraphics[width=.95\linewidth]{figures/outcome_tsne.pdf}
% %   \caption{Mortality outcome prediction}
% %   \label{fig:outcome_tsne}
% % \end{subfigure}%
% \begin{subfigure}{.4\textwidth}
%   \centering
%   \includegraphics[width=.95\linewidth]{figures/los_tsne.pdf}
%   \caption{Patient embeddings of the two-stage TCN}
%   \label{fig:los_tsne}
% \end{subfigure}
% \begin{subfigure}{.4\textwidth}
%   \centering
%   \includegraphics[width=.95\linewidth]{figures/multitask_tsne.pdf}
%   \caption{Patient embeddings of the multi-task TCN}
%   \label{fig:multitask_tsne}
% \end{subfigure}
% \caption{Patient embedding visualizations for two-stage and multi-task LOS prediction settings using t-SNE.}
% \label{fig:hidden_state_tsne}
% \end{figure}

\section{Baseline Models Descriptions}

\noindent\textbf{Clinical Scoring model}
\begin{itemize}[leftmargin=*]
\item \textbf{4C mortality score}~\cite{knight2020risk} is a risk stratification tool that predict in-hospital mortality or in-hospital clinical deterioration (defined as any requirement of ventilatory support or critical care, or death) for hospitalised COVID-19 patients. They are designed to require only parameters that are commonly available at hospital presentation.
\end{itemize}

\noindent\textbf{Machine learning models}

\begin{itemize}[leftmargin=*]

    \item \textbf{Decision tree (DT)} is a non-parametric supervised learning algorithm with a hierarchical tree structure. DT has been widely applied in many clinical predictive tasks, such as mortality prediction for peritoneal dialysis patients\cite{noh2020prediction}. %(Nature Scientific Reports, 2020).

    \item \textbf{Random forest (RF)} is a supervised ensemble learning method for classification, regression and other tasks. It constructs a multitude of decision trees during the training period. RF has been used to predict the patients' severity of the COVID-19 case and possible mortality outcome, recovery or death~\cite{iwendi2020covid}.

    \item \textbf{Gradient Boosting Decision Tree (GBDT)} is a widely-used algorithm for solving prediction problems in both classification and regression tasks. GBDT takes decision trees as the base learner and determines final prediction results based on a series of DTs' outputs. GBDT exhibits good performance in performing COVID-19 mortality prediction tasks with structured clinical data~\cite{li2021development}.

    \item \textbf{XGBoost}~\cite{chen2016xgboost} is a recursive tree-based supervised machine learning classifier. XGBoost has been used to predict the mortality for COVID-19 infected patients\cite{yan2020interpretable}. %(Nature Machine Intelligence, 2020).

    \item \textbf{CatBoost}~\cite{dorogush2018catboost} is also a gradient boost machine learning framework. It builds a symmetric tree and uses an ordered boosting strategy with a faster training and prediction speed compared to XGBoost. Catboost has been used to predict and detect the number of confirmed and death cases of COVID-19~\cite{kim2021covid}.
    
\end{itemize}

\noindent\textbf{Basic deep learning models}

\begin{itemize}[leftmargin=*]

    \item \textbf{Multi-layer perceptron (MLP)} is the feed-forward-based  neural network. MLP has been used to predict acute kidney injury\cite{tomavsev2019clinically}. %(Nature) \cite{tomavsev2019clinically}.

    \item \textbf{Recurrent neural network (RNN)} \cite{rumelhart1986learning} is the most popular framework to learn the abstract embedding of variable-length time series. 
    RNN has been widely used to predict the risk of the first episode of psychosis or heart failure\cite{raket2020dynamic,choi2016doctor}. %(Lancet Digital Health) \cite{raket2020dynamic}. 

    \item \textbf{Long-short term memory network (LSTM)}~\cite{hochreiter1997long} is a variant of the Recurrent Neural Network (RNN), capable of learning long-term dependencies. LSTM has been used to perform the 90-day all-cause mortality in the intensive care unit (ICU), based on the concatenated static features and dynamic features\cite{thorsen2020dynamic}. %(Lancet Digital Health, 2020) \cite{thorsen2020dynamic}.

    \item \textbf{Gated recurrent units (GRU)}~\cite{chung2014empirical} embeds the time series as the input to perform the target prediction. It is a widely applied variant of the Recurrent Neural Network (RNN), which improves the capability to maintain historical memories and reduces parameters in the update and reset gates. GRU has been used to predict several severe complications (mortality, renal failure with a need for renal replacement therapy, and postoperative bleeding leading to operative revision) in post-cardiosurgical care in real-time\cite{meyer2018machine}. %(Lancet Respiratory Medicine, 2018) \cite{meyer2018machine}.

    \item \textbf{Temporal convolutional networks (TCN)} ~\cite{bai2018empirical} is a generic temporal convolutional network architecture for sequence modeling.
    TCN has been used to forecast hospital resource utilization (i.e., the number of hospital beds and ventilators) for COVID-19 patients~\cite{zhang2022learning}.

    \item \textbf{Transformer}~\cite{vaswani2017attention} in this experiment is the encoder part of the original Transformer, which comprises the positional encoding module and the self-attention module.
    The transformer has been used to perform the mortality risk analysis for liver transplant recipients~\cite{nitski2021long}.

\end{itemize}

\noindent\textbf{EHR-specific predictive models}

\begin{itemize}[leftmargin=*]

    \item \textbf{RETAIN}~\cite{choi2016retain} is the deep-based Reverse Time Attention model for analyzing EHR data. It utilizes a two-level neural attention module to attend important clinical visits and features. RETAIN has been used to predict heart failure by taking previous diagnoses (categorical variables) as features. In our benchmark, we modify the input layer of RETAIN with multi-layer perceptron (MLP) to make it capable of using numerical variables as features. %(Conference on Neural Information Processing Systems, 2016).

    \item \textbf{StageNet}~\cite{gao2020stagenet} is the Stage-aware neural Network, which extracts disease stage information from EHR and integrates it into risk prediction. 
    This model comprises a stage-aware LSTM module that extracts health stage variations unsupervisedly and a stage-adaptive convolutional module that incorporates stage-related progression patterns. 
    StageNet has been used to perform the decompensation prediction for ICU patients in the MIMIC-III dataset and the mortality prediction of End-Stage Renal Disease (ESRD) patients in Peking University Third Hospital. %(International World Wide Web Conference, 2020).
    
    \item \textbf{Dr. Agent}~\cite{gao2020dr} augments RNN with 2 policy gradient agents.
    It learns a dynamic skip connection to focus on the relevant information over time. 
    Dr. Agent has been used to perform the in-hospital mortality prediction, acute care phenotype classification, physiologic decompensation prediction and length of stay forecasting task on the MIMIC-III (Medical Information Mart for Intensive Care) dataset. %(Journal of the American Medical Informatics Association, 2020).
    
    \item \textbf{AdaCare}~\cite{ma2020adacare} employs a multi-scale adaptive dilated convolutional module to capture the long and short-term variations of biomarkers to depict the health status in multiple time scales. 
    AdaCare has been used to perform decompensation prediction on the MIMIC-III dataset and the mortality prediction on the End-Stage Renal Disease (ESRD) dataset. %(Association for the Advancement of Artificial Intelligence, 2020).
    
    \item \textbf{ConCare}~\cite{ma2020concare} employs multi-channel embedding architecture and self-attention mechanism to model the relationship of feature sequences and build the health representation.
    ConCare has been used to perform the morality prediction on the MIMIC-III and ESRD datasets. %(Association for the Advancement of Artificial Intelligence, 2020).
    
    \item \textbf{GRASP}~\cite{zhang2021grasp} is a generic framework for healthcare models, which leverages the information extracted from patients with similar conditions to enhance the cohort representation learning results.
    GRASP has been used to perform the sepsis prediction on the cardiology dataset and mortality prediction on the ESRD dataset.%(Association for the Advancement of Artificial Intelligence, 2021).
    % \item \textbf{CovidCare}~\cite{ma2020covidcare}
\end{itemize}

\section{Experiment Environments and Model Hyperparameters}

For DT, RF, GBDT models, we use the scikit-learn package (version 1.2.2), XGBoost (version 1.7.5), CatBoost (version 1.2) to construct and train the models. For deep learning models, we use the PyTorch (version 2.0.1) and PyTorch Lighting (version 2.0.2)  to implement the models. We use the mini-batch gradient descent to train the models and the batch size is set to 64 for \textcolor{blue}{TJH} dataset and 128 for \textcolor{red}{CDSL} dataset. The $\zeta$ and $\eta$ are set to $0.1$. For the metric calculations, $E$ is the maximum value at the 95\% percentile of the length-of-stay of all patient in both CDSL and TJH datasets. In the hold-out set, $E$ is 27.25 in the CDSL dataset and 21.44 in the TJH dataset. We use AdamW optimizer with tuned learning rate to train the models over 100 epochs. The model training process will be stopped if the score on the validation set do not improve over 10 consecutive iterations. We set the random seed to 0 in all computational runs.

Experiments are conducted on a server equipped with dual Intel Xeon Silver 4210R CPUs, each with 10 cores supporting 20 threads, two NVIDIA RTX 3090 GPUs and 64GB RAM. Notably, all experiments are executed solely on one of the GPUs for consistent results.

The model hyperparameters are shown in Table~\ref{tab:ml_config}, Table~\ref{tab:dl_config_tjh}, and ~\ref{tab:dl_config_cdsl}. To perform the 4C calculations, we followed the guidelines and methodology provided at \url{https://isaric4c.net/risk/4c/}. Also, we have incorporated additional details about the 4C mortality score in the main text. The computation of the 4C mortality score is a straightforward process. Each patient in our study is evaluated based on a set of variables — four variables in the TJH dataset and five in the CDSL dataset. For each variable, a specific score is assigned, reflecting the patient's status in that variable. The total 4C mortality score for a patient is the sum of these individual scores. Each level of the accumulated score correlates with a distinct probability of mortality. We have presented the variables used for calculating the 4C mortality scores in both datasets in Table~\ref{tab:4c_mortality_score_variables}.

\begin{table}[htbp]
    \footnotesize
    \centering
    \caption{\textit{Hyperparameter settings of machine learning models.} All hyperparameters are obtained using grid-search on the validation set.}
    \label{tab:ml_config}
\resizebox{1\textwidth}{!}{
\begin{tabular}{c|c|c|l}
\toprule
Dataset & Task & Model & \makecell[c]{Detail} \\\hline
\multirow{10}{*}{TJH}  & outcome        & RF       & max depth 10, n\_estimators 100, learning rate 0.1, criterion gini   \\
 & length of stay & RF       & max depth 15, n\_estimators 50, learning rate 1.0, criterion squared\_error \\
 & outcome        & DT       & max depth 10, min\_samples\_split 2, min\_samples\_leaf 1   \\
& length of stay & DT       & max depth 5, criterion squared\_error, min\_samples\_split 2 \\
& outcome        & GBDT     & max depth 5, n\_estimators 100, learning rate 0.1, subsample 1.0    \\
& length of stay & GBDT     & max depth 5, n\_estimators 100, learning rate 0.1, subsample 1.0 \\
& outcome        & XGBoost  & max depth 5, n\_estimators 100, learning rate 0.1, objective binary:logistic   \\
& length of stay & XGBoost  & max depth 5, n\_estimators 100, learning rate 0.1, objective reg:squarederror  \\
& outcome        & CatBoost & max depth 5, n\_estimators 100, learning rate 0.1, loss CrossEntropy   \\
& length of stay & CatBoost & max depth 5, n\_estimators 100, learning rate 0.1, loss RMSE \\
\hline
\multirow{10}{*}{CDSL} & outcome        & RF       & max depth 15, n\_estimators 100, learning rate 1.0, criterion gini \\
& length of stay & RF       & max depth 15, n\_estimators 100, learning rate 1.0, criterion squared\_error  \\
& outcome        & DT       & max depth 5, min\_samples\_split 2, min\_samples\_leaf 1  \\
& length of stay & DT       & max depth 5, criterion squared\_error, min\_samples\_split 2 \\
& outcome        & GBDT     & max depth 5, n\_estimators 100, learning rate 0.1, subsample 1.0 \\
& length of stay & GBDT     & max depth 5, n\_estimators 100, learning rate 0.1, subsample 1.0 \\
& outcome        & XGBoost  & max depth 5, n\_estimators 50, learning rate 0.1, objective binary:logistic   \\
& length of stay & XGBoost  & max depth 10, n\_estimators 50, learning rate 0.1, objective reg:squarederror  \\
& outcome        & CatBoost & max depth 5, n\_estimators 50, learning rate 0.1, loss CrossEntropy\\
& length of stay & CatBoost & max depth 10, n\_estimators 100, learning rate 0.1, loss RMSE  \\
\bottomrule
\end{tabular}}
\end{table}

\begin{table}[htbp]
    \footnotesize
    \centering
    \caption{\textit{Hyperparameter settings of deep learning models in the \textcolor{blue}{TJH} dataset.} All hyperparameters are obtained using grid-search on the validation set.}
    \resizebox{1\textwidth}{!}{
    \label{tab:dl_config_tjh}
\begin{tabular}{c|c|l}
\toprule
Task & Model & \makecell[c]{Detail} \\\hline
outcome        & MLP & hidden dim 32, learning rate 0.01, feed-forward dim 128\\
 length of stay & MLP & hidden dim 128, learning rate 0.001, feed-forward dim 512\\
 multi-task     & MLP & hidden dim 32, learning rate 0.001, feed-forward dim 128\\
 outcome        & RNN & hidden dim 64, learning rate 0.001, layers 1\\
 length of stay & RNN & hidden dim 64, learning rate 0.01, layers 1\\
 multi-task     & RNN & hidden dim 64, learning rate 0.01, layers 1\\
 outcome        & LSTM & hidden dim 128, learning rate 0.01, layers 1\\
 length of stay & LSTM & hidden dim 128, learning rate 0.01, layers 1\\
 multi-task     & LSTM & hidden dim 128, learning rate 0.01, layers 1\\
 outcome        & GRU & hidden dim 32, learning rate 0.01, layers 1\\
 length of stay & GRU & hidden dim 128, learning rate 0.01, layers 1\\
 multi-task     & GRU & hidden dim 32, learning rate 0.0001, layers 1\\
 outcome        & TCN & hidden dim 64, learning rate 0.0001, kernel size 2, dropout rate 0.2\\
 length of stay & TCN & hidden dim 64, learning rate 0.01, kernel size 2, dropout rate 0.2\\
 multi-task     & TCN & hidden dim 128, learning rate 0.01, kernel size 2, dropout rate 0.2\\
 outcome        & Transformer & hidden dim 64, learning rate 0.0001, layers 1, heads 1\\
 length of stay & Transformer & hidden dim 32, learning rate 0.001, layers 1, heads 1\\
 multi-task     & Transformer & hidden dim 32, learning rate 0.001, layers 1, heads 1\\
 outcome        & RETAIN & hidden dim 64, learning rate 0.001, dropout rate 0.1\\
 length of stay & RETAIN & hidden dim 64, learning rate 0.001, dropout rate 0.1\\
 multi-task     & RETAIN & hidden dim 64, learning rate 0.001, dropout rate 0.1\\
 outcome        & StageNet & hidden dim 64, learning rate 0.0001, levels 3, chunk size 64\\
 length of stay & StageNet & hidden dim 128, learning rate 0.01, levels 3, chunk size 128\\
 multi-task     & StageNet & hidden dim 32, learning rate 0.001, levels 3, chunk size 32\\
 outcome        & Dr. Agent & hidden dim 32, learning rate 0.01, actions 10, units 64\\
 length of stay & Dr. Agent & hidden dim 64, learning rate 0.01, actions 10, units 64\\
 multi-task     & Dr. Agent & hidden dim 64, learning rate 0.01, actions 10, units 64\\
 outcome        & AdaCare & hidden dim 32, learning rate 0.0001, kernel size 2, kernel num 64, RNN type GRU\\
 length of stay & AdaCare & hidden dim 64, learning rate 0.01, kernel size 2, kernel num 64, RNN type GRU\\
 multi-task     & AdaCare & hidden dim 64, learning rate 0.01, kernel size 2, kernel num 64, RNN type GRU\\
 outcome        & GRASP & hidden dim 128, learning rate 0.01, block GRU, clusters 12, dropout rate 0.5\\
 length of stay & GRASP & hidden dim 64, learning rate 0.01, block GRU, clusters 12, dropout rate 0.5\\
 multi-task     & GRASP & hidden dim 32, learning rate 0.01, block GRU, clusters 12, dropout rate 0.5\\
 outcome        & ConCare & hidden dim 64, learning rate 0.001, GRU layers 1, attention None\\
 length of stay & ConCare & hidden dim 64, learning rate 0.001, GRU layers 1, attention None\\
 multi-task     & ConCare & hidden dim 64, learning rate 0.001, GRU layers 1, attention None\\
\bottomrule
\end{tabular}}
\end{table}

\begin{table}[h!]
    \footnotesize
    \centering
    \caption{\textit{Hyperparameter settings of deep learning models in the \textcolor{red}{CDSL} dataset.} All hyperparameters are obtained using grid-search on the validation set.}
    \label{tab:dl_config_cdsl}
    \resizebox{1\textwidth}{!}{
\begin{tabular}{c|c|l}
\toprule
Task & Model & \makecell[c]{Detail} \\\hline
outcome        & MLP & hidden dim 32, learning rate 0.001, feed-forward dim 128\\
length of stay & MLP & hidden dim 64, learning rate 0.001, feed-forward dim 256\\
 multi-task     & MLP & hidden dim 128, learning rate 0.001, feed-forward dim 512\\
 outcome        & RNN & hidden dim 128, learning rate 0.001, layers 1\\
 length of stay & RNN & hidden dim 128, learning rate 0.01, layers 1\\
 multi-task     & RNN & hidden dim 32, learning rate 0.01, layers 1\\
 outcome        & LSTM & hidden dim 64, learning rate 0.001, layers 1\\
 length of stay & LSTM & hidden dim 32, learning rate 0.0001, layers 1\\
 multi-task     & LSTM & hidden dim 32, learning rate 0.001, layers 1\\
 outcome        & GRU & hidden dim 64, learning rate 0.01, layers 1\\
 length of stay & GRU & hidden dim 64, learning rate 0.01, layers 1\\
 multi-task     & GRU & hidden dim 32, learning rate 0.01, layers 1\\
 outcome        & TCN & hidden dim 128, learning rate 0.01, kernel size 2, dropout rate 0.2\\
 length of stay & TCN & hidden dim 128, learning rate 0.001, kernel size 2, dropout rate 0.2\\
 multi-task     & TCN & hidden dim 128, learning rate 0.001, kernel size 2, dropout rate 0.2\\
 outcome        & Transformer & hidden dim 32, learning rate 0.01, layers 1, heads 1\\
 length of stay & Transformer & hidden dim 128, learning rate 0.01, layers 1, heads 1\\
 multi-task     & Transformer & hidden dim 128, learning rate 0.01, layers 1, heads 1\\
 outcome        & RETAIN & hidden dim 64, learning rate 0.01, dropout rate 0.1\\
 length of stay & RETAIN & hidden dim 128, learning rate 0.001, dropout rate 0.1\\
 multi-task     & RETAIN & hidden dim 128, learning rate 0.01, dropout rate 0.1\\
 outcome        & StageNet & hidden dim 32, learning rate 0.001, levels 3, chunk size 32\\
 length of stay & StageNet & hidden dim 128, learning rate 0.001, levels 3, chunk size 128\\
 multi-task     & StageNet & hidden dim 32, learning rate 0.0001, levels 3, chunk size 32\\
 outcome        & Dr. Agent & hidden dim 64, learning rate 0.001, actions 10, units 64\\
 length of stay & Dr. Agent & hidden dim 32, learning rate 0.01, actions 10, units 64\\
 multi-task     & Dr. Agent & hidden dim 32, learning rate 0.01, actions 10, units 64\\
 outcome        & AdaCare & hidden dim 32, learning rate 0.01, kernel size 2, kernel num 64, RNN type GRU\\
 length of stay & AdaCare & hidden dim 128, learning rate 0.001, kernel size 2, kernel num 64, RNN type GRU\\
 multi-task     & AdaCare & hidden dim 64, learning rate 0.01, kernel size 2, kernel num 64, RNN type GRU\\
 outcome        & GRASP & hidden dim 32, learning rate 0.01, block GRU, clusters 12, dropout rate 0.5\\
 length of stay & GRASP & hidden dim 32, learning rate 0.001, block GRU, clusters 12, dropout rate 0.5\\
 multi-task     & GRASP & hidden dim 128, learning rate 0.001, block GRU, clusters 12, dropout rate 0.5\\
 outcome        & ConCare & hidden dim 64, learning rate 0.001, GRU layers 1, attention None\\
 length of stay & ConCare & hidden dim 64, learning rate 0.001, GRU layers 1, attention None\\
 multi-task     & ConCare & hidden dim 64, learning rate 0.001, GRU layers 1, attention None\\
\bottomrule
\end{tabular}}
\end{table}

\begin{table}[htbp]
% \footnotesize
\centering
    \caption{\textit{Variables employed in the 4C mortality score calculation for both TJH and CDSL datasets.}} 
    \label{tab:4c_mortality_score_variables}
        \begin{tabular}{c|c|c}
        \toprule
        TJH & CDSL & 4C Mortality Score Component\\ 
        \midrule
        Sex & Sex & Sex at birth     \\
        Age & Age & Age     \\
        Urea & Urea & Urea  \\
        CRP & CRP & C reactive protein  \\
         -  & SO2C & Peripheral oxygen saturation on room air \\
        \bottomrule
        \end{tabular}
\end{table}

\section{Evaluation of Prediction Fairness}

To evaluate the models' fairness, we test the GRU model using fairness experiments. The fairness metrics include disparate impact (DI)~\cite{feldman2015certifying}, average odds difference (AOD)~\cite{hardt2016equality}, equal opportunity difference (EOD)~\cite{hardt2016equality} and statistical parity difference (SPD)~\cite{calders2010three}. The results are shown in Table~\ref{tab:fairness}.

\begin{itemize}
    \item \textbf{Disparate Impact (DI)~\cite{feldman2015certifying}:} DI measures the ratio of the positive outcome rates between the unprivileged and privileged groups. The ideal value is 1.0, indicating equal positive outcome rates for both groups. A value less than 1 indicates that the unprivileged group is less likely to receive the positive outcome than the privileged group. This metric is given by:
    \[
    \text{DI} = \frac{P(Y=1 | D=\text{unprivileged})}{P(Y=1 | D=\text{privileged})}
    \]
    Where: \( Y \) is the outcome variable (e.g., dead or alive) and \( D \) represents the group (privileged or unprivileged). For the purpose of our study, the privileged groups are identified as male and younger patients.
    
    \item \textbf{Average Odds Difference (AOD)~\cite{hardt2016equality}:} AOD measures the average difference of FPR and TPR for the unprivileged and privileged groups. The ideal value is 0, indicating both groups have equal FPR and TPR. The formula for AOD is:
    \[
    \text{AOD} = \frac{1}{2} \left[ \left( FPR_{D=\text{unprivileged}} - FPR_{D=\text{privileged}} \right) + \left( TPR_{D=\text{unprivileged}} - TPR_{D=\text{privileged}} \right) \right]
    \]

    \item \textbf{Equal Opportunity Difference (EOD)~\cite{hardt2016equality}:} EOD is similar to AOD but only considers the TPR. The ideal value is 0, indicating both groups have equal TPR. EOD is computed as:
    \[
    \text{EOD} = TPR_{D=\text{unprivileged}} - TPR_{D=\text{privileged}}
    \]

    \item \textbf{Statistical Parity Difference (SPD)~\cite{calders2010three}:} SPD measures the difference in the probability of positive decisions between the unprivileged and privileged groups. The ideal value is 0, indicating both groups have equal probabilities of receiving a positive decision. SPD can be calculated using:
    \[
    \text{SPD} = P(Y=1 | D=\text{unprivileged}) - P(Y=1 | D=\text{privileged})
    \]
\end{itemize}

\begin{table*}[!ht]
    \footnotesize
    \centering
    \caption{\textit{Comprehensive fairness assessment across deep learning models on TJH and CDSL datasets.} The table presents a detailed evaluation of fairness metrics, including Disparate Impact (DI), Average Odds Difference (AOD), Equal Opportunity Difference (EOD), and Statistical Parity Difference (SPD), for multiple deep learning models (e.g., MLP) across different privileged groups (Sex, Age).}
    \label{tab:fairness}
\begin{tabular}{lc|cccc|cccc}
\toprule
\multicolumn{2}{c|}{Dataset} & \multicolumn{4}{c|}{TJH}                                                       & \multicolumn{4}{c}{CDSL}                                                       \\ \midrule
\multicolumn{2}{c|}{Metric}  & DI ($\uparrow$) & AOD ($\downarrow$) & EOD ($\downarrow$) & SPD ($\downarrow$) & DI ($\uparrow$) & AOD ($\downarrow$) & EOD ($\downarrow$) & SPD ($\downarrow$) \\ \midrule
\multicolumn{1}{l|}{\multirow{2}{*}{MLP}} & Sex &  0.80$\pm$0.07 & 0.01$\pm$0.02 & 0.01$\pm$0.04 & 0.09$\pm$0.04                 &  0.84$\pm$0.04 & 0.02$\pm$0.01 & 0.04$\pm$0.02 & 0.01$\pm$0.00             \\
\multicolumn{1}{l|}{}                     & Age &  0.78$\pm$0.05 & 0.02$\pm$0.01 & 0.04$\pm$0.02 & 0.09$\pm$0.02                 &  0.82$\pm$0.02 & 0.02$\pm$0.01 & 0.03$\pm$0.01 & 0.01$\pm$0.00             \\ \midrule

\multicolumn{1}{l|}{\multirow{2}{*}{RNN}} & Sex & 0.81$\pm$0.09 & 0.01$\pm$0.03 & 0.00$\pm$0.05 & 0.09$\pm$0.06                  & 0.88$\pm$0.05 & 0.02$\pm$0.01 & 0.04$\pm$0.02 & 0.01$\pm$0.01              \\
\multicolumn{1}{l|}{}                     & Age & 0.76$\pm$0.07 & 0.03$\pm$0.02 & 0.05$\pm$0.03 & 0.10$\pm$0.03                 & 0.80$\pm$0.04 & 0.03$\pm$0.01 & 0.06$\pm$0.02 & 0.02$\pm$0.00              \\ \midrule

\multicolumn{1}{l|}{\multirow{2}{*}{LSTM}} & Sex & 0.82$\pm$0.09 & 0.00$\pm$0.03 & 0.02$\pm$0.03 & 0.09$\pm$0.06                  & 0.89$\pm$0.06 & 0.02$\pm$0.01 & 0.03$\pm$0.03 & 0.01$\pm$0.01              \\
\multicolumn{1}{l|}{}                     & Age & 0.79$\pm$0.06 & 0.02$\pm$0.02 & 0.03$\pm$0.02 & 0.09$\pm$0.02                  &  0.81$\pm$0.03 & 0.03$\pm$0.01 & 0.06$\pm$0.02 & 0.02$\pm$0.00             \\ \midrule

\multicolumn{1}{l|}{\multirow{2}{*}{GRU}} & Sex & 0.81$\pm$0.10 & 0.01$\pm$0.04 & 0.01$\pm$0.06 & 0.10$\pm$0.07                  &  0.89$\pm$0.06 & 0.01$\pm$0.01 & 0.02$\pm$0.03 & 0.01$\pm$0.01             \\
\multicolumn{1}{l|}{}                     & Age & 0.76$\pm$0.07 & 0.03$\pm$0.02 & 0.05$\pm$0.03 & 0.10$\pm$0.02                  & 0.82$\pm$0.05 & 0.03$\pm$0.01 & 0.06$\pm$0.02 & 0.02$\pm$0.00               \\ \midrule

\multicolumn{1}{l|}{\multirow{2}{*}{TCN}} & Sex & 0.81$\pm$0.09 & 0.00$\pm$0.03 & 0.00$\pm$0.05 & 0.09$\pm$0.06                  &  0.88$\pm$0.07 & 0.02$\pm$0.01 & 0.05$\pm$0.03 & 0.01$\pm$0.01             \\
\multicolumn{1}{l|}{}                     & Age & 0.76$\pm$0.06 & 0.03$\pm$0.01 & 0.06$\pm$0.03 & 0.10$\pm$0.02                  & 0.79$\pm$0.03 & 0.03$\pm$0.01 & 0.06$\pm$0.02 & 0.02$\pm$0.00              \\ \midrule

\multicolumn{1}{l|}{\multirow{2}{*}{Transformer}} & Sex & 0.83$\pm$0.09 & 0.00$\pm$0.03 & 0.02$\pm$0.03 & 0.09$\pm$0.06                  &  0.86$\pm$0.05 & 0.01$\pm$0.01 & 0.01$\pm$0.03 & 0.00$\pm$0.01             \\
\multicolumn{1}{l|}{}                     & Age & 0.80$\pm$0.06 & 0.01$\pm$0.01 & 0.02$\pm$0.02 & 0.09$\pm$0.02                  &  0.79$\pm$0.05 & 0.02$\pm$0.01 & 0.04$\pm$0.02 & 0.01$\pm$0.01             \\ \midrule

\multicolumn{1}{l|}{\multirow{2}{*}{RETAIN}} & Sex & 0.79$\pm$0.09 & 0.01$\pm$0.02 & 0.00$\pm$0.04 & 0.10$\pm$0.06                  & 0.85$\pm$0.08 & 0.02$\pm$0.01 & 0.05$\pm$0.03 & 0.01$\pm$0.01              \\
\multicolumn{1}{l|}{}                     & Age & 0.76$\pm$0.06 & 0.02$\pm$0.02 & 0.05$\pm$0.03 & 0.10$\pm$0.02                  &  0.77$\pm$0.02 & 0.03$\pm$0.01 & 0.06$\pm$0.02 & 0.01$\pm$0.00             \\ \midrule

\multicolumn{1}{l|}{\multirow{2}{*}{StageNet}} & Sex & 0.80$\pm$0.10 & 0.01$\pm$0.03 & 0.00$\pm$0.06 & 0.10$\pm$0.07                  & 0.89$\pm$0.06 & 0.02$\pm$0.01 & 0.03$\pm$0.02 & 0.01$\pm$0.01              \\
\multicolumn{1}{l|}{}                     & Age & 0.76$\pm$0.07 & 0.03$\pm$0.02 & 0.06$\pm$0.03 & 0.10$\pm$0.03                  &  0.81$\pm$0.04 & 0.03$\pm$0.01 & 0.06$\pm$0.02 & 0.02$\pm$0.01             \\ \midrule

\multicolumn{1}{l|}{\multirow{2}{*}{Dr. Agent}} & Sex & 0.80$\pm$0.09 & 0.01$\pm$0.02 & 0.01$\pm$0.03 & 0.10$\pm$0.06                  &  0.86$\pm$0.06 & 0.02$\pm$0.02 & 0.03$\pm$0.04 & 0.01$\pm$0.01             \\
\multicolumn{1}{l|}{}                     & Age &  0.79$\pm$0.05 & 0.01$\pm$0.01 & 0.03$\pm$0.01 & 0.09$\pm$0.02                 & 0.82$\pm$0.06 & 0.03$\pm$0.01 & 0.06$\pm$0.02 & 0.02$\pm$0.01              \\ \midrule

\multicolumn{1}{l|}{\multirow{2}{*}{AdaCare}} & Sex & 0.79$\pm$0.10 & 0.01$\pm$0.02 & 0.01$\pm$0.04 & 0.10$\pm$0.06                  &  0.89$\pm$0.06 & 0.02$\pm$0.01 & 0.04$\pm$0.02 & 0.01$\pm$0.01             \\
\multicolumn{1}{l|}{}                     & Age & 0.77$\pm$0.06 & 0.02$\pm$0.01 & 0.03$\pm$0.03 & 0.09$\pm$0.03                  & 0.83$\pm$0.03 & 0.03$\pm$0.01 & 0.05$\pm$0.02 & 0.01$\pm$0.00              \\ \midrule

\multicolumn{1}{l|}{\multirow{2}{*}{GRASP}} & Sex  & 0.81$\pm$0.09 & 0.01$\pm$0.03 & 0.01$\pm$0.04 & 0.10$\pm$0.06                   &  0.90$\pm$0.08 & 0.01$\pm$0.01 & 0.02$\pm$0.02 & 0.01$\pm$0.01             \\
\multicolumn{1}{l|}{}                     & Age & 0.79$\pm$0.06 & 0.02$\pm$0.01 & 0.04$\pm$0.02 & 0.09$\pm$0.02                  & 0.79$\pm$0.04 & 0.03$\pm$0.01 & 0.06$\pm$0.02 & 0.02$\pm$0.01              \\ \midrule

\multicolumn{1}{l|}{\multirow{2}{*}{ConCare}} & Sex & 0.79$\pm$0.08 & 0.01$\pm$0.02 & 0.00$\pm$0.04 & 0.10$\pm$0.05                  &  0.88$\pm$0.06 & 0.03$\pm$0.01 & 0.05$\pm$0.03 & 0.01$\pm$0.01             \\
\multicolumn{1}{l|}{}                     & Age & 0.77$\pm$0.06 & 0.02$\pm$0.01 & 0.04$\pm$0.02 & 0.10$\pm$0.02                  &  0.78$\pm$0.05 & 0.03$\pm$0.01 & 0.06$\pm$0.02 & 0.02$\pm$0.00             \\ 

\bottomrule
\end{tabular}
\end{table*}

The results in Table~\ref{tab:fairness} show that most model predictions are more fair on the CDSL dataset than the TJH dataset, especially in terms of sex. This may be because the population size is limited in the TJH dataset and thus has induced population bias. Future work may include conducting a more comprehensive analysis to improve the bias and fairness of different predicting models.

\section{Benchmarking Experiment under Hold-Out Data Split Settings}
We employ a 10-fold cross-validation strategy to mitigate the bias associated with small sample sizes. Randomly splitting the test set can lead to unstable model performance due to its limited size. Cross-validation allows the model to be tested on all samples, thereby yielding more consistent performance conclusions. However, to evaluate the model's generalizability over time more comprehensively, we conduct a standard holdout experiment. We divide the dataset into training, validation, and test sets in a 7:1:2 ratio, based on admission times (using the latest 20\% of patients as the holdout dataset, as recommended). We retrain all models five times with different random seeds and report the standard deviations. The model performances are detailed in Table~\ref{tab:tjh_cdsl_holdout_los_performance} and Table~\ref{tab:tjh_cdsl_holdout_mortality_performance}.

\begin{table}[htbp]
    \footnotesize
    \centering
    \caption{\textit{Benchmarking performance of outcome-specific length-of-stay prediction on TJH and CDSL hold-out test sets.}. The reported score is in the form of $mean \pm std$. Subscript $m$ signifies a multi-task learning strategy, while subscript $t$ indicates a two-stage learning strategy. \textbf{Bold} denotes the best performance. \underline{Underline} indicates that the multi-task setting outperforms the two-stage learning strategy. The asterisk * denotes that the performance improvement against the two-stage model is statistically significant (p-value < 0.05).}
    \label{tab:tjh_cdsl_holdout_los_performance}
\begin{tabular}{l|ccc|ccc}
\toprule
Dataset & \multicolumn{3}{c}{TJH} & \multicolumn{3}{|c}{CDSL} \\
\midrule
Metric & MAE($\downarrow$) & MSE($\downarrow$) & OSMAE($\downarrow$) & MAE($\downarrow$) & MSE($\downarrow$) & OSMAE($\downarrow$) \\ 
\midrule
$\text{RF}_t$ & 5.81 $\pm$ 0.18 & 45.89 $\pm$ 2.03 & 8.18 $\pm$ 0.19 & 4.21 $\pm$ 0.00 & 40.47 $\pm$ 0.05 & 4.24 $\pm$ 0.01 \\ 
$\text{DT}_t$ & 5.00 $\pm$ 0.06 & 51.22 $\pm$ 1.83 & 9.34 $\pm$ 0.54 & 4.34 $\pm$ 0.00 & 41.82 $\pm$ 0.00 & 4.43 $\pm$ 0.00 \\ 
$\text{GBDT}_t$ & 5.98 $\pm$ 0.06 & 50.39 $\pm$ 1.40 & \textbf{7.21 $\pm$ 0.16} & 4.24 $\pm$ 0.00 & 40.58 $\pm$ 0.02 & 4.27 $\pm$ 0.01 \\ 
$\text{CatBoost}_t$ & 5.51 $\pm$ 0.14 & 41.25 $\pm$ 2.12 & 7.71 $\pm$ 0.33 & 4.22 $\pm$ 0.00 & 40.49 $\pm$ 0.08 & 4.24 $\pm$ 0.01 \\ 
$\text{XGBoost}_t$ & 5.84 $\pm$ 0.00 & 46.86 $\pm$ 0.00 & 7.87 $\pm$ 0.00 & 4.22 $\pm$ 0.00 & 40.57 $\pm$ 0.00 & 4.23 $\pm$ 0.00 \\ 
\midrule
\midrule
$\text{MLP}_t$ & 5.98 $\pm$ 0.27 & 51.09 $\pm$ 5.21 & 9.77 $\pm$ 1.18 & 4.22 $\pm$ 0.02 & 40.52 $\pm$ 0.04 & 4.20 $\pm$ 0.02 \\ 
$\text{MLP}_m$ & 6.35 $\pm$ 0.24 & 71.81 $\pm$ 17.97 & \underline{9.19 $\pm$ 0.27}* & 4.22 $\pm$ 0.04 & \underline{40.14 $\pm$ 0.27}* & \underline{4.19 $\pm$ 0.07} \\ 
$\text{RNN}_t$ & 5.56 $\pm$ 0.11 & 41.27 $\pm$ 1.42 & 8.73 $\pm$ 0.45 & 4.08 $\pm$ 0.03 & 41.10 $\pm$ 1.42 & \textbf{4.00 $\pm$ 0.05} \\ 
$\text{RNN}_m$ & 7.81 $\pm$ 0.33 & 87.99 $\pm$ 9.73 & 12.14 $\pm$ 0.78 & 4.11 $\pm$ 0.06 & \underline{\textbf{39.25 $\pm$ 0.61}}* & 4.15 $\pm$ 0.15 \\ 
$\text{LSTM}_t$ & 5.28 $\pm$ 0.57 & 40.00 $\pm$ 11.06 & 8.65 $\pm$ 0.60 & 4.06 $\pm$ 0.02 & 39.71 $\pm$ 0.14 & 4.06 $\pm$ 0.04 \\ 
$\text{LSTM}_m$ & 6.71 $\pm$ 0.48 & 70.72 $\pm$ 9.57 & 10.94 $\pm$ 1.12 & 4.18 $\pm$ 0.07 & 40.33 $\pm$ 0.60 & 4.25 $\pm$ 0.10 \\ 
$\text{GRU}_t$ & 5.87 $\pm$ 0.31 & 47.79 $\pm$ 5.05 & 8.80 $\pm$ 0.70 & 4.08 $\pm$ 0.07 & 40.57 $\pm$ 2.20 & 4.03 $\pm$ 0.09 \\ 
$\text{GRU}_m$ & \underline{5.01 $\pm$ 0.29}* & \underline{32.03 $\pm$ 3.06}* & 9.44 $\pm$ 1.47 & 4.17 $\pm$ 0.03 & \underline{40.28 $\pm$ 1.63}* & 4.29 $\pm$ 0.09 \\ 
$\text{TCN}_t$ & 6.29 $\pm$ 0.84 & 77.96 $\pm$ 34.06 & 9.53 $\pm$ 0.93 & 4.08 $\pm$ 0.06 & 40.38 $\pm$ 0.27 & 4.12 $\pm$ 0.24 \\ 
$\text{TCN}_m$ & 7.50 $\pm$ 1.16 & 162.34 $\pm$ 108.99 & 11.93 $\pm$ 1.23 & \underline{\textbf{4.04 $\pm$ 0.03}}* & \underline{39.39 $\pm$ 0.41}* & \underline{4.03 $\pm$ 0.07}* \\ 
$\text{Transformer}_t$ & 6.38 $\pm$ 0.20 & 62.65 $\pm$ 12.11 & 11.11 $\pm$ 0.91 & 4.14 $\pm$ 0.03 & 41.23 $\pm$ 0.55 & 4.60 $\pm$ 0.25 \\ 
$\text{Transformer}_m$ & 6.46 $\pm$ 0.19 & 74.44 $\pm$ 26.43 & 11.27 $\pm$ 1.35 & 4.16 $\pm$ 0.06 & 43.27 $\pm$ 1.13 & \underline{4.54 $\pm$ 0.27} \\ 
\midrule
\midrule
$\text{RETAIN}_t$ & 5.36 $\pm$ 0.33 & 39.42 $\pm$ 5.26 & 12.60 $\pm$ 2.19 & 4.17 $\pm$ 0.08 & 40.87 $\pm$ 0.35 & 4.24 $\pm$ 0.09 \\ 
$\text{RETAIN}_m$ & 6.30 $\pm$ 0.57 & 115.36 $\pm$ 72.40 & 12.63 $\pm$ 1.56 & \underline{4.11 $\pm$ 0.02}* & \underline{39.41 $\pm$ 0.49}* & 4.35 $\pm$ 0.15 \\ 
$\text{StageNet}_t$ & 5.49 $\pm$ 0.85 & 47.51 $\pm$ 22.86 & 9.45 $\pm$ 0.46 & 4.08 $\pm$ 0.03 & 40.75 $\pm$ 1.04 & 4.14 $\pm$ 0.08 \\ 
$\text{StageNet}_m$ & \underline{5.10 $\pm$ 0.43}* & \underline{33.51 $\pm$ 6.01}* & 15.09 $\pm$ 7.87 & 4.09 $\pm$ 0.02 & \underline{40.40 $\pm$ 0.43} & 4.16 $\pm$ 0.02 \\ 
$\text{Dr. Agent}_t$ & 6.36 $\pm$ 0.29 & 57.96 $\pm$ 3.57 & 9.69 $\pm$ 0.84 & 4.09 $\pm$ 0.03 & 41.17 $\pm$ 1.03 & 4.28 $\pm$ 0.20 \\ 
$\text{Dr. Agent}_m$ & 7.06 $\pm$ 0.43 & 72.02 $\pm$ 2.90 & 10.15 $\pm$ 1.17 & 4.13 $\pm$ 0.09 & \underline{39.83 $\pm$ 0.95}* & \underline{4.22 $\pm$ 0.19} \\ 
$\text{AdaCare}_t$ & \textbf{3.58 $\pm$ 0.66} & \textbf{19.65 $\pm$ 5.81} & 9.26 $\pm$ 1.58 & \textbf{4.04 $\pm$ 0.01} & 40.37 $\pm$ 0.86 & 4.10 $\pm$ 0.08 \\ 
$\text{AdaCare}_m$ & 5.13 $\pm$ 0.76 & 34.44 $\pm$ 8.48 & 16.89 $\pm$ 8.79 & 4.12 $\pm$ 0.19 & 43.40 $\pm$ 4.03 & 4.11 $\pm$ 0.12 \\ 
$\text{GRASP}_t$ & 5.40 $\pm$ 0.53 & 40.75 $\pm$ 10.74 & 8.94 $\pm$ 0.74 & 4.20 $\pm$ 0.07 & 41.00 $\pm$ 0.20 & 4.29 $\pm$ 0.10 \\ 
$\text{GRASP}_m$ & 5.80 $\pm$ 0.90 & 47.69 $\pm$ 18.62 & 11.91 $\pm$ 4.34 & \underline{4.18 $\pm$ 0.06} & 41.03 $\pm$ 0.93 & 4.31 $\pm$ 0.09 \\ 
$\text{ConCare}_t$ & 4.79 $\pm$ 0.44 & 29.92 $\pm$ 5.63 & 8.15 $\pm$ 0.22 & 4.16 $\pm$ 0.02 & 41.12 $\pm$ 0.45 & 4.08 $\pm$ 0.03 \\ 
$\text{ConCare}_m$ & 7.62 $\pm$ 0.20 & 78.77 $\pm$ 6.50 & 10.24 $\pm$ 0.43 & \underline{4.13 $\pm$ 0.16} & \underline{39.36 $\pm$ 0.62}* & 4.12 $\pm$ 0.23 \\ 
\bottomrule
\end{tabular}
\end{table}

%%% outcome
    \begin{table}[htbp]
        \footnotesize
        \centering
        \caption{\textit{Benchmarking performance on the task of early mortality prediction on TJH and CDSL hold-out test sets.} The reported score is of the form $mean \pm std$. `TA' denotes the model trained with the time-aware loss. \textbf{Bold} denotes the best performance. \underline{Underline} indicates that the model with time-aware loss outperforms the original model. The asterisk * denotes that the performance improvement against the model without TA version is statistically significant (p-value < 0.05). All three metrics are multiplied by 100 for readability purposes.}
        \label{tab:tjh_cdsl_holdout_mortality_performance}
    \begin{tabular}{l|ccc|ccc}
\toprule
Dataset & \multicolumn{3}{c}{TJH} & \multicolumn{3}{|c}{CDSL} \\
    \midrule
    Metric & AUPRC($\uparrow$) & AUROC($\uparrow$) & ES($\uparrow$) & AUPRC($\uparrow$) & AUROC($\uparrow$) & ES($\uparrow$) \\ 
\midrule
RF & 99.48 $\pm$ 0.04 & 98.13 $\pm$ 0.19 & 52.16 $\pm$ 2.32 & 52.16 $\pm$ 2.32 & 83.28 $\pm$ 0.21 & -8.08 $\pm$ 0.30 \\ 
DT & 94.31 $\pm$ 0.80 & 85.68 $\pm$ 1.80 & 44.84 $\pm$ 3.33 & 27.71 $\pm$ 0.00 & 76.58 $\pm$ 0.00 & -4.47 $\pm$ 0.00 \\ 
GBDT & 99.34 $\pm$ 0.03 & 97.57 $\pm$ 0.11 & 50.80 $\pm$ 0.62 & 41.91 $\pm$ 0.28 & \textbf{84.71 $\pm$ 0.04} & 3.55 $\pm$ 0.35 \\ 
CatBoost & 99.38 $\pm$ 0.11 & 97.74 $\pm$ 0.39 & 67.62 $\pm$ 6.16 & 40.00 $\pm$ 0.36 & 84.17 $\pm$ 0.23 & -6.91 $\pm$ 0.78 \\ 
XGBoost & 99.46 $\pm$ 0.00 & 98.17 $\pm$ 0.00 & 39.92 $\pm$ 0.00 & 40.32 $\pm$ 0.00 & 84.44 $\pm$ 0.00 & -1.22 $\pm$ 0.00 \\ 
\midrule
\midrule
MLP & 99.03 $\pm$ 0.82 & 97.63 $\pm$ 0.98 & 69.16 $\pm$ 8.37 & 39.63 $\pm$ 0.79 & 83.41 $\pm$ 0.36 & -3.68 $\pm$ 1.24 \\ 
MLP-TA & 98.75 $\pm$ 0.74 & 96.94 $\pm$ 1.01 & \underline{69.44 $\pm$ 6.61} & \underline{40.46 $\pm$ 0.89}* & \underline{83.49 $\pm$ 0.32} & \underline{-2.26 $\pm$ 2.14}* \\ 
RNN & 99.69 $\pm$ 0.14 & 98.94 $\pm$ 0.46 & 75.56 $\pm$ 4.28 & 41.91 $\pm$ 1.20 & 83.36 $\pm$ 0.90 & 5.44 $\pm$ 3.40 \\ 
RNN-TA & \underline{99.71 $\pm$ 0.13}* & \underline{98.99 $\pm$ 0.44}* & \underline{75.78 $\pm$ 4.04} & \underline{42.65 $\pm$ 1.17}* & 83.35 $\pm$ 0.63 & \underline{7.00 $\pm$ 5.62}* \\ 
LSTM & 99.14 $\pm$ 0.14 & 97.50 $\pm$ 0.24 & 73.70 $\pm$ 1.79 & 39.76 $\pm$ 1.40 & 81.73 $\pm$ 0.77 & 10.56 $\pm$ 1.76 \\ 
LSTM-TA & \underline{99.16 $\pm$ 0.14} & \underline{97.56 $\pm$ 0.23} & \underline{74.15 $\pm$ 2.61}* & \underline{40.77 $\pm$ 1.32}* & \underline{82.39 $\pm$ 0.53}* & \underline{13.03 $\pm$ 2.66}* \\ 
GRU & 99.43 $\pm$ 0.25 & 98.07 $\pm$ 0.78 & 75.74 $\pm$ 3.88 & 41.43 $\pm$ 1.93 & 83.28 $\pm$ 0.90 & 5.38 $\pm$ 5.13 \\ 
GRU-TA & \underline{99.48 $\pm$ 0.20} & \underline{98.21 $\pm$ 0.64}* & \underline{76.42 $\pm$ 2.06}* & \underline{42.18 $\pm$ 1.35}* & \underline{83.64 $\pm$ 0.82}* & \underline{9.01 $\pm$ 5.59}* \\ 
TCN & 98.71 $\pm$ 0.23 & 96.84 $\pm$ 0.25 & 74.38 $\pm$ 1.29 & 39.64 $\pm$ 1.48 & 83.50 $\pm$ 0.44 & 4.17 $\pm$ 8.71 \\ 
TCN-TA & \underline{98.72 $\pm$ 0.20 }& 96.84 $\pm$ 0.22 & 73.70 $\pm$ 1.79 & \underline{40.61 $\pm$ 1.68}* & 83.46 $\pm$ 0.44 & 3.45 $\pm$ 3.26 \\ 
Transformer & 99.06 $\pm$ 0.86 & 97.84 $\pm$ 1.23 & 64.22 $\pm$ 6.73 & 31.18 $\pm$ 4.86 & 80.04 $\pm$ 1.15 & -1.93 $\pm$ 12.86 \\ 
Transformer-TA & \underline{99.53 $\pm$ 0.22} & \underline{98.52 $\pm$ 0.56} & 64.22 $\pm$ 6.81 & \underline{32.31 $\pm$ 2.44}* & \underline{80.89 $\pm$ 1.22}* & \underline{9.48 $\pm$ 6.30}* \\ 
\midrule
\midrule
RETAIN & 97.23 $\pm$ 1.21 & 92.70 $\pm$ 2.90 & 50.39 $\pm$ 15.40 & 37.65 $\pm$ 2.17 & 81.84 $\pm$ 1.33 & 1.92 $\pm$ 9.56 \\ 
RETAIN-TA & \underline{97.50 $\pm$ 1.11} & \underline{93.14 $\pm$ 2.82} & \underline{52.66 $\pm$ 16.36} & \underline{38.83 $\pm$ 2.30}* & \underline{82.42 $\pm$ 1.65}* & \underline{5.31 $\pm$ 9.79}* \\ 
StageNet & 99.35 $\pm$ 0.26 & 97.94 $\pm$ 0.64 & 70.07 $\pm$ 2.59 & 42.08 $\pm$ 0.53 & 83.09 $\pm$ 0.36 & 12.88 $\pm$ 3.08 \\ 
StageNet-TA & \underline{99.38 $\pm$ 0.25}* & \underline{98.06 $\pm$ 0.63}* & \underline{70.98 $\pm$ 3.98}* & \underline{42.51 $\pm$ 0.52}* & \underline{83.17 $\pm$ 0.32} & \underline{\textbf{14.06 $\pm$ 2.80}}* \\ 
Dr. Agent & 99.35 $\pm$ 0.24 & 97.85 $\pm$ 0.57 & 72.57 $\pm$ 3.84 & 40.83 $\pm$ 1.18 & 82.98 $\pm$ 0.78 & 11.18 $\pm$ 6.92 \\ 
Dr. Agent-TA & \underline{99.37 $\pm$ 0.24} & \underline{97.92 $\pm$ 0.57} & \underline{73.47 $\pm$ 2.93}* & \underline{41.87 $\pm$ 1.45}* & \underline{83.07 $\pm$ 1.12}* & \underline{12.46 $\pm$ 6.58}* \\ 
AdaCare & 99.85 $\pm$ 0.13 & 99.49 $\pm$ 0.41 & 62.32 $\pm$ 10.08 & 40.78 $\pm$ 0.94 & 83.46 $\pm$ 0.12 & 8.27 $\pm$ 1.81 \\ 
AdaCare-TA & \textbf{\underline{99.87 $\pm$ 0.09}}* & \textbf{\underline{99.53 $\pm$ 0.32}}* & \underline{69.35 $\pm$ 4.41}* & \underline{41.90 $\pm$ 1.15}* & \underline{83.72 $\pm$ 0.28}* & \underline{9.70 $\pm$ 1.59}* \\ 
GRASP & 99.42 $\pm$ 0.16 & 98.19 $\pm$ 0.39 & 72.79 $\pm$ 4.63 & 33.13 $\pm$ 1.10 & 75.04 $\pm$ 0.74 & -2.32 $\pm$ 6.00 \\ 
GRASP-TA & \underline{99.49 $\pm$ 0.22} & \underline{98.40 $\pm$ 0.57}* & 72.79 $\pm$ 4.90 & \underline{34.57 $\pm$ 1.12}* & \underline{75.43 $\pm$ 1.20} & \underline{1.92 $\pm$ 2.39} \\ 
ConCare & 99.50 $\pm$ 0.20 & 98.24 $\pm$ 0.55 & 77.33 $\pm$ 2.82 & 43.12 $\pm$ 0.09 & 84.39 $\pm$ 0.08 & -1.23 $\pm$ 5.16 \\ 
ConCare-TA & 99.49 $\pm$ 0.24 & 98.22 $\pm$ 0.66 & \textbf{\underline{78.91 $\pm$ 1.72}}* & \underline{\textbf{43.47 $\pm$ 0.14}}* & 84.28 $\pm$ 0.17 & \underline{2.55 $\pm$ 3.38}* \\ 
\bottomrule
\end{tabular}
\end{table}

For outcome-specific LOS prediction on the TJH dataset, AdaCare achieves the lowest MAE and MSE. On the CDSL dataset, TCN, AdaCare, and RNN exhibit the lowest MAE, MSE, and OSMAE. In the early mortality prediction task, the results align closely with those from the cross-validation setting. AdaCare and ConCare demonstrate superior performance in AUROC, AUPRC, and ES on the TJH dataset. Meanwhile, ConCare, StageNet, and GBDT achieve the best performance on the CDSL dataset. We observe significant variability in some model performances, particularly in MSE on the TJH dataset, attributable mainly to its smaller size and consequent less stable conclusions compared to the cross-validation setting. Furthermore, we note that overall model performance on the hold-out dataset is inferior to that in the cross-validation setting, suggesting a shift in dataset distribution over time and a corresponding decline in the models' predictive capabilities. Overall, AdaCare, Dr. Agent and StageNet are top-performing models for both tasks on both datasets. For a light-weight choice, GBDT outperformed other machine learning models and could be considered as the choice for machine learning models.

\section{Benchmarking Experiment on MIMIC Datasets}

To enhance the comprehensiveness and validity of the evaluation, we include a broader array of ICU datasets (MIMIC-III~\cite{johnson2016mimic} and MIMIC-IV~\cite{johnson2020mimic}). We conduct the in-hospital mortality prediction task on the two datasets. The label definition and data preprocessing are following previous benchmark works~\cite{harutyunyan2019multitask}, but we provide a more comprehensive baseline comparison. The results are shown in Table~\ref{tab:mimic_holdout_mortality_performance}.

We find that Dr. Agent and StageNet achieve better performance on the MIMIC-III dataset, while Dr. Agent and GRU achieve better performance on the MIMIC-IV dataset. This conclusion is similar to the CDSL and TJH dataset, which proves the generalizability of these models. Since the LOS information in the two MIMIC datasets has high variance, we do not report the OSMAE on these datasets. In future works, these metrics can be applied to more specific cohorts that have lower LOS variance. 

\begin{figure}[htbp]
\centering
\begin{subfigure}{0.45\textwidth}
  \centering
  \includegraphics[width=\textwidth]{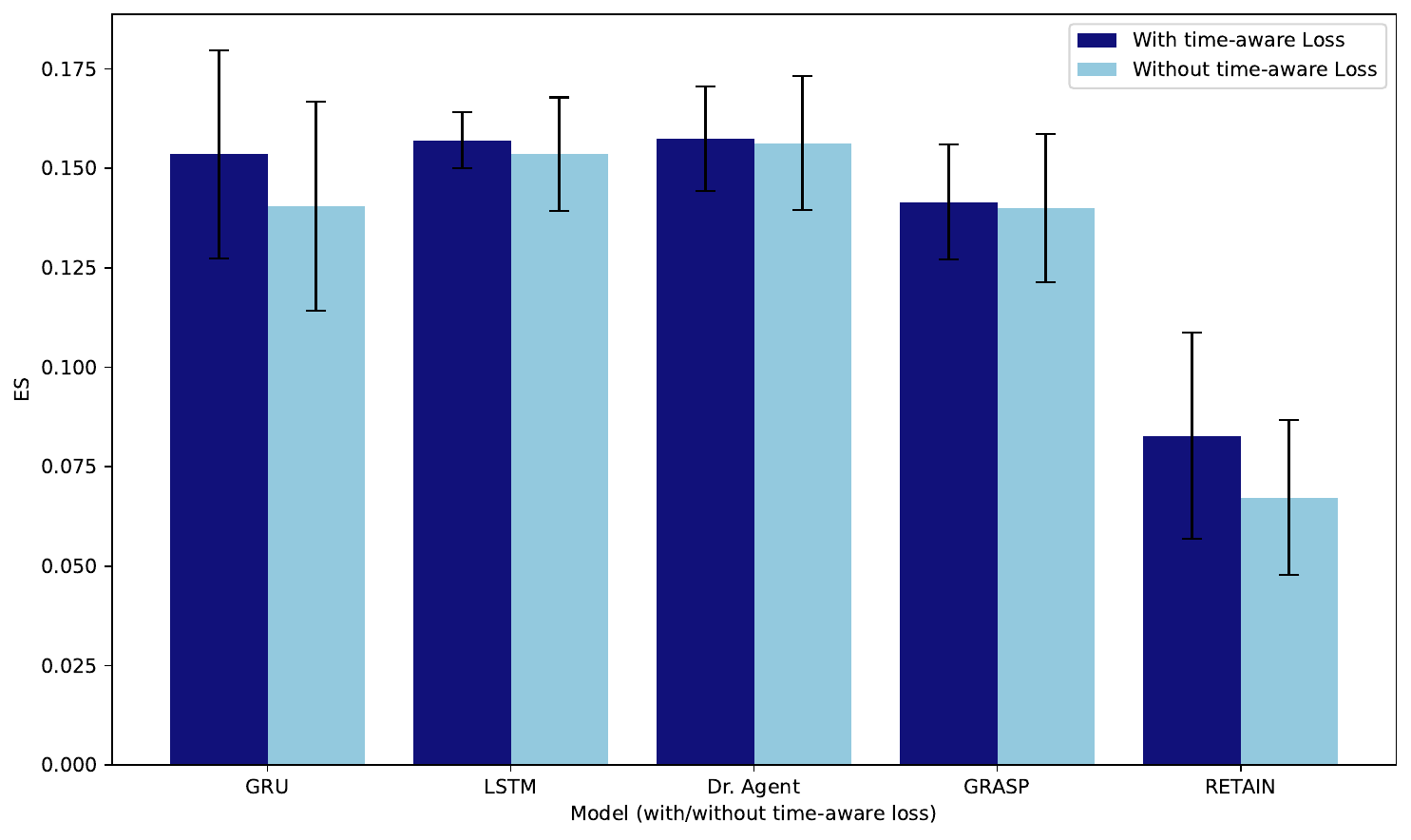} % first figure itself
  \caption{MIMIC-III}
  \label{fig:ta_term_iii}
\end{subfigure}%
\begin{subfigure}{0.45\textwidth}
  \centering
  \includegraphics[width=\textwidth]{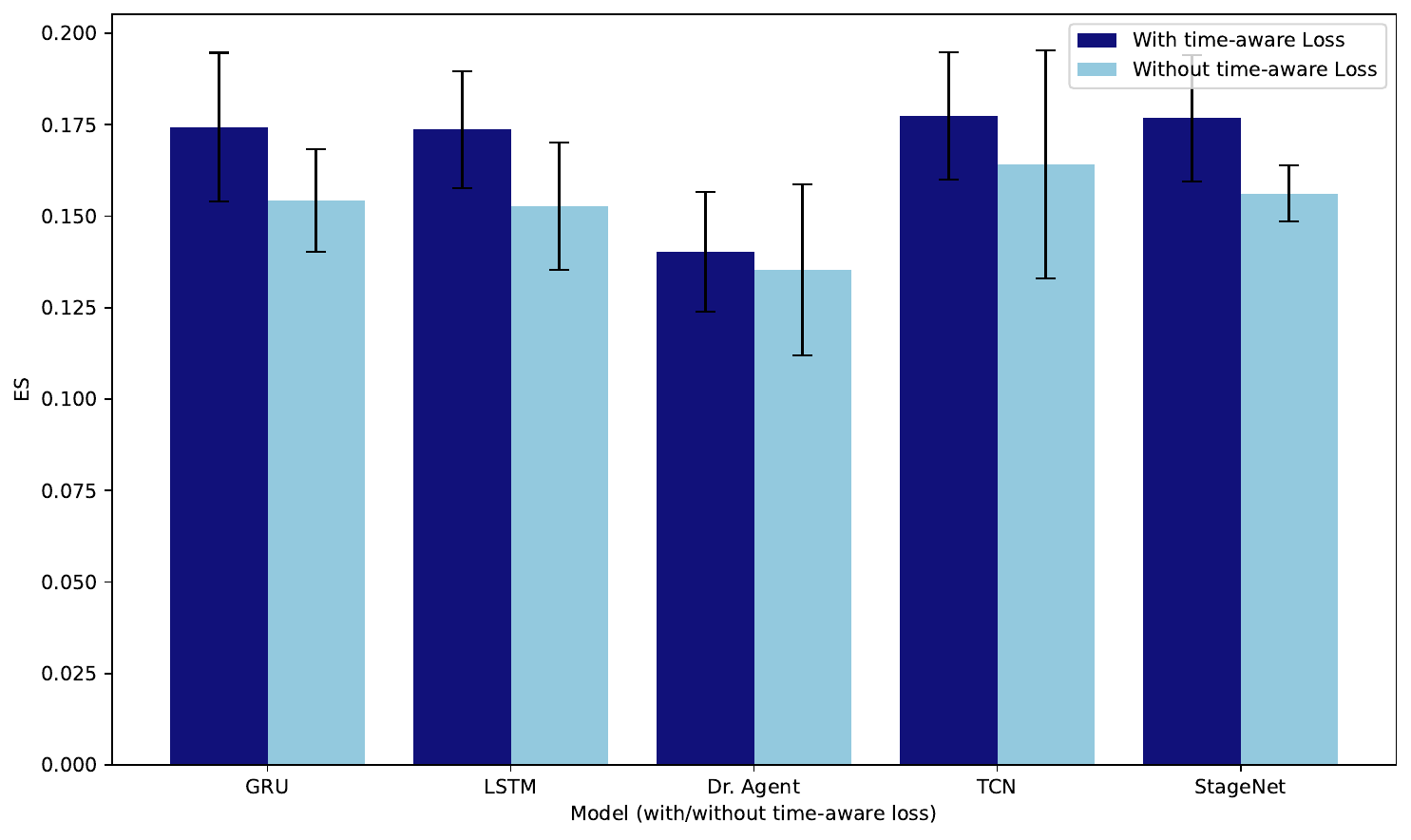} % second figure itself
  \caption{MIMIC-IV}
  \label{fig:ta_term_iv}
\end{subfigure}
\caption{\textit{Early prediction performance of 5 models with the highest ES on the MIMIC-III and MIMIC-IV datasets.} All models are trained using the first half of patient records. Error bars are standard deviations.}
\label{fig:ta_loss_mimic}
\end{figure}

Similar to the TJH and CDSL datasets, we also explore the early prediction setting on the MIMIC-III and MIMIC-IV datasets. As shown in Figure~\ref{fig:ta_loss_mimic}, the time-aware loss can effectively improve various models' prediction performance on more general datasets. The MIMIC dataset statistics used in our experiments are shown in Table~\ref{tab:summary_statistics_mimic3} and \ref{tab:summary_statistics_mimic4}.

\begin{table}[htbp]
    \footnotesize
    \centering
    \caption{\textit{Benchmarking performance on the task of early mortality prediction on MIMIC-III and MIMIC-IV hold-out test sets.} The reported score is of the form $mean \pm std$. `TA' denotes the model trained with the time-aware loss. \textbf{Bold} denotes the best performance. \underline{Underline} indicates that the model with time-aware loss outperforms the original model. The asterisk * denotes that the performance improvement against the model without TA version is statistically significant (p-value < 0.05). All three metrics are multiplied by 100 for readability purposes.}
    \label{tab:mimic_holdout_mortality_performance}
\begin{tabular}{l|ccc|ccc}
\toprule
Dataset & \multicolumn{3}{c}{MIMIC-III} & \multicolumn{3}{|c}{MIMIC-IV} \\
    \midrule
    Metric & AUPRC($\uparrow$) & AUROC($\uparrow$) & ES($\uparrow$) & AUPRC($\uparrow$) & AUROC($\uparrow$) & ES($\uparrow$) \\ 
\midrule
RF & 43.56 $\pm$ 0.24 & 81.94 $\pm$ 0.08 & -1.58 $\pm$ 0.31 & 39.46 $\pm$ 0.21 & 80.58 $\pm$ 0.06 & 0.50 $\pm$ 0.23 \\ 
DT & 31.35 $\pm$ 0.16 & 70.65 $\pm$ 0.17 & 7.45 $\pm$ 0.20 & 29.56 $\pm$ 0.22 & 74.04 $\pm$ 0.22 & 9.49 $\pm$ 0.16 \\ 
GBDT & 45.73 $\pm$ 0.02 & 83.67 $\pm$ 0.00 & 11.80 $\pm$ 0.03 & 42.06 $\pm$ 0.04 & 81.80 $\pm$ 0.01 & 13.57 $\pm$ 0.05 \\ 
CatBoost & 45.44 $\pm$ 0.34 & 83.50 $\pm$ 0.08 & 6.72 $\pm$ 0.14 & 41.44 $\pm$ 0.11 & 81.66 $\pm$ 0.04 & 8.67 $\pm$ 0.27 \\ 
XGBoost & 45.73 $\pm$ 0.20 & 83.65 $\pm$ 0.04 & 10.68 $\pm$ 0.43 & 41.54 $\pm$ 0.00 & 81.76 $\pm$ 0.00 & 12.29 $\pm$ 0.00 \\ 
\midrule
\midrule
MLP & 45.01 $\pm$ 0.28 & 83.04 $\pm$ 0.07 & 12.73 $\pm$ 1.51 & 41.39 $\pm$ 0.54 & 81.35 $\pm$ 0.17 & 10.66 $\pm$ 1.29 \\ 
MLP-TA & 44.92 $\pm$ 0.57 & 82.86 $\pm$ 0.18 & \underline{13.68 $\pm$ 1.70}* & 40.97 $\pm$ 0.35 & 81.18 $\pm$ 0.14 & \underline{13.80 $\pm$ 1.22}* \\ 
RNN & 45.87 $\pm$ 0.61 & 83.34 $\pm$ 0.25 & 13.45 $\pm$ 0.96 & 44.08 $\pm$ 0.21 & 82.85 $\pm$ 0.10 & 15.93 $\pm$ 2.35 \\ 
RNN-TA & 45.38 $\pm$ 0.52 & 82.99 $\pm$ 0.20 & \underline{13.69 $\pm$ 1.12} & 43.59 $\pm$ 0.34 & 82.59 $\pm$ 0.16 & \underline{\textbf{17.88 $\pm$ 1.81}}* \\ 
LSTM & 47.46 $\pm$ 0.34 & 83.66 $\pm$ 0.06 & 15.35 $\pm$ 1.44 & 44.52 $\pm$ 0.49 & 83.14 $\pm$ 0.21 & 15.27 $\pm$ 1.73 \\ 
LSTM-TA & 47.14 $\pm$ 0.35 & 83.29 $\pm$ 0.15 & \underline{15.70 $\pm$ 1.32} & 44.34 $\pm$ 0.44 & 82.86 $\pm$ 0.15 & \underline{17.37 $\pm$ 1.65}* \\ 
GRU & 47.19 $\pm$ 0.45 & 83.80 $\pm$ 0.18 & 14.05 $\pm$ 0.70 & \textbf{45.58 $\pm$ 0.18} & 83.50 $\pm$ 0.09 & 15.43 $\pm$ 1.60 \\ 
GRU-TA & 46.87 $\pm$ 0.45 & 83.39 $\pm$ 0.19 & \underline{15.35 $\pm$ 2.62}* & 45.33 $\pm$ 0.22 & 83.18 $\pm$ 0.14 & \underline{17.44 $\pm$ 2.03}* \\ 
TCN & 47.71 $\pm$ 0.46 & 84.17 $\pm$ 0.13 & \textbf{18.72 $\pm$ 1.85} & 45.00 $\pm$ 0.51 & \textbf{83.53 $\pm$ 0.23} & 16.42 $\pm$ 2.34 \\ 
TCN-TA & \underline{47.73 $\pm$ 0.24} & 84.04 $\pm$ 0.13 & 18.10 $\pm$ 2.31 & \underline{45.03 $\pm$ 0.59} & 83.47 $\pm$ 0.20 & \underline{17.74 $\pm$ 1.75}* \\ 
Transformer & 42.51 $\pm$ 0.49 & 82.43 $\pm$ 0.16 & 8.55 $\pm$ 1.88 & 42.32 $\pm$ 0.24 & 82.54 $\pm$ 0.16 & 10.12 $\pm$ 3.95 \\ 
Transformer-TA & 42.38 $\pm$ 0.54 & 82.24 $\pm$ 0.18 & \underline{9.68 $\pm$ 2.31}* & 42.24 $\pm$ 0.36 & 82.54 $\pm$ 0.14 & \underline{13.77 $\pm$ 4.17}* \\ 
\midrule
\midrule
RETAIN & 46.72 $\pm$ 0.44 & 83.61 $\pm$ 0.23 & 6.72 $\pm$ 1.94 & 44.89 $\pm$ 0.54 & 83.34 $\pm$ 0.26 & 6.65 $\pm$ 0.59 \\ 
RETAIN-TA & 46.49 $\pm$ 0.35 & 83.21 $\pm$ 0.15 & \underline{8.27 $\pm$ 1.86}* & 44.48 $\pm$ 0.25 & 82.68 $\pm$ 0.22 & \underline{7.33 $\pm$ 1.19}* \\ 
StageNet & 47.33 $\pm$ 0.38 & 83.60 $\pm$ 0.16 & 16.66 $\pm$ 2.23 & 44.79 $\pm$ 0.26 & 83.21 $\pm$ 0.35 & 15.62 $\pm$ 0.77 \\ 
StageNet-TA & 46.81 $\pm$ 0.41 & 83.22 $\pm$ 0.19 & 16.49 $\pm$ 1.26 & 44.22 $\pm$ 0.44 & 82.70 $\pm$ 0.33 & \underline{17.68 $\pm$ 3.12}* \\ 
Dr. Agent & 47.94 $\pm$ 0.33 & \textbf{84.30 $\pm$ 0.10} & 15.71 $\pm$ 2.32 & 44.53 $\pm$ 0.69 & 83.15 $\pm$ 0.24 & 13.54 $\pm$ 1.53 \\ 
Dr. Agent-TA & \underline{\textbf{48.10 $\pm$ 0.23}}* & 84.17 $\pm$ 0.13 & \underline{16.04 $\pm$ 2.59}* & 44.00 $\pm$ 0.29 & 82.67 $\pm$ 0.15 & \underline{13.84 $\pm$ 1.64} \\ 
AdaCare & 47.32 $\pm$ 0.37 & 83.62 $\pm$ 0.33 & 15.90 $\pm$ 1.16 & 44.41 $\pm$ 0.13 & 82.74 $\pm$ 0.13 & 15.33 $\pm$ 2.10 \\ 
AdaCare-TA & 47.29 $\pm$ 0.24 & 83.49 $\pm$ 0.23 & 15.45 $\pm$ 1.85 & 44.19 $\pm$ 0.16 & 82.42 $\pm$ 0.11 & \underline{17.78 $\pm$ 2.22}* \\ 
GRASP & 46.96 $\pm$ 0.19 & 83.55 $\pm$ 0.11 & 13.95 $\pm$ 1.66 & 44.22 $\pm$ 0.31 & 82.87 $\pm$ 0.12 & 14.54 $\pm$ 0.81 \\ 
GRASP-TA & 46.64 $\pm$ 0.33 & 83.13 $\pm$ 0.18 & \underline{14.13 $\pm$ 1.56} & 44.03 $\pm$ 0.41 & 82.49 $\pm$ 0.15 & \underline{16.03 $\pm$ 1.33}* \\ 
ConCare & 46.40 $\pm$ 0.30 & 83.03 $\pm$ 0.10 & 15.30 $\pm$ 1.25 & 44.10 $\pm$ 0.47 & 82.70 $\pm$ 0.22 & 13.92 $\pm$ 3.04 \\ 
ConCare-TA & 45.91 $\pm$ 0.49 & 82.74 $\pm$ 0.26 & 15.29 $\pm$ 0.91 & 43.92 $\pm$ 0.46 & 82.48 $\pm$ 0.18 & \underline{15.62 $\pm$ 1.55}* \\ 
\bottomrule
\end{tabular}
\end{table}

\begin{table}[htbp]
    \footnotesize
    \centering
    \caption{\textit{Statistics of the \textcolor{blue}{MIMIC-III} dataset.} The reported statistics are of the form $Median [Q1, Q3]$.}
    \label{tab:summary_statistics_mimic3}
\begin{tabular}{l|ccc}
\toprule
Mortality Outcome & Total & Alive & Dead\\
\midrule
\# Patients &  41517  &  37108 (89.38 \%) & 4409 (10.62 \%)  \\
\# Records & 3509005  & 2952672 (84.15 \%) & 556333 (15.85 \%) \\
\# Avg. records & 46.0 [26.0, 88.0]  & 45.0 [26.0, 80.0] & 66.0 [26.0, 154.0]  \\
\midrule
Age & 65.5 [52.4, 77.9] & 64.5 [51.6, 77.0] & 74.6 [60.7, 83.3]    \\
Age > Avg. (75.1) & 12852(30.96 \%) & 10703(28.84 \%) & 2149(48.74 \%) \\
Age $\leq$ Avg. (75.1) & 28665(69.04 \%) & 26405(71.16 \%) & 2260(51.26 \%) \\
\midrule
Gender & 55.9\% Male & 56.3\% Male & 52.8\% Male \\
Male & 23214(55.91 \%) & 20885(56.28 \%) & 2329(52.82 \%) \\
Female & 18303(44.09 \%) & 16223(43.72 \%) & 2080(47.18 \%) \\
\midrule
\# Features & \multicolumn{3}{c}{61} \\
Length of stay & 46.9 [26.5, 88.5] & 45.9 [26.3, 80.5] & 68.6 [29.3, 158.8]\\
\bottomrule
\end{tabular}
\end{table}

\begin{table}[htbp]
    \footnotesize
    \centering
    \caption{\textit{Statistics of the \textcolor{blue}{MIMIC-IV} dataset.} The reported statistics are of the form $Median [Q1, Q3]$.}
    \label{tab:summary_statistics_mimic4}
\begin{tabular}{l|ccc}
\toprule
Mortality Outcome & Total & Alive & Dead\\
\midrule
\# Patients &  56888  & 51451 (90.44 \%) & 5437 (9.56 \%)  \\
\# Records & 4055519  & 3466575 (85.48 \%) & 588944 (14.52 \%) \\
\# Avg. records & 42.0 [24.0, 75.0]  & 41.0 [24.0, 72.0] & 59.0 [24.0, 137.0]  \\
\midrule
Age & 65.0 [53.0, 76.0] & 64.0 [52.0, 75.0] & 72.0 [60.0, 82.0]    \\
Age > Avg. (63.1) & 29950(52.65 \%) & 26232(50.98 \%) & 3718(68.38 \%) \\
Age $\leq$ Avg. (63.1) & 26938(47.35 \%) & 25219(49.02 \%) & 1719(31.62 \%) \\
\midrule
Gender & 55.7\% Male & 55.8\% Male & 54.1\% Male \\
Male & 31669(55.67 \%) & 28726(55.83 \%) & 2943(54.13 \%) \\
Female & 25219(44.33 \%) & 22725(44.17 \%) & 2494(45.87 \%) \\
\midrule
\# Features & \multicolumn{3}{c}{61} \\
Length of stay & 42.1 [24.0, 75.2] & 41.0 [24.0, 71.9] & 61.2[25.2, 138.6] \\
\bottomrule
\end{tabular}
\end{table}

\section{Model Embedding Visualization for Two-Stage and Multi-task Settings}
\begin{figure}[htbp]
\centering
\begin{subfigure}{0.45\textwidth}
  \centering
  \includegraphics[width=\textwidth]{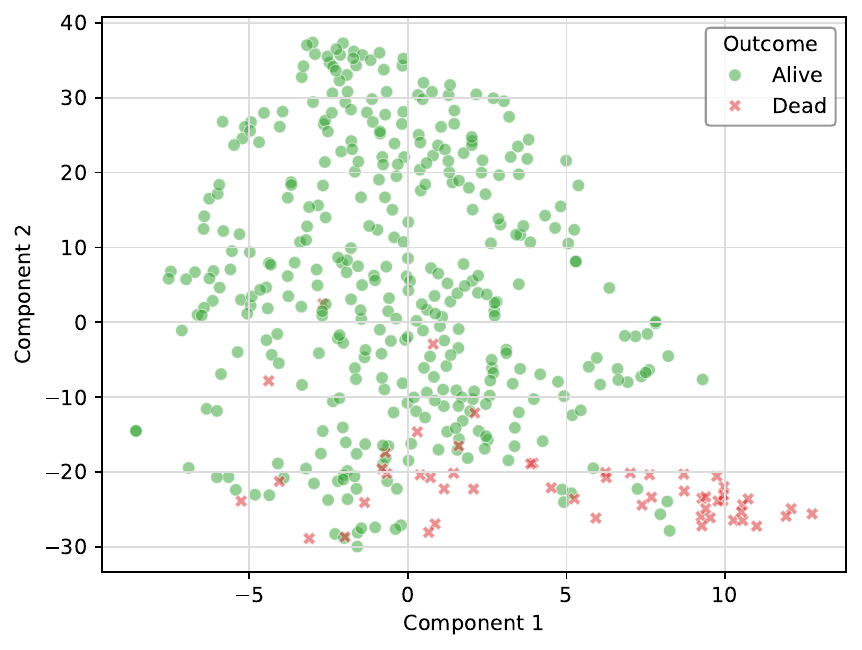} % first figure itself
  \caption{TCN model's embedding under two-stage setting}
  \label{fig:tcn_los_embedding_tsne}
\end{subfigure}%
\begin{subfigure}{0.45\textwidth}
  \centering
  \includegraphics[width=\textwidth]{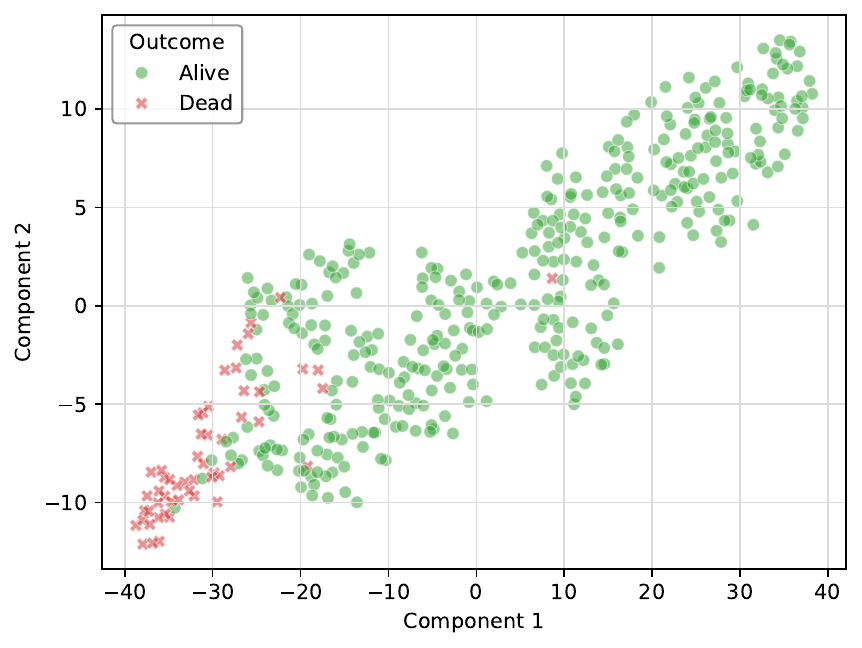} % second figure itself
  \caption{TCN model's embedding under multi-task setting}
  \label{fig:tcn_multi_embedding_tsne}
\end{subfigure}
\caption{\textit{Visualization of TCN model's embedding for LOS prediction under two settings using t-SNE.} The embeddings are extracted at each patient's last visit on the CDSL dataset.}
\label{fig:tcn_embedding_tsne}
\end{figure}

To more effectively illustrate the learned embeddings from both model types, we employ t-SNE for visualizing the hidden states of patient embeddings. As depicted in Figure \ref{fig:tcn_embedding_tsne}, the left figure presents the embeddings of the TCN model under a two-stage setting, while the right figure displays the embeddings from a multi-task setting. The embeddings derived under the multi-task setting appear more clustered, suggesting that the model is capable of learning more consistent embeddings in the latent space. This clustering is likely due to the model being tasked with optimizing for two objectives simultaneously.

\section{P-Values of T-Tests}
To assess the statistical significance of our model's performance, we performed t-tests on all results. The bootstrap T-test~\cite{dwivedi2017analysis} was employed to calculate p-values. We set the sample number for the bootstrap at 1000 and the number of sampling processes at 50, including all results from the 10-fold prediction process. The comparisons were made between different variations of the model — for instance, comparing the model with and without Time-Aware (TA) loss in the early mortality prediction task, and between two-stage and multi-task settings in the LOS prediction task. We report the p-value of AUPRC in the early mortality prediction task and MAE in the LOS prediction task in Table~\ref{tab:pvalue_los} and \ref{tab:pvalue_mortality}.

\begin{table}[htbp]
\centering
\caption{\textit{P-values of two-stage vs. multi-task performance in LOS predictions using t-test.} P-values are rounded to three decimal places; a value of 0.000 indicates a rounded p-value < 0.0005.}
\begin{tabular}{l|ccc|ccc}
\toprule
Dataset     & \multicolumn{3}{c|}{TJH} & \multicolumn{3}{c}{CDSL} \\
\midrule
Model       & MAE    & MSE    & OSMAE & MAE    & MSE    & OSMAE  \\
\midrule
MLP         & 0.000  & 0.000  & 0.738 & 0.000  & 0.000  & 0.000  \\
RNN         & 0.000  & 0.000  & 0.000 & 0.000  & 0.000  & 0.000  \\
LSTM        & 0.000  & 0.000  & 0.000 & 0.000  & 0.000  & 0.000  \\
GRU         & 0.000  & 0.011  & 0.000 & 0.000  & 0.000  & 0.000  \\
TCN         & 0.000  & 0.000  & 0.000 & 0.000  & 0.000  & 0.000  \\
Transformer & 0.000  & 0.349  & 0.000 & 0.000  & 0.000  & 0.000  \\
\hline
RETAIN      & 0.000  & 0.000  & 0.000 & 0.000  & 0.000  & 0.000  \\
StageNet    & 0.000  & 0.000  & 0.000 & 0.000  & 0.000  & 0.000  \\
Dr. Agent   & 0.000  & 0.000  & 0.000 & 0.000  & 0.000  & 0.000  \\
AdaCare     & 0.000  & 0.000  & 0.000 & 0.000  & 0.000  & 0.000  \\
GRASP       & 0.000  & 0.000  & 0.000 & 0.000  & 0.000  & 0.000  \\
ConCare     & 0.000  & 0.000  & 0.000 & 0.000  & 0.000  & 0.000  \\
\bottomrule
\end{tabular}
\label{tab:pvalue_los}
\end{table}

\begin{table}[htbp]
\centering
\caption{\textit{P-values of naive models vs. models with the TA loss in early mortality predictions using t-test.} P-values are rounded to three decimal places; a value of 0.000 indicates a rounded p-value < 0.0005.}
\begin{tabular}{l|ccc|ccc}
\toprule
Dataset     & \multicolumn{3}{c|}{TJH} & \multicolumn{3}{c}{CDSL} \\
\midrule
Model       & AUPRC    & AUROC    & ES & AUPRC    & AUROC    & ES  \\
\midrule
MLP         & 0.000        & 0.000        & 0.195   & 0.896       & 0.570        & 0.430     \\
RNN         & 0.136      & 0.274      & 0.001   & 0.000         & 0.000         & 0.000      \\
LSTM        & 0.000        & 0.000        & 0.042   & 1.000         & 0.378       & 1.000      \\
GRU         & 0.584      & 0.176      & 0.000     & 0.000         & 0.000         & 0.000      \\
TCN         & 0.132      & 0.002      & 0.092   & 0.972       & 0.540        & 0.005    \\
Transformer & 0.817      & 0.881      & 0.000     & 0.966       & 0.000         & 0.000      \\
\hline
RETAIN      & 0.000        & 0.000        & 0.000     & 1.000         & 1.000         & 1.000      \\
StageNet    & 0.000        & 0.000        & 0.003   & 0.010        & 0.010        & 1.000      \\
Dr. Agent   & 0.030       & 0.000        & 0.236   & 0.000         & 0.000         & 0.000      \\
AdaCare     & 0.586      & 0.118      & 0.000     & 0.700         & 0.964       & 0.007    \\
GRASP       & 0.151      & 0.176      & 0.135   & 0.000         & 0.000         & 0.000      \\
ConCare     & 0.380       & 0.953      & 0.001   & 0.815       & 0.755       & 1.000      \\
\bottomrule
\end{tabular}
\label{tab:pvalue_mortality}
\end{table}

\section{Error Analysis}
\begin{figure}[htbp]
\centering
\begin{subfigure}{0.45\linewidth}
  \centering
  \includegraphics[width=\textwidth]{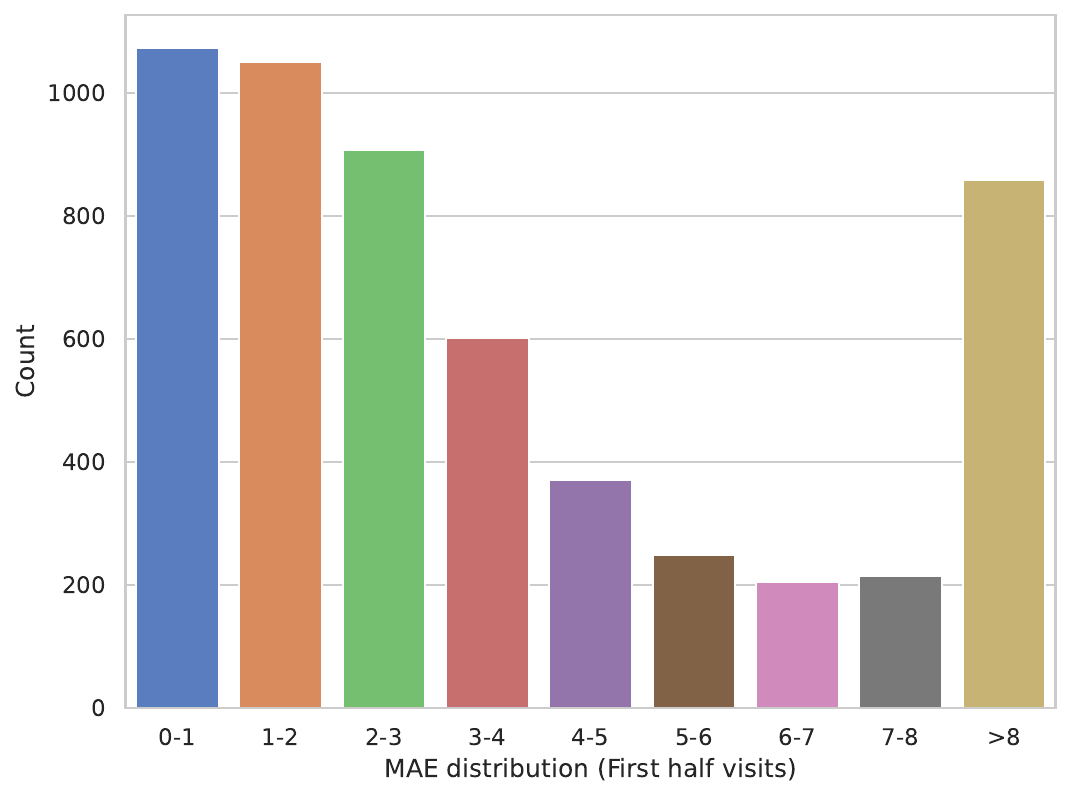}
  \caption{The first half of records (alive patients)}
  \label{fig:alive_first_half}
\end{subfigure}%
\begin{subfigure}{0.45\linewidth}
  \centering
  \includegraphics[width=\textwidth]{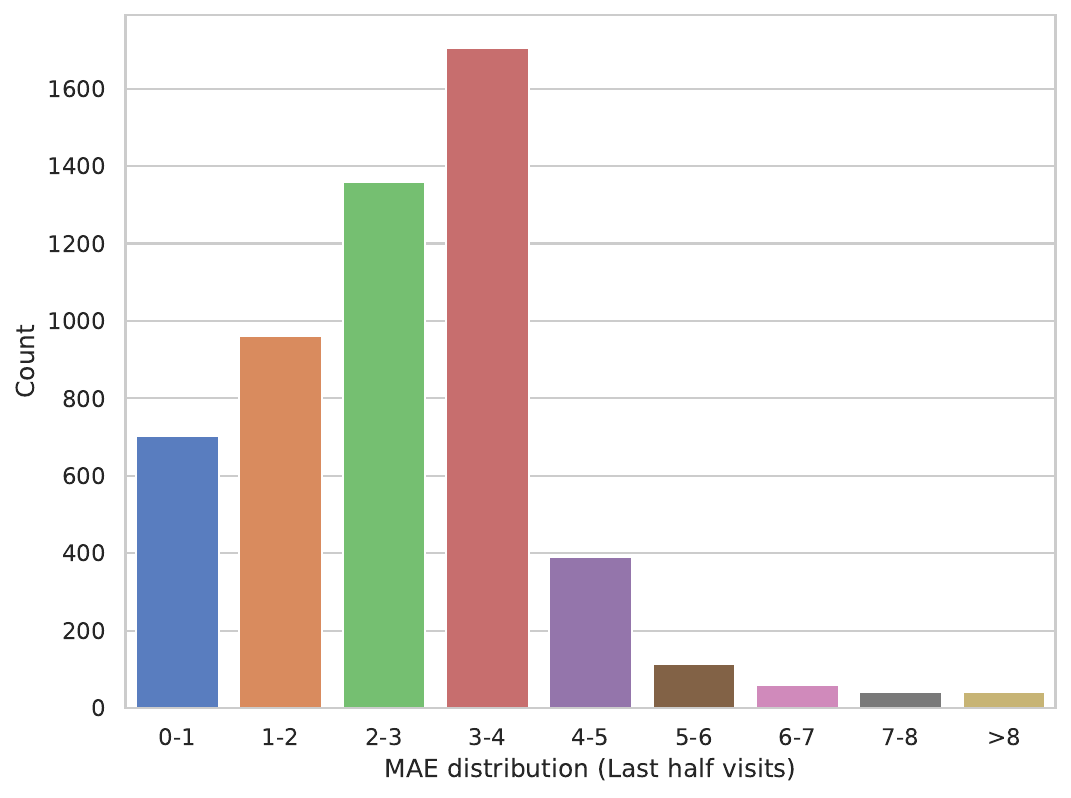}
  \caption{The last half of records (alive patients)}
  \label{fig:alive_last_half}
\end{subfigure}
\begin{subfigure}{0.45\linewidth}
  \centering
  \includegraphics[width=\textwidth]{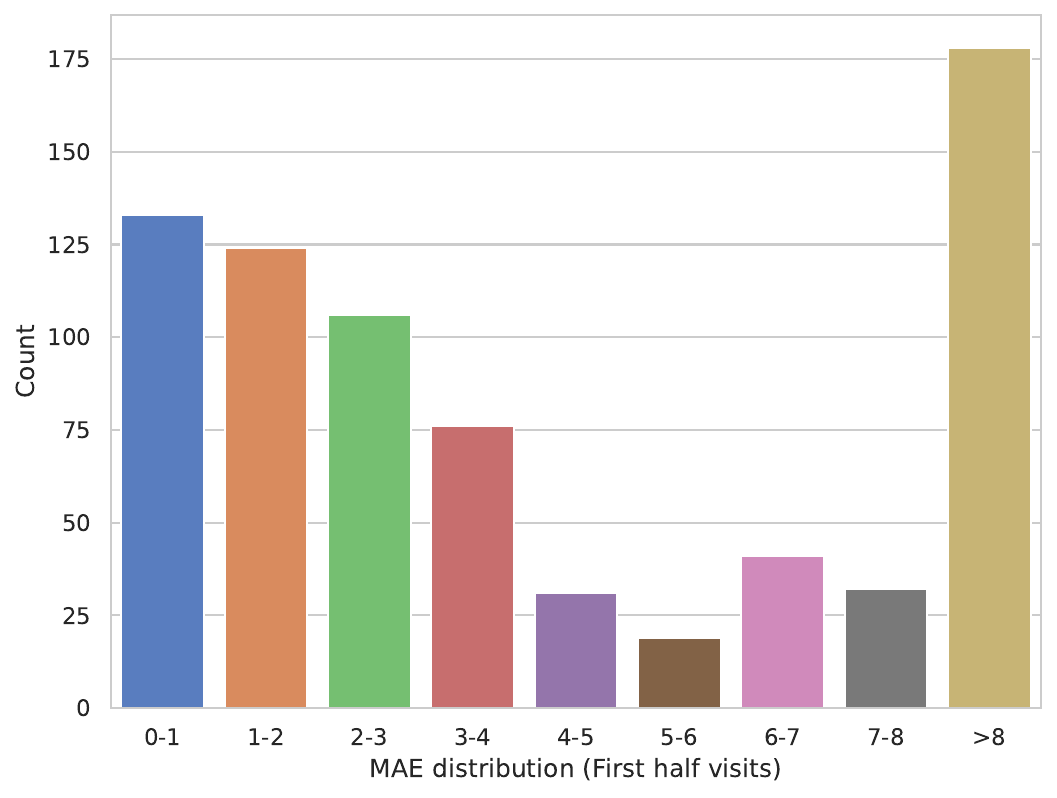}
  \caption{The first half of records (deceased patients)}
  \label{fig:died_first_half}
\end{subfigure}%
\begin{subfigure}{0.45\linewidth}
  \centering
  \includegraphics[width=\textwidth]{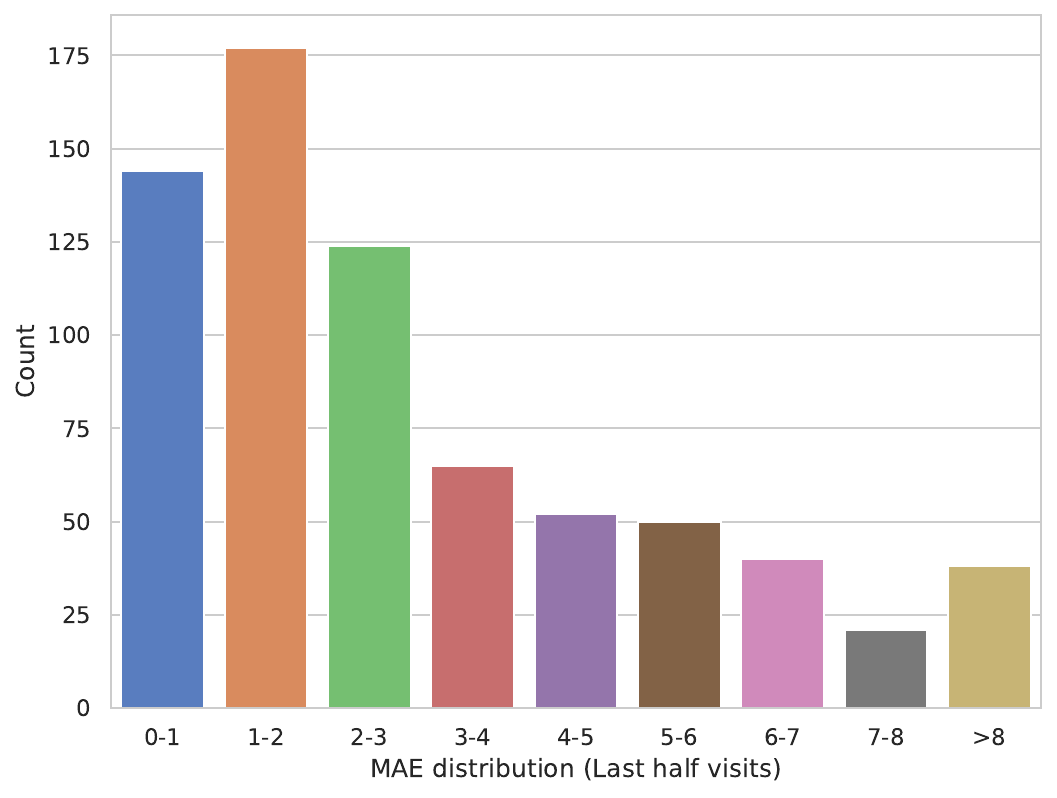}
  \caption{The last half of records (deceased patients)}
  \label{fig:died_last_half}
\end{subfigure}
\caption{\textit{MAE score distribution on the CDSL test set.} The figure compares the discrepancies in MAE performance distributions between the first and last half of patient records. The analysis utilizes the StageNet model with the multi-task setting.}
\label{fig:mae_distribution_cdsl}
\end{figure}

To more effectively analyze the sources of MAE, we plotted the MAE distributions for groups of patients who were alive and deceased. Additionally, we segmented each patient's record into two halves: the initial half and the latter half. As depicted in Figure~\ref{fig:mae_distribution_cdsl}, we observe that the MAE distributions for the first halves of the records are similar across both patient groups, with many records of MAE greater than 8. This is primarily because, in the initial stages of their hospital stay, patients' health statuses are often unstable, making it challenging for the model to accurately predict the remaining LOS. Conversely, during the latter half of the patients' stay, their health status tends to be more defined, resulting in a left-skewed MAE distribution for both groups. Notably, the MAE for deceased patients is predominantly less than 3, reflecting the more apparent nature of their health status and easier classification. In contrast, predicting LOS for alive patients is more complex, as their discharge from the ICU can be influenced by various external factors. Hence, their MAE values are larger (mostly clustered between 3 and 4).

These findings suggest that while the average MAE for the total cohort may be as large as the true LOS of patients, it can still effectively identify patients in critical condition, as indicated by the generally low MAE values for such cases. On the CDSL dataset, the similarity in MAE values is attributed to the fact that 87\% of the patients are alive, with their MAE predominantly ranging between 2 and 4, leading to closely aligned average values.

\section{Feature Statistics and Distributions}

\begin{figure}[htbp]
    \centering
    \includegraphics[width=\linewidth]{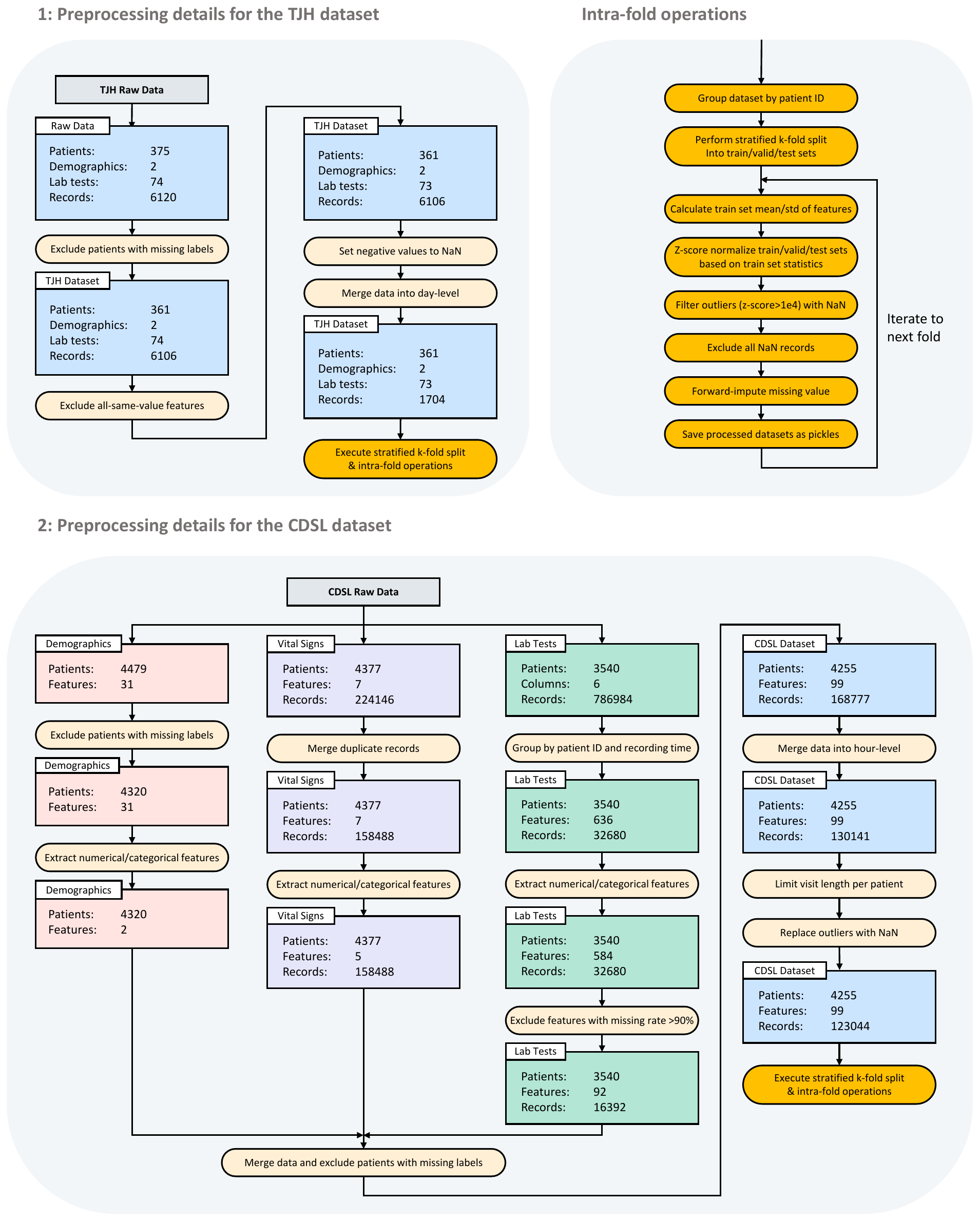}
    \caption{Data preprocessing details of two datasets.}
    \label{fig:preprocessing}
\end{figure}

\begin{figure}[htbp]
    \centering
    \includegraphics[width=0.6\linewidth]{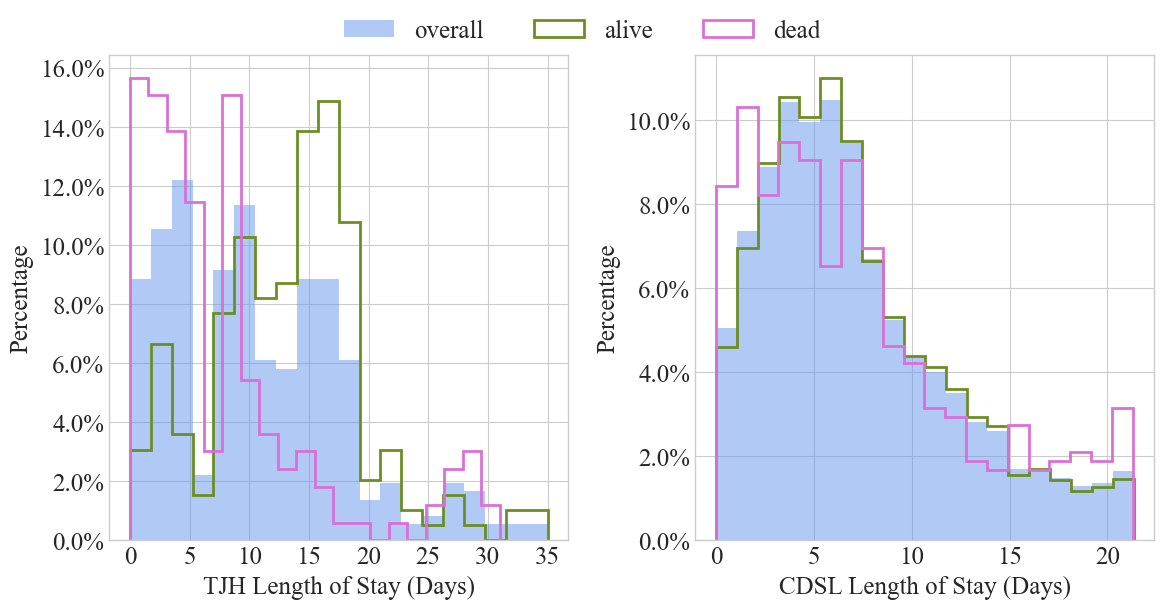}
    \caption{\textit{Length of stay distributions in two datasets.} To keep the figure informative, we only show the statistics in the 0\%-95\% range for the \textcolor{red}{CDSL} dataset.}
    % axis: 
    % - y: percentage
    % - x: days of los(CDSL 0~95%, TJH: total)
    \label{fig:los_fig}
\end{figure}

\begin{figure}[htbp]
    \centering
    \includegraphics[width=\linewidth]{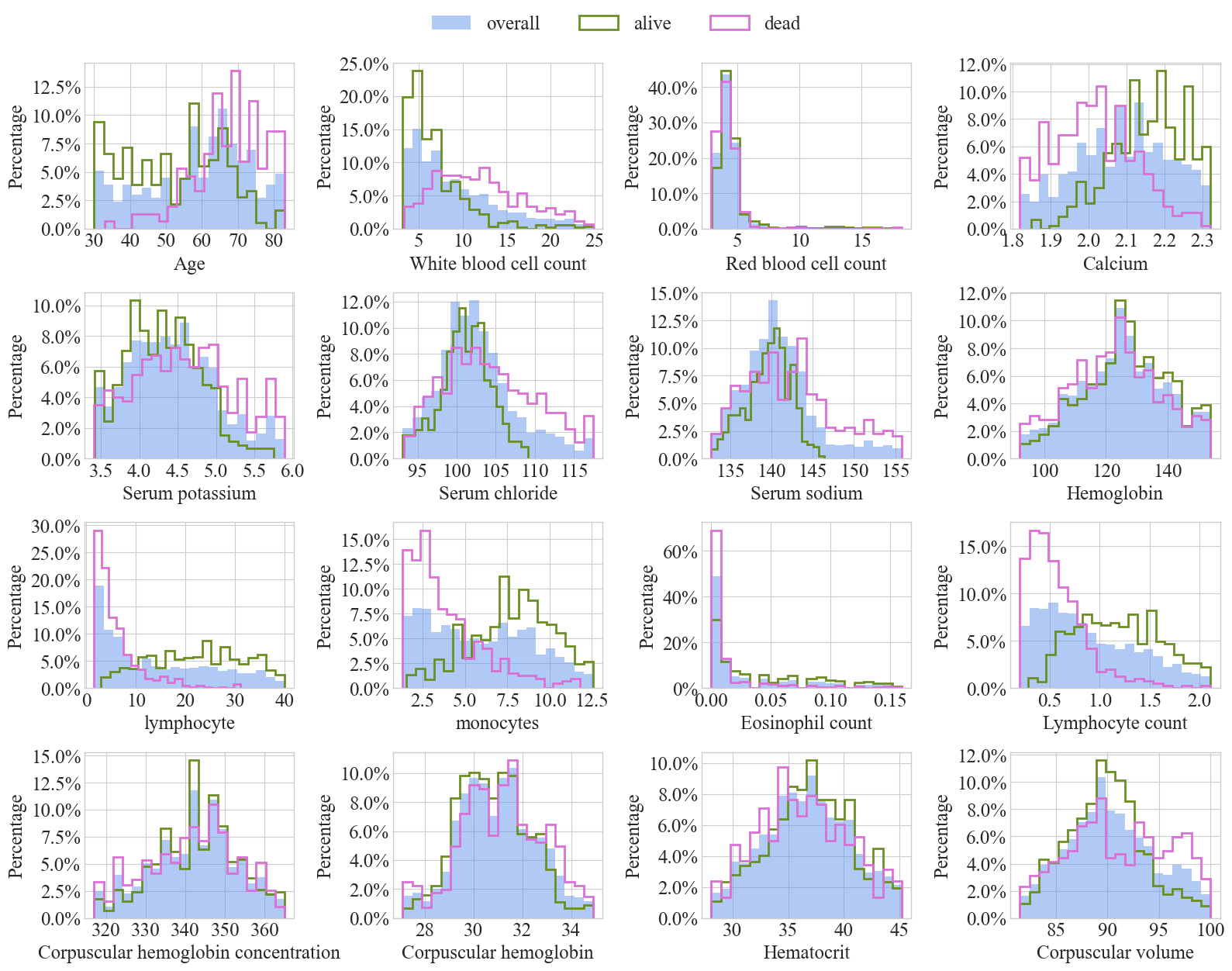}
    \caption{\textit{Distributions of 16 features with the lowest missing rates in the \textcolor{blue}{TJH} dataset.} We plot the data distributions of 16 features with the lowest missing rates for the overall, alive, and dead patients. The blue bars are distributions of total patients. The green and pink curves are distributions for alive and dead patients, respectively. To keep the figure informative, we only show the statistics in the 0\%-95\% range.} 
    \label{fig:distribution_tjh}
\end{figure}

\begin{figure}[htbp]
    \centering
    \includegraphics[width=\linewidth]{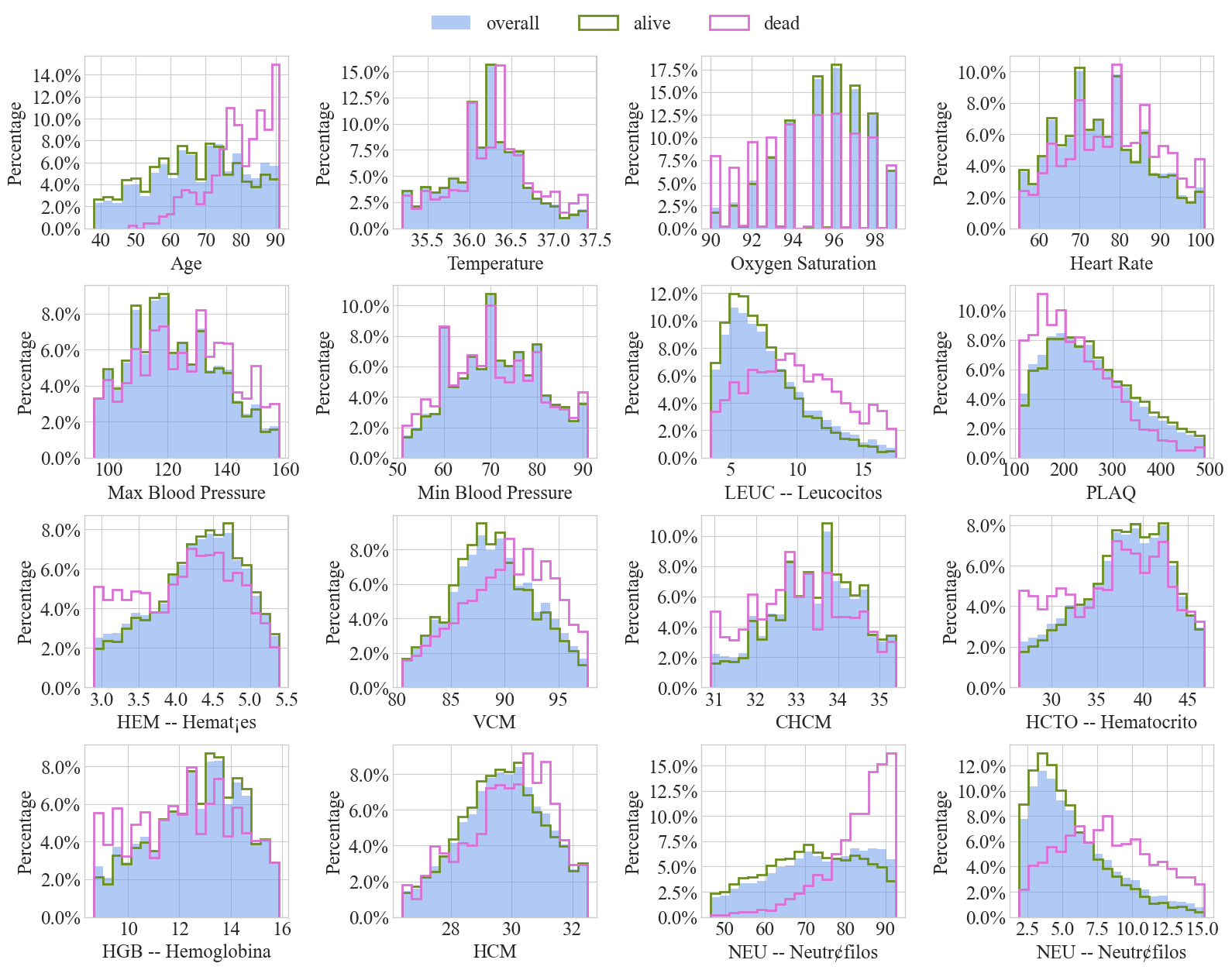}
    \caption{\textit{Distributions of 16 features with the lowest missing rates in the \textcolor{red}{CDSL} dataset.} We plot the data distributions of 16 features with the lowest missing rates for the overall, alive, and dead patients. The blue bars are distributions of total patients. The green and pink curves are distributions for alive and dead patients, respectively. To keep the figure informative, we only show the statistics in the 0\%-95\% range.}
    \label{fig:distribution_cdsl}
\end{figure}

We provide statistics of all features which are used in the modeling process in Table~\ref{tab:tjh_lab_tests_charactistics} and \ref{tab:cdsl_lab_tests_charactistics}. The data preprocessing details are shown in Figure~\ref{fig:preprocessing}. The length of stay distributions are shown in Figure~\ref{fig:los_fig}. We plot the distributions of 16 features with the lowest missing rates in two datasets in Figure~\ref{fig:distribution_tjh} and Figure~\ref{fig:distribution_cdsl}.

% red line: dead
% green line: alive

\clearpage
\begin{longtable}{lll}
\caption{\textit{Statistics of lab test features in the \textcolor{blue}{TJH} dataset.} The reported statistics are of the form $Median[Q1, Q3]$.} \\\hline
Feature name & Statistics & Missing Rate\\\hline
\label{tab:tjh_lab_tests_charactistics}
\endfirsthead
\caption{(Continued) Statistics of lab test features in the \textcolor{blue}{TJH} dataset.} \\\hline
Feature name & Statistics & Missing Rate\\\hline
\endhead
\hline
\endfoot
Hypersensitive cardiac troponinI          & 19.80 {[}4.20, 216.25{]}     & 29.23\%      \\
hemoglobin                                & 125.00 {[}113.00, 137.00{]}  & 55.52\%      \\
Serum chloride                            & 102.00 {[}98.95, 105.50{]}   & 55.58\%      \\
Prothrombin time                          & 14.70 {[}13.60, 16.60{]}     & 38.44\%      \\
procalcitonin                             & 0.10 {[}0.04, 0.40{]}        & 26.94\%      \\
eosinophils(\%)                           & 0.10 {[}0.00, 0.90{]}        & 55.40\%      \\
Interleukin 2 receptor                    & 680.50 {[}460.00, 1169.75{]} & 15.38\%      \\
Alkaline phosphatase                      & 69.00 {[}54.00, 95.00{]}     & 53.81\%      \\
albumin                                   & 32.40 {[}27.50, 36.70{]}     & 53.99\%      \\
basophil(\%)                              & 0.20 {[}0.10, 0.30{]}        & 55.40\%      \\
Interleukin 10                            & 5.80 {[}5.00, 12.50{]}       & 15.32\%      \\
Total bilirubin                           & 10.60 {[}7.30, 16.70{]}      & 53.81\%      \\
Platelet count                            & 181.00 {[}112.75, 250.25{]}  & 55.16\%      \\
monocytes(\%)                             & 5.70 {[}2.90, 8.70{]}        & 55.46\%      \\
antithrombin                              & 86.00 {[}74.00, 97.00{]}     & 19.19\%      \\
Interleukin 8                             & 15.95 {[}8.62, 33.58{]}      & 15.38\%      \\
indirect bilirubin                        & 5.40 {[}3.80, 8.00{]}        & 52.46\%      \\
Red blood cell distribution width         & 12.60 {[}12.00, 13.60{]}     & 53.40\%      \\
neutrophils(\%)                           & 82.05 {[}65.07, 92.20{]}     & 55.40\%      \\
total protein                             & 66.00 {[}61.10, 70.40{]}     & 53.81\%      \\
Treponema pallidum antibodies             & 0.05 {[}0.04, 0.07{]}        & 16.31\%      \\
Prothrombin activity                      & 81.50 {[}65.00, 95.00{]}     & 38.26\%      \\
HBsAg                                     & 0.01 {[}0.00, 0.01{]}        & 16.31\%      \\
mean corpuscular volume                   & 90.00 {[}86.90, 93.80{]}     & 55.40\%      \\
hematocrit                                & 36.70 {[}33.50, 39.90{]}     & 55.40\%      \\
White blood cell count                    & 7.60 {[}5.08, 12.52{]}       & 60.09\%      \\
Tumor necrosis factor$\alpha$                    & 8.60 {[}6.70, 11.58{]}       & 15.38\%      \\
corpuscular hemoglobin concentration      & 343.00 {[}333.75, 350.00{]}  & 55.40\%      \\
fibrinogen                                & 4.12 {[}3.06, 5.51{]}        & 32.86\%      \\
Interleukin 1$\beta$                            & 5.00 {[}5.00, 5.00{]}        & 15.38\%      \\
Urea                                      & 5.90 {[}4.00, 11.10{]}       & 53.99\%      \\
lymphocyte count                          & 0.80 {[}0.47, 1.32{]}        & 55.40\%      \\
PH value                                  & 6.50 {[}6.00, 7.00{]}        & 21.01\%      \\
Red blood cell count                      & 4.16 {[}3.67, 4.69{]}        & 60.09\%      \\
Eosinophil count                          & 0.01 {[}0.00, 0.06{]}        & 55.40\%      \\
Corrected calcium                         & 2.36 {[}2.27, 2.44{]}        & 52.82\%      \\
Serum potassium                           & 4.41 {[}3.95, 4.86{]}        & 55.75\%      \\
glucose                                   & 6.98 {[}5.53, 10.15{]}       & 45.07\%      \\
neutrophils count                         & 5.82 {[}3.08, 10.82{]}       & 55.40\%      \\
Direct bilirubin                          & 4.80 {[}3.20, 8.00{]}        & 53.81\%      \\
Mean platelet volume                      & 10.80 {[}10.10, 11.50{]}     & 50.12\%      \\
ferritin                                  & 711.60 {[}385.80, 1425.30{]} & 16.49\%      \\
RBC distribution width SD                 & 40.90 {[}38.50, 44.68{]}     & 53.40\%      \\
Thrombin time                             & 16.80 {[}15.60, 18.30{]}     & 32.86\%      \\
lymphocyte(\%)                            & 11.70 {[}4.00, 25.00{]}      & 55.46\%      \\
HCV antibody quantification               & 0.06 {[}0.04, 0.09{]}        & 16.31\%      \\
D-D dimer                                 & 2.12 {[}0.60, 21.00{]}       & 36.74\%      \\
Total cholesterol                         & 3.63 {[}3.01, 4.27{]}        & 53.87\%      \\
aspartate aminotransferase                & 27.00 {[}20.00, 42.00{]}     & 53.93\%      \\
Uric acid                                 & 244.00 {[}184.00, 332.60{]}  & 53.87\%      \\
HCO3-                                     & 23.50 {[}21.00, 25.90{]}     & 53.87\%      \\
calcium                                   & 2.09 {[}1.98, 2.19{]}        & 55.75\%      \\
NT-proBNP                                 & 571.50 {[}147.00, 2589.00{]} & 27.58\%      \\
Lactate dehydrogenase                     & 339.00 {[}217.00, 596.25{]}  & 53.87\%      \\
platelet large cell ratio                 & 30.80 {[}25.50, 37.10{]}     & 50.12\%      \\
Interleukin 6                             & 19.56 {[}4.66, 61.12{]}      & 15.61\%      \\
Fibrin degradation products               & 17.80 {[}4.00, 150.00{]}     & 19.19\%      \\
monocytes count                           & 0.41 {[}0.27, 0.58{]}        & 55.40\%      \\
PLT distribution width                    & 12.40 {[}11.10, 14.30{]}     & 50.12\%      \\
globulin                                  & 32.70 {[}29.70, 36.50{]}     & 53.81\%      \\
glutamyl transpeptidase                   & 34.00 {[}22.00, 58.00{]}     & 53.81\%      \\
International standard ratio              & 1.14 {[}1.03, 1.33{]}        & 38.26\%      \\
basophil count(\#)                        & 0.01 {[}0.01, 0.02{]}        & 55.40\%      \\
mean corpuscular hemoglobin               & 30.90 {[}29.70, 32.20{]}     & 55.40\%      \\
Activation of partial thromboplastin time & 39.20 {[}35.50, 44.10{]}     & 32.92\%      \\
Hypersensitive c-reactive protein         & 50.50 {[}5.35, 118.50{]}     & 42.66\%      \\
HIV antibody quantification               & 0.09 {[}0.07, 0.11{]}        & 16.26\%      \\
serum sodium                              & 140.30 {[}137.70, 143.30{]}  & 55.58\%      \\
thrombocytocrit                           & 0.21 {[}0.15, 0.27{]}        & 50.12\%      \\
ESR                                       & 28.00 {[}14.00, 45.50{]}     & 22.48\%      \\
glutamic-pyruvic transaminase             & 24.00 {[}16.00, 40.00{]}     & 53.87\%      \\
eGFR                                      & 88.10 {[}64.70, 104.20{]}    & 53.99\%      \\
creatinine                                & 76.00 {[}58.00, 98.00{]}     & 53.99\%      \\
\bottomrule
\end{longtable}

\begin{longtable}{lll}
\caption{\textit{Statistics of lab test features in the \textcolor{red}{CDSL} dataset.} The reported statistics are of the form $Median[Q1, Q3]$.} \\\hline
Feature name & Statistics & Missing Rate\\\hline
\label{tab:cdsl_lab_tests_charactistics}
\endfirsthead
\caption{(Continued) Statistics of lab test features in the \textcolor{red}{CDSL} dataset.} \\\hline
Feature name & Statistics & Missing Rate\\\hline
\endhead
\hline
\endfoot
ADW -- Coeficiente de anisocitosis          & 13.20 {[}12.10, 14.55{]}     & 8.34\%       \\
ADW -- SISTEMATICO DE SANGRE                & 13.85 {[}12.30, 15.30{]}     & 2.68\%       \\
ALB -- ALBUMINA                             & 3.10 {[}2.70, 3.50{]}        & 1.27\%       \\
AMI -- AMILASA                              & 69.00 {[}46.92, 111.90{]}    & 1.25\%       \\
AP -- ACTIVIDAD DE PROTROMBINA              & 79.00 {[}70.00, 87.00{]}     & 5.86\%       \\
APTT -- TIEMPO DE CEFALINA (APTT)           & 30.80 {[}28.10, 33.70{]}     & 5.37\%       \\
AU -- ACIDO URICO                           & 4.70 {[}3.50, 6.50{]}        & 0.53\%       \\
BAS -- Bas¢filos                            & 0.02 {[}0.01, 0.04{]}        & 8.31\%       \\
BAS -- SISTEMATICO DE SANGRE                & 0.02 {[}0.01, 0.04{]}        & 2.70\%       \\
BAS\% -- Bas¢filos \%                       & 0.30 {[}0.10, 0.50{]}        & 8.32\%       \\
BAS\% -- SISTEMATICO DE SANGRE              & 0.30 {[}0.10, 0.54{]}        & 2.70\%       \\
BD -- BILIRRUBINA DIRECTA                   & 0.26 {[}0.18, 0.40{]}        & 1.76\%       \\
BE(b) -- BE(b)                              & 2.80 {[}0.00, 5.90{]}        & 2.57\%       \\
BE(b)V -- BE (b)                            & 1.90 {[}-0.60, 4.50{]}       & 1.00\%       \\
BEecf -- BEecf                              & 3.20 {[}-0.30, 6.78{]}       & 2.57\%       \\
BEecfV -- BEecf                             & 2.50 {[}-0.60, 5.40{]}       & 1.00\%       \\
BT -- BILIRRUBINA TOTAL                     & 0.40 {[}0.27, 0.57{]}        & 1.68\%       \\
BT -- BILIRRUBINA TOTAL                     & 0.52 {[}0.36, 0.77{]}        & 3.64\%       \\
CA -- CALCIO                                & 8.30 {[}8.00, 8.70{]}        & 0.77\%       \\
CA++ -- Ca++ Gasometria                     & 4.41 {[}4.21, 4.64{]}        & 2.30\%       \\
CHCM -- Conc. Hemoglobina Corpuscular Media & 33.40 {[}32.50, 34.20{]}     & 8.39\%       \\
CHCM -- SISTEMATICO DE SANGRE               & 33.10 {[}32.10, 34.00{]}     & 2.70\%       \\
CK -- CK (CREATINQUINASA)                   & 66.10 {[}36.68, 140.00{]}    & 3.10\%       \\
CL -- CLORO                                 & 102.30 {[}98.90, 106.43{]}   & 2.19\%       \\
CREA -- CREATININA                          & 0.82 {[}0.64, 1.06{]}        & 8.28\%       \\
DD -- DIMERO D                              & 985.00 {[}525.75, 2138.75{]} & 6.31\%       \\
EOS -- Eosin¢filos                          & 0.03 {[}0.00, 0.13{]}        & 8.31\%       \\
EOS -- SISTEMATICO DE SANGRE                & 0.10 {[}0.01, 0.23{]}        & 2.70\%       \\
EOS\% -- Eosin¢filos \%                     & 0.50 {[}0.00, 1.90{]}        & 8.32\%       \\
EOS\% -- SISTEMATICO DE SANGRE              & 1.30 {[}0.10, 3.20{]}        & 2.70\%       \\
FA -- FOSFATASA ALCALINA                    & 72.70 {[}55.50, 104.90{]}    & 3.29\%       \\
FER -- FERRITINA                            & 908.55 {[}445.92, 1645.00{]} & 1.91\%       \\
FIB -- FIBRINàGENO                          & 562.00 {[}405.00, 724.00{]}  & 2.95\%       \\
FOS -- FOSFORO                              & 3.28 {[}2.70, 3.90{]}        & 2.15\%       \\
G-CORONAV (RT-PCR)                          & 1.00 {[}0.00, 1.00{]}        & 0.49\%       \\
GGT -- GGT                                  & 65.00 {[}30.00, 140.50{]}    & 5.53\%       \\
GLU -- GLUCOSA                              & 111.20 {[}94.00, 144.20{]}   & 7.67\%       \\
GOT -- GOT (AST)                            & 30.30 {[}20.80, 48.50{]}     & 7.42\%       \\
GPT -- GPT (ALT)                            & 32.60 {[}18.90, 61.00{]}     & 7.23\%       \\
HCM -- Hemoglobina Corpuscular Media        & 29.80 {[}28.60, 30.80{]}     & 8.39\%       \\
HCM -- SISTEMATICO DE SANGRE                & 29.90 {[}28.70, 30.90{]}     & 2.70\%       \\
HCO3 -- HCO3-                               & 27.40 {[}24.10, 31.10{]}     & 2.57\%       \\
HCO3V -- HCO3-                              & 27.20 {[}24.05, 29.90{]}     & 1.00\%       \\
HCTO -- Hematocrito                         & 38.30 {[}33.60, 42.00{]}     & 8.39\%       \\
HCTO -- SISTEMATICO DE SANGRE               & 34.40 {[}29.60, 39.20{]}     & 2.70\%       \\
HEM -- Hemat¡es                             & 4.33 {[}3.75, 4.78{]}        & 8.39\%       \\
HEM -- SISTEMATICO DE SANGRE                & 3.83 {[}3.25, 4.45{]}        & 2.70\%       \\
HGB -- Hemoglobina                          & 12.80 {[}11.10, 14.10{]}     & 8.39\%       \\
HGB -- SISTEMATICO DE SANGRE                & 11.30 {[}9.70, 13.10{]}      & 2.70\%       \\
INR -- INR                                  & 1.17 {[}1.09, 1.27{]}        & 5.87\%       \\
K -- POTASIO                                & 4.25 {[}3.86, 4.67{]}        & 8.10\%       \\
LAC -- LACTATO                              & 1.50 {[}1.10, 2.10{]}        & 2.46\%       \\
LDH -- LDH                                  & 532.00 {[}410.05, 704.00{]}  & 7.29\%       \\
LEUC -- Leucocitos                          & 7.25 {[}5.34, 10.17{]}       & 8.39\%       \\
LEUC -- SISTEMATICO DE SANGRE               & 7.70 {[}5.74, 10.85{]}       & 2.70\%       \\
LIN -- Linfocitos                           & 1.08 {[}0.72, 1.55{]}        & 8.39\%       \\
LIN -- SISTEMATICO DE SANGRE                & 1.24 {[}0.79, 1.78{]}        & 2.70\%       \\
LIN\% -- Linfocitos \%                      & 15.50 {[}8.80, 24.50{]}      & 8.38\%       \\
LIN\% -- SISTEMATICO DE SANGRE              & 16.33 {[}9.06, 26.30{]}      & 2.70\%       \\
MG -- MAGNESIO                              & 2.05 {[}1.81, 2.31{]}        & 2.40\%       \\
MONO -- Monocitos                           & 0.55 {[}0.36, 0.77{]}        & 8.32\%       \\
MONO -- SISTEMATICO DE SANGRE               & 0.60 {[}0.42, 0.82{]}        & 2.70\%       \\
MONO\% -- Monocitos \%                      & 7.60 {[}4.80, 10.40{]}       & 8.32\%       \\
MONO\% -- SISTEMATICO DE SANGRE             & 7.80 {[}5.15, 10.30{]}       & 2.70\%       \\
NA -- SODIO                                 & 138.00 {[}135.60, 140.70{]}  & 8.10\%       \\
NEU -- Neutr¢filos                          & 5.23 {[}3.46, 8.10{]}        & 8.39\%       \\
NEU -- SISTEMATICO DE SANGRE                & 5.39 {[}3.53, 8.43{]}        & 2.70\%       \\
NEU\% -- Neutr¢filos \%                     & 74.10 {[}62.90, 84.60{]}     & 8.39\%       \\
NEU\% -- SISTEMATICO DE SANGRE              & 71.70 {[}59.20, 83.40{]}     & 2.70\%       \\
PCO2 -- pCO2                                & 41.50 {[}35.50, 48.50{]}     & 2.57\%       \\
PCO2V -- pCO2                               & 44.00 {[}38.00, 50.00{]}     & 1.00\%       \\
PCR -- PROTEINA C REACTIVA                  & 34.75 {[}8.72, 93.02{]}      & 8.03\%       \\
PH -- pH                                    & 7.44 {[}7.39, 7.47{]}        & 2.57\%       \\
PHV -- pH                                   & 7.40 {[}7.36, 7.43{]}        & 1.00\%       \\
PLAQ -- Recuento de plaquetas               & 241.00 {[}176.00, 326.00{]}  & 8.39\%       \\
PLAQ -- SISTEMATICO DE SANGRE               & 238.00 {[}174.00, 315.00{]}  & 2.70\%       \\
PO2 -- pO2                                  & 82.80 {[}63.00, 106.00{]}    & 2.57\%       \\
PO2V -- pO2                                 & 42.00 {[}27.00, 59.00{]}     & 1.00\%       \\
PROCAL -- PROCALCITONINA                    & 0.14 {[}0.08, 0.30{]}        & 0.77\%       \\
PT -- PROTEINAS TOTALES                     & 5.62 {[}5.10, 6.10{]}        & 0.84\%       \\
SO2C -- sO2c (Saturaci¢n de ox¡geno)        & 95.67 {[}92.00, 97.67{]}     & 2.57\%       \\
SO2CV -- sO2c (Saturaci¢n de ox¡geno)       & 76.00 {[}48.00, 90.00{]}     & 1.00\%       \\
TCO2 -- tCO2(B)c                            & 28.70 {[}25.20, 32.53{]}     & 2.57\%       \\
TCO2V -- tCO2 (B)                           & 28.60 {[}25.10, 31.49{]}     & 1.00\%       \\
TP -- TIEMPO DE PROTROMBINA                 & 13.00 {[}12.10, 14.10{]}     & 5.86\%       \\
TROPO -- TROPONINA                          & 15.16 {[}7.95, 34.92{]}      & 0.99\%       \\
U -- UREA                                   & 41.60 {[}29.00, 61.10{]}     & 8.06\%       \\
VCM -- SISTEMATICO DE SANGRE                & 90.00 {[}87.00, 93.40{]}     & 2.70\%       \\
VCM -- Volumen Corpuscular Medio            & 88.90 {[}86.00, 92.30{]}     & 8.39\%       \\
VPM -- SISTEMATICO DE SANGRE                & 10.50 {[}9.80, 11.30{]}      & 2.67\%       \\
VPM -- Volumen plaquetar medio              & 10.40 {[}9.70, 11.10{]}      & 8.30\%       \\
VSG -- VSG                                  & 47.50 {[}16.00, 75.00{]}     & 0.47\%       \\
\bottomrule
\end{longtable}

\end{document}